\documentclass{article}

    \PassOptionsToPackage{numbers, compress}{natbib}

\usepackage[final]{neurips_2024}

\usepackage[utf8]{inputenc} 
\usepackage[T1]{fontenc}    
\usepackage{hyperref}       
\usepackage{url}            
\usepackage{booktabs}       
\usepackage{amsfonts}       
\usepackage{nicefrac}       
\usepackage{microtype}      
\usepackage{xcolor,authblk}    

\usepackage{hyperref}
\usepackage{url}

\usepackage{comment}
\usepackage{array, multicol, multirow} 
\usepackage{algpseudocode}
\usepackage{algorithm}
\usepackage{gensymb}
\usepackage{graphicx}
\usepackage{subcaption}
\usepackage{booktabs}
\usepackage{import}
\usepackage{pifont}
\usepackage{subcaption}
\usepackage{adjustbox}
\usepackage{multirow}
\usepackage{enumitem}
\usepackage{wrapfig}
\usepackage{soul}
\usepackage{amsmath}
\usepackage{amssymb,dutchcal}
\usepackage{diagbox}

\usepackage{amssymb}
\usepackage{amsfonts}
\usepackage{amsthm}
\usepackage{mathtools}
\usepackage{bbding}
\usepackage{multicol}
\usepackage{multirow}
\usepackage{scalerel}
\usepackage{subcaption}
\usepackage{color}
\usepackage{xcolor}
\usepackage{colortbl}
\usepackage{diagbox}
\usepackage{stfloats}
\usepackage{xpatch}
\usepackage{tablefootnote}

\makeatletter
\newif\ifdiagbox@cellEmpty@

\define@key{diagbox}{widest}{%
  \ifdiagbox@cellEmpty@
    \setkeys{diagbox}{innerwidth=\widthof{#1}}%
    \PackageWarning{diagbox}{innerwidth is set to \the\diagbox@wd}%
  \fi}

\define@key{diagbox}{highest}{%
  \ifdiagbox@cellEmpty@
    \setkeys{diagbox}{height=#1}%
  \fi}

\define@key{diagbox}{text}{%
  \def\diagbox@text{#1}}

\define@key{diagbox}{align}{%
  \in@{#1}{l, c, r}%
  \ifin@
    \def\diagbox@align{#1}%
  \else
    \PackageError{diagbox}%
      {The value of key "text" must be one of "l", "c", and "r".}%
  \fi}

\xpatchcmd{\diagbox@double}{%
  \setkeys{diagbox}{dir=NW,#1}%
}{%
  \if\relax\detokenize{#2}\relax
    \if\relax\detokenize{#3}\relax
      \diagbox@cellEmpty@true
      \setkeys{diagbox}{highest=1\line, align=l, text=\@empty}%
    \fi
  \fi
  \setkeys{diagbox}{dir=NW, #1}%
  \ifdiagbox@cellEmpty@
    \rlap{\makebox
      [\dimexpr\diagbox@wd-\diagbox@insepl-\diagbox@insepr\relax]%
      [\diagbox@align]%
      {\diagbox@text}}%
  \fi
}{}{\ddt}
\makeatother

\newfloat{figtab}{htb}{fgtb}
\makeatletter
  \newcommand\figcaption{\def\@captype{figure}\caption}
  \newcommand\tabcaption{\def\@captype{table}\caption}
\makeatother

\newtheorem{assumption}{Assumption}
\newtheorem{theorem}{Theorem}

\newtheorem{defi}{Definition}

\newtheorem{rmk}{Remark}

\newcommand{\calD}{\mathcal{D}}

\newcommand{\calF}{\mathcal{F}}

\newcommand{\EE}{\mathbb{E}}
\newcommand{\RR}{\mathbb{R}}

\newcommand{\gray}[1]{{\color{gray}#1}}

\usepackage{pifont}

\DeclareMathOperator*{\argmin}{arg\,min}

\newcommand\shortname{Ferrari}
\title{\shortname: Federated Feature Unlearning via \\ Optimizing Feature Sensitivity}

\author[2\thanks{equal contribution; authors are listed alphabetically by first name.}]{\,\,\,\,\,\,\,\,Hanlin Gu}
\author[1$^{*}$]{\,\,\,\,\,\,\,\,Win Kent Ong}
\author[1\thanks{corresponding author ({\it cs.chan@um.edu.my}).}]{\,\,\,\,\,\,Chee Seng Chan}
\author[2]{\,\,\,\,\,\,\,\,Lixin Fan}
\affil[1]{CISiP, Universiti Malaya, Malaysia}
\affil[2]{AI Lab, Webank, PR China}

\begin{document}

\maketitle

\newcommand{\eg}{\textit{e.g.,}}
\newcommand{\ie}{\textit{i.e.,}}

\newcommand{\eq}{Eq.}
\newcommand{\fig}{Fig.}
\newcommand{\tab}{Tab.}
\newcommand{\s}{Sec.}
\newcommand{\alg}{Alg.}
\newcommand{\appen}{Appendix}

\newcommand{\p}{$\pm$} 

\begin{abstract}
The advent of Federated Learning (FL) highlights the practical necessity for the {\it 'right to be forgotten'} for all clients, allowing them to request data deletion from the machine learning model's service provider. This necessity has spurred a growing demand for Federated Unlearning (FU). Feature unlearning has gained considerable attention due to its applications in unlearning sensitive, backdoor, and biased features. Existing methods employ the influence function to achieve feature unlearning, which is impractical for FL as it necessitates the participation of other clients, {\it if not all}, in the unlearning process. Furthermore, current research lacks an evaluation of the effectiveness of feature unlearning. To address these limitations, we define feature sensitivity in evaluating feature unlearning according to Lipschitz continuity. This metric characterizes the model output's rate of change or sensitivity to perturbations in the input feature. We then propose an effective federated feature unlearning framework called \shortname, which minimizes feature sensitivity. Extensive experimental results and theoretical analysis demonstrate the effectiveness of \shortname~across various feature unlearning scenarios, including sensitive, backdoor, and biased features. The code is publicly available at \href{https://github.com/OngWinKent/Federated-Feature-Unlearning}{https://github.com/OngWinKent/Federated-Feature-Unlearning}

\end{abstract}
\section{Introduction}
\label{sec: introduction}

Federated Learning (FL) \cite{konevcny2015federated,mcmahan2017communication,yang2019federated} allows for model training across decentralized devices or servers holding local private data samples, without the need to exchange them directly. An essential requirement within FL is the participants {\it \textquotedblleft right to be forgotten\textquotedblright}, as explicitly outlined in regulations such as the European Union General Data Protection Regulation (GDPR)\footnote{\url{https://gdpr-info.eu/art-17-gdpr/}} and the California Consumer Privacy Act (CCPA)\footnote{\url{https://oag.ca.gov/privacy/ccpa}} \cite{harding2019understanding}. To address this requirement, Federated Unlearning (FU) has been introduced, enabling clients to selectively remove the influence of specific subsets of their data from a trained FL model while preserving the model's accuracy on the remaining data \cite{che2023fast}.

Different from unlearning at the \textit{client, class, or sample} level \cite{FU_Challenges, FU_Survey, FU_Taxonomy} in FL, the feature unlearning \cite{MU_Features_Labels} holds significant applications across various scenarios. Firstly, in contexts where sentences contain sensitive information such as names and addresses \cite{MU_Features_Labels, MU_Attribute_Unlearning}, it becomes crucial to remove these sensitive components to prevent potential exposure through model inversion attacks \cite{model_inversion_attack_1, model_inversion_attack, generative_model_inversion, tabular_model_inversion}. Secondly, when datasets contain backdoor triggers that can compromise model integrity \cite{backdoor_fl, backdoor_fl_review, backdoor_fl_li, badnets}, it is imperative to eliminate these patterns. Thirdly, unlearning biased features becomes essential in scenarios where data imbalances significantly impact model accuracy due to bias \cite{bias_review, bias_survey, bias_sagawa, bias_seo}. However, existing works of FU focus on client, class, or sample unlearning \cite{FU_Challenges, FU_Survey, FU_Taxonomy} but do not address feature unlearning, limiting their ability to unlearn specific features across multiple data points.


There are two challenges in feature unlearning in FL. Firstly, evaluating the unlearning effectiveness for feature unlearning is difficult. Typically, unlearning effectiveness is assessed by comparing the unlearned model with a retrained model without the feature. However, building data without the feature is challenging; for example, training the data with noise or a black block on the feature region may cause severe degradation in model accuracy (see \s~\ref{subsec:challenge}). Secondly, previous work on feature unlearning within centralized machine learning settings \cite{MU_Features_Labels, MU_Attribute_Unlearning} is not practical for FL due to its requirement for access to all datasets, necessitating the participation of all clients.
 
 
To address the aforementioned limitations, we first define the feature sensitivity in \s~\ref{sec: methodology feature sensitivity} to evaluate the feature unlearning inspired by the Lipschitz continuity, which characterizes the rate of change or sensitivity of the model output to perturbations in the input feature. Then we propose a simple but effective federated feature unlearning method, called \shortname~({\bf F}ed{\bf er}ated Featu{\bf r}e Unle{\bf ar}n{\bf i}ng), by minimizing the feature sensitivity in \s~\ref{sec: methodology design rationale}. Our \shortname~framework offers three key advantages: Firstly, \shortname~ requires only local datasets from the unlearned clients for feature unlearning. Secondly, \shortname~demonstrates high practicality and efficiency, which support various feature unlearning scenarios, including sensitive, backdoor, and biased features and only consumes a few epochs of optimization. Thirdly, theoretical analysis in \s~\ref{subsec:verification} elucidates that our proposed \shortname~ achieves lower model utility loss compared to the exact feature unlearning. 

The key contributions of this work are summarized as follows:


\begin{itemize}
    \item We identify two key challenges for feature unlearning in FL. The first is how to successfully unlearn features without requiring the participation of other clients, as discussed in \s~\ref{subsec:challenge}. The second is how to design an effective evaluation method in federated feature unlearning. 
    \item We define the feature sensitivity and introduce this metric in federated feature unlearning in \s~\ref{sec: methodology}. By minimizing feature sensitivity, we propose an effective federated feature unlearning method, named \shortname, which enables clients to selectively unlearn specific features from the trained global model without requiring the participation of other clients.
    \item We provide a theoretical proof in Theorem \ref{theo:thm1}, which dictates that \emph{\shortname~achieves better model performances than exact feature unlearning}.  This analytical result is also echoed in the empirical evidence, highlighting \shortname's effectiveness across various settings, including the unlearning of sensitive, backdoor, and biased features.
\end{itemize}
\section{Related Work}
\label{sec: related work}


\paragraph{Machine Unlearning} Machine Unlearning (MU), introduced by Cao et al. \cite{machine_unlearning2015}, involves selectively removing specific training data from a trained model without retraining from scratch\cite{machineunlearningprivacy2021, eurocrypt-2020-30234}. It categorizes into exact unlearning \cite{Thudi2021OnTN,nguyen2022}, aiming to completely remove data influence with techniques like SISA \cite{SISA2021} and ARCANE \cite{ARCANE2022}, though with computational costs, and approximate unlearning \cite{Jie2023,Xu2023}, which reduces data impact through techniques like data manipulation (fine-tuning with mislabeled data \cite{AmnesiacUnlearning, Two-stageModelRetraining, UNDO, BoundaryUnlearning, LSF} or introducing noise \cite{huang2021unlearnable, di2022hidden, UNSIR}), knowledge distillation \cite{Chundawat_Tarun_Mandal_Kankanhalli_2023, StochasticTeacher, kurmanji2023towards,jung2024attack} (training a student model), gradient ascent \cite{Hoang_2024_WACV, abbasi2023covarnav, choi2023machine, goel2023adversarial} (maximizing loss associated with forgotten data), and weight scrubbing \cite{FisherForgetting, NTK, Mixed-PrivacyForgetting, CertifiedDataRemoval, foster2023fast, Guo_Certified_Data_Removal} (discarding heavily influenced weights). 



\paragraph{Federated Unlearning} In FL, traditional centralized MU methods are unsuitable due to inherent differences like incremental learning and limited dataset access \cite{FU_Knowledge_Distillation}. Research on Federated Unlearning (FU) mainly focuses on client, class, and sample unlearning \cite{FU_Challenges, FU_Survey, FU_Taxonomy}. Client unlearning, pioneered by Liu et al. \cite{FU_FedEraser} introducing FedEraser \cite{FU_FedEraser}, includes approaches like FRU \cite{FU_on-Device_Reccomendation},  FedRecover \cite{FU_FedRecover}, VeriFI \cite{FU_Verifi}, HDUS \cite{FU_HDUS}, KNOT \cite{FU_Asynchoronous}, FedRecovery \cite{FU_FedRecovery}, Knowledge Distillation \cite{FU_Knowledge_Distillation}, and Gradient Ascent \cite{FU_Efficiently_Erase_a_Client, FU_Subspace, FU_Erasing_Backdoors}, aiming to remove specific clients or recover poisoned global models. Class unlearning, introduced by Wang et al. \cite{FU_Discriminative_Pruning}, involves frameworks like discriminative pruning and Momentum Degradation \cite{FU_Momentum_Degradation} (MoDE) to remove entire data classes. Sample unlearning, initiated by Liu et al. \cite{FU_TheRightToBeForgotten}, targets individual sample removal within FL settings, with advancements like the QuickDrop \cite{FU_QuickDrop} framework and FedFilter \cite{FU_FedFilter} enhancing efficiency and effectiveness. Recent works, such as $FedMe^2$ by Xia et al. \cite{FU_Fedme2}, optimize both unlearning facilitation and privacy guarantees.

Existing literature on FU primarily focuses on client, class, or sample unlearning \cite{FU_Challenges, FU_Survey, FU_Taxonomy}. However, a significant gap arises when a client seeks to remove only sensitive features while remaining engaged in FL. Unfortunately, current FU approaches do not address this specific scenario, as they do not explore feature unlearning within FL settings. In contrast to prior works focusing on feature unlearning in centralized settings of MU, such as classification models \cite{MU_Features_Labels, MU_Attribute_Unlearning}, generative models \cite{MU_GAN1, MU_GAN3, MU_GAN4, MU_GAN5}, and large language models \cite{MU_LLM1, MU_LLM2, MU_LLM3}, this study uniquely addresses feature unlearning of classification model within the FL paradigm. This distinction arises because traditional feature unlearning methods in centralized settings of MU are impractical for FL scenarios, where participation from all clients is often infeasible. In such cases, the process fails if even a single client opts out of the operation.

Therefore, to fill this critical gap, we proposed a novel federated feature unlearning framework, namely \shortname~based on the concept of Lipschitz continuity \cite{SpectralNorm, LipschitzRegularization, LipschitzContinuityFundementalAspect}. Our proposed \shortname~requires exclusively from the target client's dataset while still preserving the model's original performance. Lipschitz continuity, a fundamental mathematical concept that measures a function's sensitivity to changes in its input variables \cite{usama2018robustnn,weng2018evaluating,yoshida2017spectral}, is central to our feature unlearning approach. For a detailed exposition of our proposed federated feature unlearning framework utilizing Lipschitz continuity, please refer to \s~\ref{sec: methodology}. To the best of our knowledge, this is the {\bf first work} in feature unlearning within FL settings that does not necessitate participation from all other clients, showcasing the potential to enhance privacy, practicality and efficiency.
\section{Challenges on Feature Unlearning in FL}
\label{sec: problem definition}

\subsection{Federated Feature Unlearning} 

Consider a federated system comprising $K$ clients and one server, collaboratively learning a global model $f_\theta$ as:
\begin{equation} \label{eq:objective}
    \min_{\theta} \sum_{k=1}^K\sum_{i=1}^{n_k}\frac{\ell(f_\theta(x_{k,i}), y_{k,i})}{n_1+\cdots+n_K},
\end{equation}
where $\ell$ is the loss, \eg~the cross-entropy loss, $\calD_k=\{(x_{k,i}, y_{k,i})\}_{i=1}^{n_k}$ is the dataset with size $n_k$ owned by client $k$. 
One client (\ie~referred to as the unlearn client $C_u$) requests the removal of a feature $\mathcal F$ from the global model $\theta$ such that $\theta$ does not retain any information about $\calF$. Specifically, we assume that the data $x \in \mathbb R^d$ and denote the j-th feature of $x$ by $x[j]$. The partial element of the data $x$ corresponding the feature $\calF$ is defined as $x[\mathcal{F}]$, \ie:
\begin{equation}
    x[\mathcal{F}] = \{ x[j], j\in \mathcal F\} 
\end{equation}
Therefore, the unlearn client $C_u$ aims to remove $\{x_{i,u}[\mathcal{F}]\}_{i=1}^{n_u} $, called unlearned data $\calD_u$. Denote $\calD_r = \calD-\calD_u$ to be the remaining data.

\subsection{Challenges for Feature Unlearning in FL}\label{subsec:challenge}

\begin{wrapfigure}{r}{0.4\textwidth}
    \centering
    \vspace{-10pt}
    \begin{adjustbox}{max width= 0.4\textwidth}
        \begin{tabular}{c c c}
             \includegraphics[width=0.15\textwidth]{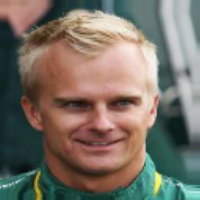}&  \includegraphics[width=0.15\textwidth]{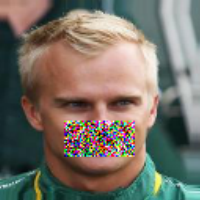}& \includegraphics[width=0.15\textwidth]{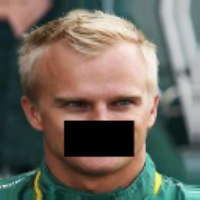}\\
             $x$& $\overline{x}_G$& $\overline{x}_B$\\
             Acc 95.86\%& 75.51\%& 68.37\%\\
        \end{tabular}
    \end{adjustbox}
    \caption{Sample data $x$ with Gaussian noise ($\overline{x}_G$) and black pixels ($\overline{x}_B$) perturbations, illustrating feature removal and performance comparison.}
    \label{fig: illustration feature perturbation}
\end{wrapfigure}

Unlike sample or class unlearning \cite{FU_Challenges, FU_Survey, FU_Taxonomy}, evaluating the unlearning effectiveness for feature unlearning is difficult. Typically, unlearning effectiveness is assessed by comparing the unlearned model with a retrained model trained on remaining data $\calD_r$. However, building $\calD_r$ for the feature unlearning takes much work. For example, suppose we want to remove the mouth from a face image. In that case, one possible solution is to replace the mouth region with Gaussian noise or black block, as illustrated in Fig. \ref{fig: illustration feature perturbation}. However, this added Gaussian noise or black block can adversely affect model training and degrade performance, \eg~the degradation of model accuracy is beyond 27\%. 

Another challenge is implementing feature unlearning for $C_u$ without the help of other clients. Previous work on feature unlearning \cite{MU_Features_Labels, MU_Attribute_Unlearning} typically requires access to the remaining data, necessitating the participation of other clients in the FL process. This requirement is impractical in the FL context, as other clients may be unwilling or unable to share data or computational resources. Therefore, finding a method to effectively unlearn features without relying on other clients is crucial to maintain the model accuracy and practicality in the FL settings.

\section{The Proposed Method}
\label{sec: methodology}

In this section, we introduce feature sensitivity (see Def. \ref{def:feature-sensitivity}) in \s~\ref{sec: methodology feature sensitivity} to evaluate the effectiveness of feature unlearning. We then propose \shortname~based on this concept in \s~\ref{sec: methodology design rationale}). Finally, we demonstrate that \shortname~achieves a lower utility loss compared to exact feature unlearning in \s~\ref{subsec:verification}).

\subsection{Feature Sensitivity}
\label{sec: methodology feature sensitivity}

\begin{wrapfigure}{r}{0.5\textwidth}
    \vspace{-20pt}
    \begin{minipage}{0.5\textwidth}
        \begin{algorithm}[H]
        \caption{Federated Feature Unlearning}
        \label{alg: federated feature unlearning pseudocode}
        \begin{algorithmic}[1] 
        
            \Statex \textbf{Input:} Unlearn client $C_u$, Local dataset $\mathcal{D}_u$ with data size $n_u$, Unlearn feature $\{\mathcal{F}_i\}_{i=1}^N$, Global model parameters $\theta$, Gaussian noise $\sigma$, Learning rate $\eta$, Sample number $N$
            \Statex \textbf{Output:} Unlearned model parameters $\theta^u$
        
\State \gray{$\triangleright$ \textit{The unlearn client $C_u$ performs:}}
        
            \For {$(x, \mathcal{F}_i)$ in $(\mathcal{D}_u,\{\mathcal{F}_i\}_{i=1}^N)$} 
            \State $\theta^u=  \theta$ 
                \For {$i=1$ \textbf{to} $N$}
                    \State Sample $\delta_{\calF_i}$ according to Eq. \eqref{eq:sampling}
                    \State Compute $L_{i} = \frac{\|f_{\theta^u}(x) - f_{\theta^u}(x + \delta_{\calF,i})\|_2}{\|\delta_{\calF,i}\|_2}$ 
                \EndFor
                \State $L = \frac{1}{N}\sum_{ i=1}^NL_{i}$ 
                \State $\theta^u \leftarrow  \theta^u - \eta \cdot \nabla_{\theta^u}(L)$ 
            \EndFor
            \State Upload $\theta^u $ to the server
            \State \gray{$\triangleright$ \textit{The server performs:}}
             \State Replace the global model $\theta$ with the $\theta^u$
            \State \textbf{return} $\theta^u$
        \end{algorithmic}
        \end{algorithm}
    \end{minipage}
    \vspace{-15pt}
\end{wrapfigure}

Inspired by Lipschitz Continuity \cite{LipschitzRegularization,LipschitzContinuityFundementalAspect,weng2018evaluating}, which provides an approximate method for removing information from images by perturbing the input data and observing the effect on the output, we introduce the concept of \textbf{feature sensitivity} 
$s$ as Def. \ref{def:feature-sensitivity}. This metric measures the memorization of a model $f_\theta$ for the feature $\calF$
by considering the local changes in the given input rather than the global change as defined in the traditional Lipschitz continuity.

\begin{defi}\label{def:feature-sensitivity}  The feature sensitivity $s$ of the model $f$ with respect to the feature $\calF$ on the data $(x,y)$ is defined as:
\begin{equation}
    s = \EE_{\delta_\calF}\frac{\|f(x) - f(x + \delta_\calF)\|_2}{\|\delta_\calF\|_2},
\end{equation}
where $\delta_\calF$ denote the perturbation on feature $\calF$.
\end{defi}

Def. \ref{def:feature-sensitivity} characterizes the rate of change or sensitivity of the model output to perturbations in the input data.  A small feature sensitivity $s$ represents the model $f$ doesn't memorize the feature $\calF$. This definition does not require building the remaining data, as it considers the expectation over the perturbation $\delta_\calF$. Specifically, it represents the average output change rate over any magnitude of the perturbation. Furthermore, we will provide the relationship between Def. \ref{def:feature-sensitivity} and exact feature unlearning in \s~\ref{subsec:verification}.

\begin{rmk}
The perturbation $\delta_\calF$ can be chosen from various distributions, such as the Gaussian distribution, the uniform distribution, and so on.
\end{rmk}

\subsection{\shortname}
\label{sec: methodology design rationale}

\begin{figure}[t]
    \centering
    \includegraphics[width= \textwidth]{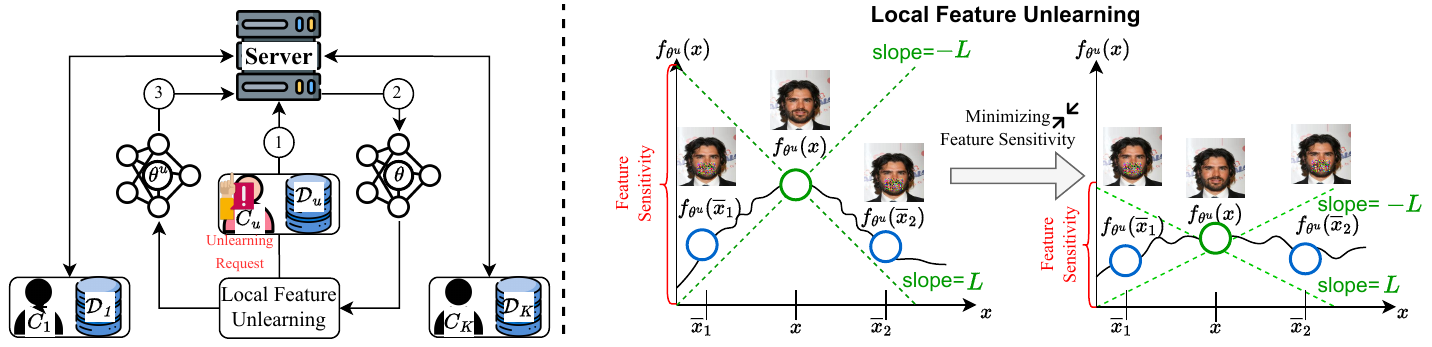}
    \caption{Overview of our proposed \shortname~framework: Initiated by the feature unlearning request from the unlearn client $C_u$, the server initializes the trained global model $\theta$ to $C_u$ for local feature unlearning. Upon completion, $C_u$ uploads the unlearned model $\theta^u$ to the server. Local feature unlearning minimizes the Lipschitz constant $L$ between the original input and its perturbed feature subset, reducing feature sensitivity yet preserving the overall model performance.}
    \label{fig: proposed method}
\end{figure}


As discussed the feature sensitivity $s$ in \s~\ref{sec: methodology feature sensitivity}, the core idea of the proposed method \shortname~is to achieve the feature unlearning by minimizing the feature sensitivity. More specifically, it controls the change in the model’s output relative to changes in the input within the feature region, \ie~the slope, to prevent the model from memorizing the feature as illustrated in Fig. \ref{fig: proposed method}.  

One unlearning client $C_u$ requests to unlearning the feature $\calF$. The proposed \shortname~aims to unlearn the global model $\theta$ to $\theta^u$.  The proposed method can be divided into three steps (see details in \alg~\ref{alg: federated feature unlearning pseudocode}). In order to compute the feature sensitivity, the perturbation $\delta_\calF$ in terms of the feature $\calF$ is \textbf{firstly} computed as the following (take the Gaussian distribution as an example):
\begin{equation} \label{eq:sampling}
\delta_\calF[j] = \left\{
\begin{aligned}
\sim N(0,\sigma^2) \quad &j\in \calF\\
0   \qquad \qquad & \text{Otherwise}\\
\end{aligned}
\right.
\end{equation} 

\textbf{Secondly,} we leverage a finite sample Monte Carlo approximation to the maximization as Def. \ref{def:feature-sensitivity} as:
\begin{equation}\label{eq:sampling1}
\begin{split}
       \EE_{\delta_\calF}\frac{\|f_\theta(x) - f_\theta(x + \delta_\calF)\|_2}{\|\delta_\calF\|_2} 
    \sim \frac{1}{N}\sum_{i=1}^N \frac{\|f_{\theta}(x) - f_{\theta}(x + \delta_{\calF,i})\|_2}{\|\delta_{\calF,i}\|_2},
\end{split}
\end{equation}
where $\delta_{\calF,i}$ is $i_{th}$ sampling as Eq. \eqref{eq:sampling}.

\textbf{Finally}, for the unlearning client $C_u$ who aims to remove the feature $\calF$ from his/her data $\calD_u$, the unlearned model $\theta^u$ is obtained as the following:
\begin{equation} \label{eq:loss}
    \theta^u = \argmin_\theta \EE_{(x,y) \in \calD_u}\frac{1}{N}\sum_{i=1}^N\frac{\|f_\theta(x) - f_\theta(x + \delta_{\calF,i})\|_2}{\|\delta_{\calF,i}\|_2},
\end{equation}
where Eq. \eqref{eq:loss} is computed over the dataset $\calD_u$. Noted that the proposed \shortname~based on Def. \ref{def:feature-sensitivity} doesn't need the participation of other clients.
\begin{rmk}
    When the unlearning happens during the federated training, the unlearning clients would also optimize the training loss and feature sensitivity simultaneously, \ie, $\EE_{(x,y)\in \calD}\big(\ell (f_\theta(x), y) +  \lambda \EE_{\delta_\calF}\frac{\|f_\theta(x) - f_\theta(x + \delta_\calF)\|_2}{\|\delta_\calF\|_2} \big ),$ where $\lambda$ is a coefficient.
\end{rmk}

\subsection{Theoretical Analysis of the Utility loss for \shortname} \label{subsec:verification}

As illustrated in \s~\ref{subsec:challenge}, retraining the model without the feature may affect the model accuracy seriously. Suppose the feature is successfully removed when the norm of perturbation is larger than $C$. We firstly define the utility loss $\ell_{1}$ with unlearning feature directly, \ie, \textbf{the exact feature unlearning}:
\begin{equation}
    \ell_{1} = \min_{\|\delta_\calF\| \geq C}\EE_{(x,y)\in \calD} \min_\theta \ell \big( f_\theta(x+\delta_\calF), y \big ) 
\end{equation}
And we define the maximum utility loss with the norm perturbation lower than $C$ as:
\begin{equation}
    \ell_{2} = \max_{\|\delta_\calF\| \leq C}\EE_{(x,y)\in \calD} \min_\theta \ell \big( f_\theta(x+\delta_\calF), y \big ) 
\end{equation}

\begin{assumption}\label{assum:1}
Assume $\ell_2 \leq \ell_1$
\end{assumption}
Assumption \ref{assum:1} elucidates that the utility loss associated with a perturbation norm lower than $C $ is smaller than the utility loss when the perturbation norm is greater than $C$. This assumption is logical, as larger perturbations would naturally lead to a greater utility loss.

\begin{assumption} \label{assum:2}
Suppose the federated model achieves zero training loss.
\end{assumption}
We have the following theorem to elucidate the relation between feature sensitivity removing via \alg~\ref{alg: federated feature unlearning pseudocode} and exact unlearning (see proof in \appen~\ref{sec: appendix theorem 1 proof}, including the extension for the non-zero training loss assumption).

\begin{theorem} \label{theo:thm1}
If Assumptions \ref{assum:1} and~\ref{assum:2} hold, the utility loss of unlearned model obtained using \alg~\ref{alg: federated feature unlearning pseudocode} is lower than the utility loss with exact feature unlearning, \ie,
\begin{equation}
    \ell_u \leq \ell_1,
\end{equation}
where $\ell_u =\EE_{(x,y)\in \calD}\ell (f_{\theta^u}(x), y)$
\end{theorem}
Theorem \ref{theo:thm1} showcases that the proposed method \shortname, results in a utility loss ($\ell_u $) that is lower than the utility loss incurred when the feature is removed, and the model is retrained, \ie~the process of exact feature unlearning.

\begin{rmk}
To further evaluate the effectiveness of feature unlearning based on feature sensitivity, we employ model inversion attacks \cite{model_inversion_attack_1, model_inversion_attack} to determine if the feature can be reconstructed and employ attention maps to assess if the model still focuses on the unlearned feature, as described in \s~\ref{sec: effectiveness sensitive}.
\end{rmk}
\section{Experimental Results}
\label{sec: results}

This section presents the empirical analysis of the proposed \shortname~framework in terms of effectiveness, utility, and time efficiency in sensitive, backdoor and biased feature unlearning scenarios.

\subsection{Experimental Setup}
\label{sec: experimental setup}

\begin{figure}[t]
    \centering
    \begin{minipage}[b]{0.55\textwidth}
        \centering
        \begin{adjustbox}{max width=\textwidth}
            \begin{tabular}{c c c c c}
                \includegraphics[width=0.3\textwidth]{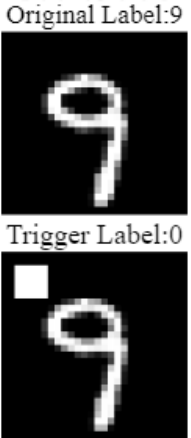} & \includegraphics[width=0.3\textwidth]{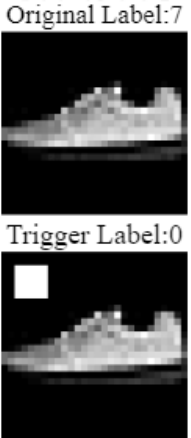}& \includegraphics[width=0.3\textwidth]{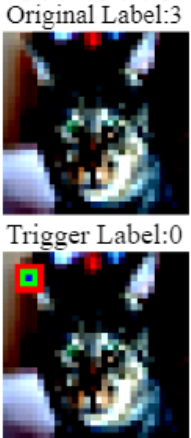}& \includegraphics[width=0.3\textwidth]{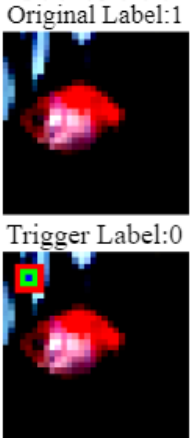}&
                \includegraphics[width=0.3\textwidth]{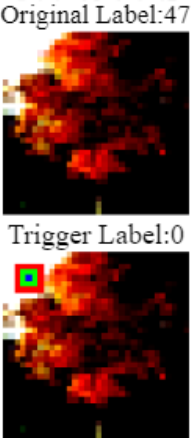}\\
                MNIST & FMNIST & CIFAR-10 & CIFAR-20 & CIFAR-100\\ 
            \end{tabular}
        \end{adjustbox}
        \caption{Pixel-pattern backdoor feature.}
        \label{fig: backdoor sample}
    \end{minipage}
    \hfill
    \begin{minipage}[b]{0.43\textwidth}
        \centering
        \begin{adjustbox}{max width=\textwidth}
            \begin{tabular}{c c}
                \includegraphics[width=0.41\textwidth]{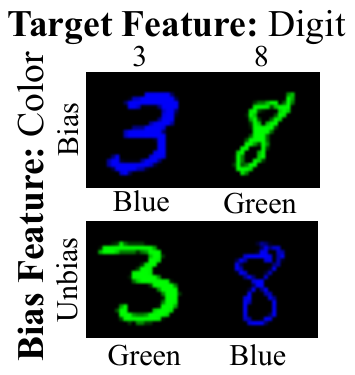} &
                \includegraphics[width=0.44\textwidth]{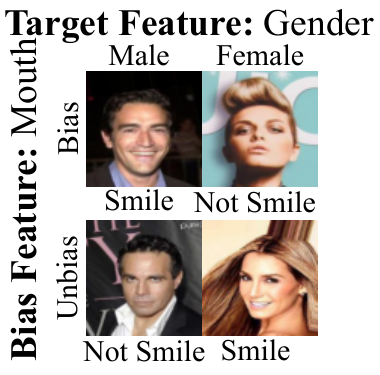} \\
                CMNIST & CelebA\\ 
            \end{tabular}
        \end{adjustbox}
        \caption{Biased datasets distribution.}
        \label{fig: bias dataset}
    \end{minipage}
\end{figure}

\paragraph{Unlearning Scenarios} 
\textit{Sensitive Feature Unlearning}: We simulate the removal of sensitive features from the $\mathcal{D}_u$ to fulfill the request of $C_u$ due to privacy concern. Specifically, we remove 'mouth' from CelebA \cite{Celeba}, 'marital status' from Adult \cite{adult_income_dataset}, and 'pregnancies number' from Diabetes \cite{diabetes_dataset}. Therefore, our proposed \shortname~aims to remove the influence of these requested features. 

\textit{Backdoor Feature Unlearning}: We simulate a pixel-pattern backdoor attack by $C_u$ based on BadNets \cite{badnets} within a FL framework \cite{backdoor_fl, backdoor_fl_review, backdoor_fl_li}. $C_u$ injects a pixel-pattern backdoor feature and trigger label into its $\mathcal{D}_u$ during training, as shown in \fig~\ref{fig: backdoor sample}. Consequently, our proposed \shortname~aims to remove the influence of these backdoor features and restore the model's original performance.

\textit{Biased Feature Unlearning}: We simulate the bias dataset $\mathcal{D}_u$ of the $C_u$ and the unbias dataset $\mathcal{D}_r$ with a bias ratio of 0.8, as shown in \fig~\ref{fig: bias dataset}. This results in a global model biased towards the biased dataset \cite{bias_dje, bias_chen} due to unintended feature memorization \cite{bias_seo}. In CMNIST \cite{Mnist}, the model focuses on color patterns instead of digits, and in CelebA \cite{Celeba}, it learns mouth features instead of facial features for gender classification. Therefore, our proposed \shortname~aims to mitigate these bias-inducing features and restore model performance.


\paragraph{Hyperparameters \& Datasets \& Model} We simulate HFL with $K=10$ clients under an IID setting, each holding $10\%$ of the datasets, except for the biased feature unlearning experiment with a bias ratio of 0.8. For federated feature unlearning experiments, we set hyperparameters: learning rate $\eta=0.0001$, sample size $N=20$, and random Gaussian noise with standard deviation ranging from $0.05 \leq \sigma \leq 1.0$ (see \s~\ref{sec: ablation studies}) across iterations of $N$. Experiments are repeated over five random trials, and results are reported as mean and standard deviation. We employ ResNet18 \cite{Resnet18} on image datasets: MNIST \cite{Mnist}, Colored-MNIST (CMNIST) \cite{Mnist}, Fashion-MNIST \cite{fmnist}, CIFAR-10, CIFAR-20, CIFAR-100 \cite{Cifar10} and ImageNet \cite{imagenet}. For tabular datasets, such as Adult Census Income (Adult) \cite{adult_income_dataset} and Diabetes \cite{diabetes_dataset}, we used a fully-connected neural network linear model. Additionally, we utilize the transformer-based BERT model \cite{bert} for the text dataset, specifically the IMDB movie reviews dataset \cite{imdb}. We conduct experiments on a single NVIDIA A100 GPU. Further details are in \appen~\ref{sec: appendix experimental setup}.

\paragraph{Evaluation Metrics} We assess effectiveness by measuring feature sensitivity (see Section \ref{sec: methodology feature sensitivity}) and conducting a model inversion attack (MIA) \cite{model_inversion_attack_1, model_inversion_attack,generative_model_inversion,tabular_model_inversion} to determine the attack success rate (ASR). The goal is to achieve low feature sensitivity and ASR, indicating successful unlearning sensitive features. Backdoor and biased feature unlearning are evaluated by comparing accuracy on the retain dataset $\mathcal{D}_r$ ($Acc_r$) and the unlearn client dataset $\mathcal{D}_u$($Acc_u$). Low $Acc_u$ indicates high effectiveness for backdoor unlearning, while similar accuracy ($Acc_r \approx Acc_u$) reflects fairness and effectiveness in biased feature unlearning. Qualitatively, effectiveness is assessed using MIA-reconstructed images (sensitive) and GradCAM \cite{gradcam} attention maps (backdoor and biased). The utility is measured by test dataset $\mathcal{D}_t$ accuracy ($Acc_t$), with higher values indicating stronger utility. Time efficiency is evaluated by comparing the runtime of each baseline.


\paragraph{Baselines} We compare our proposed \shortname~against the models of Baseline, Retrain, Fine-tune, FedCDP \cite{FU_Discriminative_Pruning} and FedRecovery \cite{FU_FedRecovery}. Additional details are provided in \appen~\ref{sec: appendix experimental setup}.

\subsection{Utility Guarantee}
\label{sec: utility guarantee}


To evaluate the utility of \shortname, we measure $Acc_t$ on $\mathcal{D}_{t}$, where a higher $Acc_t$ indicates greater utility (\tab~\ref{tab: utility}). Although the Fine-tune method shows high $Acc_t$ in the backdoor feature unlearning scenario with a clean dataset, its unlearning effectiveness is very low (see \s~\ref{sec: effectiveness backdoor}). This problem worsens with FedCDP \cite{FU_Discriminative_Pruning} and FedRecovery \cite{FU_FedRecovery}, which suffer significant $Acc_t$ declines, reducing model utility and making them unsuitable for feature unlearning. In contrast, \shortname~achieves the highest model utility in sensitive and biased feature unlearning scenarios, with the highest $Acc_t$ among baselines, minimal deterioration, and the greatest unlearning effectiveness across all scenarios.

\begin{table}[t]
    \centering
    \begin{adjustbox}{max width=\textwidth}
    \begin{tabular}{c |c| c|c |c |c |c |c |c }
        \hline
        \multirow{2}{*}{\textbf{Scenarios}} & \multirow{2}{*}{\textbf{Datasets}} & \multirow{2}{3.3em}{\textbf{Unlearn Feature}} & \multicolumn{6}{c}{\textbf{Accuracy(\%)}}\\
        \cline{4-9}
        
         & & & \textbf{Baseline} & \textbf{Retrain} & \textbf{Fine-tune} & \textbf{FedCDP\cite{FU_Discriminative_Pruning}} & \textbf{FedRecovery\cite{FU_FedRecovery}} & \textbf{\shortname~(Ours)}\\
         \hline
         
         \multirow{4}{*}{\textbf{Sensitive}} & \textbf{CelebA} & Mouth & 94.87 \p 1.38& 79.46 \p 2.32& 62.79 \p 1.62& 34.03 \p 4.20& 29.78 \p 6.69 & \textbf{92.26 \p 1.73}\\
         
         & \textbf{Adult} & Marriage & 82.45 \p 2.59& 65.27 \p 0.58& 61.02 \p 1.05& 30.19 \p 1.62& 27.89 \p 3.71& \textbf{81.02 \p 0.58}\\
         
         & \textbf{Diabetes} & Pregnancies & 82.11 \p 0.49& 64.19 \p 0.72& 59.57 \p 0.68& 36.71 \p 4.56& 17.56 \p 2.32& \textbf{79.53 \p 0.79}\\

         & \textbf{IMDB} & Names & 91.39 \p 1.57& 83.27 \p 2.05 & 72.15 \p 1.92 & 48.36 \p 2.79 & 37.93 \p 2.84 & \textbf{89.15 \p 1.32} \\
         \hline

         \multirow{6}{*}{\textbf{Backdoor}} & \textbf{MNIST} & \multirow{6}{3.5em}{Backdoor Pixel Pattern} & 94.75 \p 4.88& 96.23 \p 0.16& \textbf{96.85 \p 0.91}& 65.31 \p 4.39& 40.52 \p 7.38& 95.83 \p 1.14\\
         
         & \textbf{FMNIST} & & 90.68 \p 2.19& 92.98 \p 0.75& \textbf{93.52 \p 1.63}& 67.62 \p 0.81& 42.24 \p 4.45& 92.61 \p 1.57\\
         
         & \textbf{CIFAR-10} & & 87.55 \p 3.71& 90.92 \p 1.83& \textbf{91.23 \p 0.44}& 53.98 \p 2.17& 27.16 \p 9.68 &89.52 \p 2.18\\
         
         & \textbf{CIFAR-20} & & 74.47 \p 2.38& 81.61 \p 1.75& \textbf{82.52 \p 0.69}& 54.76 \p 0.98& 23.02 \p 3.11& 78.34 \p 2.35\\
         
         & \textbf{CIFAR-100} & & 54.13 \p 7.62& 73.12 \p 1.54& \textbf{73.59 \p 1.66}& 34.30 \p 0.42& 15.21 \p 5.83& 69.30 \p 2.27\\

         & \textbf{ImageNet} & & 52.86 \p 4.14 & 67.18 \p 2.07 & \textbf{67.52 \p 1.69} & 31.17 \p 3.96 & 12.75 \p 5.27& 65.36 \p 1.84\\
         
         \hline

        \multirow{2}{*}{\textbf{Biased}} & \textbf{CMNIST} & Color & 81.72 \p 3.41& 98.49 \p 1.46& 82.54 \p 0.78& 27.56 \p 1.71& 25.05 \p 5.09& \textbf{83.85 \p 1.63}\\
        
        & \textbf{CelebA} & Mouth & 87.35 \p 4.07& 95.87 \p 1.52& 88.93 \p 2.65& 16.98 \p 0.23& 20.19 \p 7.21& \textbf{94.62 \p 2.49}\\
         \hline
    \end{tabular}
    \end{adjustbox}
    \caption{The accuracy of $\mathcal{D}_{t}$ for each unlearning method across different unlearning scenarios.}
    \label{tab: utility}
\end{table}

\subsection{Effectiveness Guarantee}
\label{sec: effectiveness guarantee}

In this subsection, we analyze the unlearning effectiveness of \shortname~against baselines in sensitive, backdoor, and biased feature unlearning scenarios. 

\subsubsection{Sensitive Feature Unlearning}
\label{sec: effectiveness sensitive}


To evaluate \shortname's effectiveness in unlearning sensitive features, we measured feature sensitivity (see \s~\ref{sec: methodology feature sensitivity}) and conducted a model inversion attack (MIA) \cite{model_inversion_attack_1, model_inversion_attack,generative_model_inversion, tabular_model_inversion}.  


\begin{table}[t]
    \centering
    \begin{adjustbox}{max width=\textwidth}
    \begin{tabular}{c |c |c | c|c |c |c | c| c}
        \hline
         \multirow{2}{*}{\textbf{Scenario}}& \multirow{2}{*}{\textbf{Datasets}}& \multirow{2}{3.3em}{\textbf{Unlearn Feature}}& \multicolumn{6}{c}{\textbf{Feature Sensitivity}} \\
         \cline{4-9}
         
         & & & \textbf{Baseline} & \textbf{Retrain} & \textbf{Fine-tune} & \textbf{FedCDP \cite{FU_Discriminative_Pruning}} & \textbf{FedRecovery \cite{FU_FedRecovery}} & \textbf{\shortname~ (Ours)}\\
         \hline
         
         \multirow{4}{3.5em}{\textbf{Sensitive}} & \textbf{CelebA}& Mouth& 0.96 \p 1.41$\times 10^{-2}$& 0.07 \p 8.06$\times 10^{-4}$& 0.79 \p 2.05$\times 10^{-2}$& 0.93 \p 2.87$\times 10^{-2}$& 0.91\p 3.41$\times 10^{-2}$& \textbf{0.09 \p 3.04$\times 10^{-4}$}\\ 
         
         & \textbf{Adult}& Marriage & 1.31 \p 1.53$\times 10^{-2}$& 0.02 \p 6.47$\times 10^{-4}$& 0.94 \p 6.81$\times 10^{-2}$& 1.07 \p 7.43$\times 10^{-2}$& 1.14 \p 2.57$\times 10^{-2}$& \textbf{0.05 \p 1.72$\times 10^{-4}$}\\ 
         
         & \textbf{Diabetes}& Pregnancies & 1.52 \p 0.91$\times 10^{-2}$& 0.05 \p 5.07$\times 10^{-4}$& 0.96 \p 1.28$\times 10^{-2}$& 1.23 \p 3.82$\times 10^{-2}$& 0.83 \p 5.08$\times 10^{-2}$& \textbf{0.07 \p 1.07$\times 10^{-4}$}\\ 

         & \textbf{IMDB}& Names & 0.85 \p 1.07$\times 10^{-2}$& 0.07 \p 5.38$\times 10^{-4}$& 0.74 \p 3.81$\times 10^{-2}$& 0.81 \p 3.27$\times 10^{-2}$& 0.78 \p 2.41$\times 10^{-2}$& \textbf{0.08 \p 1.32$\times 10^{-4}$}\\ 
         \hline
    \end{tabular}
    \end{adjustbox}
    \caption{Feature sensitivity for each unlearning method across sensitive feature unlearning scenario.}
    \label{tab: feature sensitivity}
\end{table}

\paragraph{Feature Sensitivity} \tab~\ref{tab: feature sensitivity} shows the sensitivity of the unlearn feature. The baseline model had high sensitivity to this feature. Similar results were observed for the Fine-tune, FedCDP \cite{FU_Discriminative_Pruning}, and FedRecovery models \cite{FU_FedRecovery}, with sensitivities greater than 0.8, indicating ineffective unlearning. In contrast, our proposed \shortname~model exhibits low sensitivity, similar to the Retrain model, indicating successful unlearning of the sensitive feature.

\begin{table}[t]
    \centering
    \begin{adjustbox}{max width=\textwidth}
    \begin{tabular}{c |c |c | c|c |c |c | c| c}
        \hline
         \multirow{2}{*}{\textbf{Scenario}}& \multirow{2}{*}{\textbf{Datasets}}& \multirow{2}{3.3em}{\textbf{Unlearn Feature}}& \multicolumn{6}{c}{\textbf{Attack Success Rate(ASR) ($\%$)}} \\
         \cline{4-9}
         
         & & & \textbf{Baseline} & \textbf{Retrain} & \textbf{Fine-tune} & \textbf{FedCDP \cite{FU_Discriminative_Pruning}} & \textbf{FedRecovery \cite{FU_FedRecovery}} & \textbf{\shortname~(Ours)}\\
         \hline
         
         \multirow{3}{3.5em}{\textbf{Sensitive}} & \textbf{CelebA}& Mouth & 84.36 \p 3.22& 47.52 \p 1.04& 77.43 \p 10.98& 75.36 \p 9.31& 71.52 \p 6.07& \textbf{51.28 \p 2.41}\\ 
         
         & \textbf{Adult}& Marriage & 87.54 \p 13.89& 49.28 \p 2.13& 83.45 \p 8.44& 72.83 \p 5.18& 80.39 \p 10.68& \textbf{49.58 \p 1.38}\\ 
         
         & \textbf{Diabetes}& Pregnancies & 92.31 \p 7.55& 38.89 \p 2.52& 88.46 \p 5.01& 81.91 \p 8.17& 78.27 \p 2.47 & \textbf{42.61 \p 1.81}\\ 

         & \textbf{IMDB}& Names & 90.28 \p 2.49 & 40.29 \p 1.59 & 86.74 \p 3.81 & 83.67 \p 4.59 & 80.95 \p 3.51 & \textbf{43.75 \p 1.86}\\ 
         \hline
    \end{tabular}
    \end{adjustbox}
    \caption{The ASR of MIA for each unlearning method across sensitive feature unlearning scenario.}
    \label{tab: mia asr}
\end{table}

\paragraph{ASR of MIA} \tab~\ref{tab: mia asr} shows the ASR results. The Baseline model achieved an ASR exceeding 80\%, indicating substantial exposure of sensitive features. Similar observations were made for the Fine-tune, FedCDP \cite{FU_Discriminative_Pruning}, and FedRecovery \cite{FU_FedRecovery} models, with ASR surpassing 70\% exhibiting ineffective feature unlearning. Conversely, \shortname~achieved low ASR, suggesting successful feature unlearning with minimal unlearned feature exposure after using \shortname~via MIA.

\begin{figure}[t]
    \begin{adjustbox}{max width=\textwidth}
        \begin{tabular}{*{8}{c}}
        
             \textbf{Target}& \textbf{Baseline}& \textbf{Retrain}& \textbf{\shortname~(Ours)}& \textbf{Target}& \textbf{Baseline}& \textbf{Retrain}& \textbf{\shortname~(Ours)}\\
            
             \includegraphics[width=0.125\textwidth]{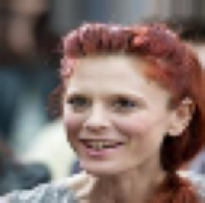}& \includegraphics[width=0.125\textwidth]{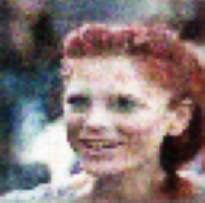}& \includegraphics[width=0.125\textwidth]{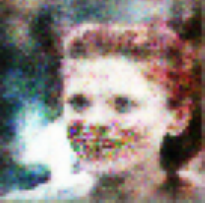}& \includegraphics[width=0.125\textwidth]{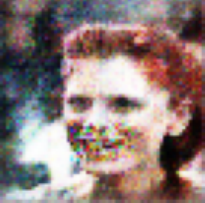}&
            
             \includegraphics[width=0.125\textwidth]{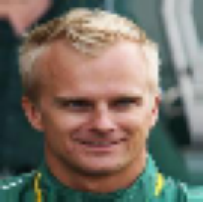}& \includegraphics[width=0.125\textwidth]{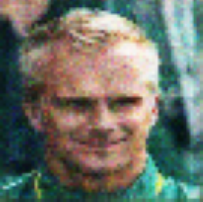}& \includegraphics[width=0.125\textwidth]{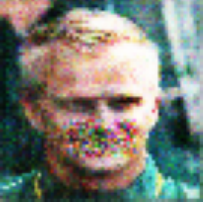}& \includegraphics[width=0.125\textwidth]{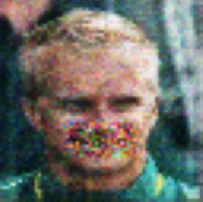}\\
             
        \end{tabular}
    \end{adjustbox}
    \caption{MIA reconstruction on CelebA (unlearned mouth)}
    \label{fig: gmi}
\end{figure}

\paragraph{MIA Reconstruction} \fig~\ref{fig: gmi} shows MIA-reconstructed images. The Baseline model achieved complete reconstruction, whereas both Retrain and \shortname~models failed to reconstruct the mouth feature accurately. This underscores \shortname's effectiveness in unlearning and preserving privacy by preventing precise reconstruction of unlearned features via MIA.

\subsubsection{Backdoor Feature Unlearning}
\label{sec: effectiveness backdoor}

\begin{table}[t]
    \centering
    \begin{adjustbox}{max width=\textwidth}
    \begin{tabular}{c | c | c | c | c | c | c | c | c | c }
        \hline
        \multirow{2}{*}{\textbf{Scenarios}} & \multirow{2}{*}{\textbf{Datasets}} & \multicolumn{2}{c|}{\multirow{2}{*}{\textbf{Unlearn Feature}}} & \multicolumn{6}{c}{\textbf{Accuracy (\%)}} \\
        
        \cline{5-10}
         & & \multicolumn{2}{c|}{} & \textbf{Baseline} & \textbf{Retrain} & \textbf{Fine-tune} & \textbf{FedCDP\cite{FU_Discriminative_Pruning}} & \textbf{FedRecovery\cite{FU_FedRecovery}} & \textbf{\shortname(Ours)} \\
         \hline

         \multirow{12}{*}{\textbf{Backdoor}}& \multirow{2}{*}{\textbf{MNIST}}& \multirow{12}{3.5em}{Backdoor pixel-pattern}& $\mathcal{D}_r$& 95.65 \p 1.39 & 97.19 \p 2.49 & \textbf{96.16 \p 0.37} & 65.82 \p 6.85 & 40.81 \p 4.31 & 95.93 \p 0.45 \\
         & & & $\mathcal{D}_u$& 97.43 \p 3.69 & 0.00 \p 0.00 & 72.64 \p 0.24 & 69.37 \p 0.83 & 53.72 \p 3.14 & \textbf{0.11 \p 0.01}\\
         \cline{4-10}

         & \multirow{2}{*}{\textbf{FMNIST}}& & $\mathcal{D}_r$& 91.07 \p 0.54 & 93.85 \p 1.08 & \textbf{94.36 \p 1.98}& 68.46 \p 3.39 & 42.93 \p 2.50 & 92.83 \p 0.61 \\
         & & & $\mathcal{D}_u$& 94.51 \p 6.29 & 0.00 \p 0.00 & 43.91 \p 0.28 & 72.19 \p 0.49 & 48.15 \p 4.37 & \textbf{0.90 \p 0.03}\\
         \cline{4-10}

         & \multirow{2}{*}{\textbf{CIFAR-10}}& & $\mathcal{D}_r$& 87.63 \p 1.16 & 91.12 \p 1.60 & \textbf{92.02 \p 3.15}& 54.91 \p 6.91 & 27.49 \p 4.96 & 89.91 \p 0.95 \\
         & & & $\mathcal{D}_u$& 95.05 \p 2.30 & 0.00 \p 0.00 & 88.44 \p 0.92 & 62.75 \p 5.07 & 49.26 \p 2.23 & \textbf{0.29 \p 0.04}\\
         \cline{4-10}

         & \multirow{2}{*}{\textbf{CIFAR-20}}& & $\mathcal{D}_r$& 75.06 \p 6.41 & 81.91 \p 4.68 & \textbf{82.67 \p 1.32}& 55.67 \p 6.35 & 23.76 \p 2.17 & 78.29 \p 3.12 \\
         & & & $\mathcal{D}_u$& 94.21 \p 4.11 & 0.00 \p 0.00 & 86.53 \p 1.47 & 50.17 \p 9.11 & 50.38 \p 4.25 & \textbf{0.78 \p 0.08}\\
         \cline{4-10}

         & \multirow{2}{*}{\textbf{CIFAR-100}}& & $\mathcal{D}_r$& 54.14 \p 3.96 & 73.54 \p 5.70 & \textbf{73.66 \p 6.57}& 34.62 \p 2.24 & 15.62 \p 7.78 & 69.57 \p 3.81 \\
         & & & $\mathcal{D}_u$& 88.98 \p 6.63 & 0.00 \p 0.00 & 65.38 \p 4.76 & 57.29 \p 3.62 & 46.17 \p 9.25 & \textbf{0.15 \p 0.01}\\
         \cline{4-10}

         & \multirow{2}{*}{\textbf{ImageNet}}& & $\mathcal{D}_r$&  52.35 \p 2.25 & 67.05 \p 1.29 & \textbf{67.34 \p 2.73} & 29.74 \p 4.72 & 13.46 \p 6.53 & 65.74 \p 1.32\\
         
         & & & $\mathcal{D}_u$& 83.16 \p 3.74 & 0.00 \p 0.00 & 71.48 \p 3.69 & 62.39 \p 3.05 & 54.92 \p 5.59 & \textbf{0.09 \p 0.02}\\

         \hline

         \multirow{4}{*}{\textbf{Biased}}& \multirow{2}{*}{\textbf{CMNIST}}& \multirow{2}{*}{Color}& $\mathcal{D}_r$& 64.94 \p 7.88 & 98.76 \p 3.65 & 67.15 \p 2.60 & 25.85 \p 1.58 & 23.92 \p 1.08 & \textbf{84.31 \p 2.63}\\
         & & & $\mathcal{D}_u$& 98.88 \p 4.90 & 98.44 \p 1.90 & 97.95 \p 1.13 & 30.17 \p 4.69 & 27.64 \p 9.37 & \textbf{84.62 \p 3.59}\\
         \cline{4-10}

         & \multirow{2}{*}{\textbf{CelebA}}& \multirow{2}{*}{Mouth}& $\mathcal{D}_r$& 79.46 \p 2.09 & 96.47 \p 6.15 & 84.45 \p 1.48 & 14.29 \p 0.81 & 16.34 \p 3.43 & \textbf{94.18 \p 3.08}\\
         & & & $\mathcal{D}_u$& 96.38 \p 3.87 & 96.11 \p 2.17 & 94.23 \p 0.66 & 21.58 \p 3.48 & 25.72 \p 8.02 & \textbf{94.79 \p 1.48}\\

         \hline
         
    \end{tabular}
    \end{adjustbox}
    \caption{The accuracy of $\mathcal{D}_r$ and $\mathcal{D}_u$ for each unlearning method across different unlearning scenarios.}
    \label{tab: effectiveness}
\end{table}

\paragraph{Accuracy} $\mathcal{D}_r$ and $\mathcal{D}_u$ represent the clean and backdoor datasets, respectively. Successful unlearning is shown by low $Acc_u$ and high $Acc_r$, indicating effective unlearning and preserved model utility. As shown in \tab~\ref{tab: effectiveness}, the Fine-tune method has higher $Acc_r$ and utility than the Retrain method but lower unlearning effectiveness due to high $Acc_u$. FedCDP \cite{FU_Discriminative_Pruning} and FedRecovery \cite{FU_FedRecovery} show low utility and unlearning effectiveness with low $Acc_r$ and $Acc_u$, rendering them unsuitable for backdoor feature unlearning. In contrast, \shortname~demonstrates the highest utility and unlearning effectiveness.

\paragraph{Attention Map} \fig~\ref{fig: backdoor gradcam} illustrates attention maps analyzing backdoor feature unlearning. Initially, the Baseline model focuses on the $5\times5$ square at the top-left corner, indicating a significant influence on output prediction by the pixel-pattern backdoor feature. In contrast, \shortname~unlearned models shift the attention towards recognizable objects like digits and cars, similar to the Retrain model. This shift suggests a reduced sensitivity to the backdoor feature, indicating a successful unlearning. See \appen~\ref{sec: appendix backdoor gradcam} for supplementary results.

\subsubsection{Biased Feature Unlearning}
\label{sec: effectiveness biased}

\begin{figure}[t]
    \centering
    \begin{subfigure}{0.59\textwidth}
        \begin{adjustbox}{max width=\textwidth}
            \begin{tabular}{*{6}{c}}
                 
                 \textbf{Input} & \includegraphics[width=0.2\textwidth]{Figures/results/backdoor_gradcam/mnist/input/1} &
                 \includegraphics[width=0.2\textwidth]{Figures/results/backdoor_gradcam/fmnist/input/9} & 
                 \includegraphics[width=0.2\textwidth]{Figures/results/backdoor_gradcam/cifar10/input/9} & 
                 \includegraphics[width=0.2\textwidth]{Figures/results/backdoor_gradcam/cifar20/input/1} & 
                 \includegraphics[width=0.2\textwidth]{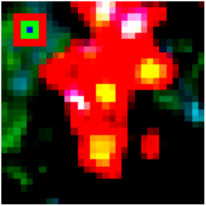}\\

                \textbf{Baseline} & \includegraphics[width=0.2\textwidth]{Figures/results/backdoor_gradcam/mnist/backdoor/1} &
                \includegraphics[width=0.2\textwidth]{Figures/results/backdoor_gradcam/fmnist/backdoor/9} & \includegraphics[width=0.2\textwidth]{Figures/results/backdoor_gradcam/cifar10/backdoor/9} & \includegraphics[width=0.2\textwidth]{Figures/results/backdoor_gradcam/cifar20/backdoor/1} & \includegraphics[width=0.2\textwidth]{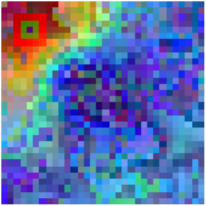} \\

                \textbf{Retrain} & \includegraphics[width=0.2\textwidth]{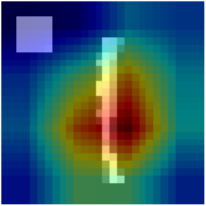} &
                \includegraphics[width=0.2\textwidth]{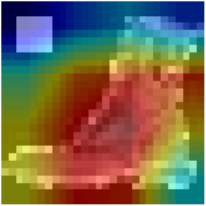} & \includegraphics[width=0.2\textwidth]{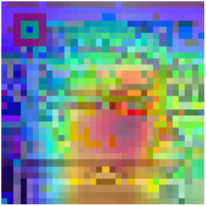} & \includegraphics[width=0.2\textwidth]{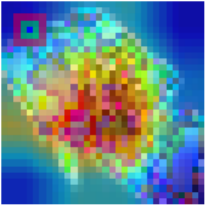} & \includegraphics[width=0.2\textwidth]{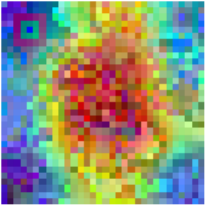} \\

                \textbf{\shortname} & \includegraphics[width=0.2\textwidth]{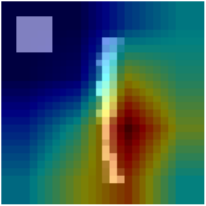} &
                \includegraphics[width=0.2\textwidth]{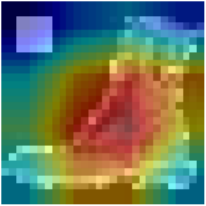} & \includegraphics[width=0.2\textwidth]{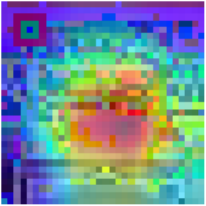} & \includegraphics[width=0.2\textwidth]{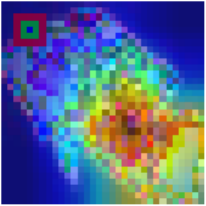} & \includegraphics[width=0.2\textwidth]{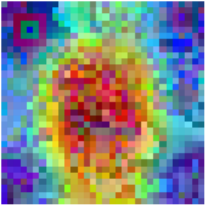} \\

                & \textbf{MNIST}& \textbf{FMNIST}& \textbf{CIFAR-10}& \textbf{CIFAR-20}& \textbf{CIFAR-100}\\
            \end{tabular}
        \end{adjustbox}
        \caption{Backdoor Feature Unlearning}
        \label{fig: backdoor gradcam}
    \end{subfigure}
    \begin{subfigure}{0.4\textwidth}
        \begin{adjustbox}{max width=\textwidth}
            \begin{tabular}{*{4}{c}}

             \includegraphics[width=0.325\textwidth]{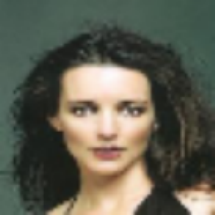} &
             \includegraphics[width=0.325\textwidth]{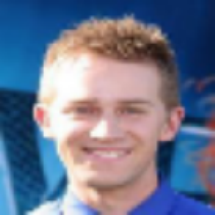} & \includegraphics[width=0.325\textwidth]{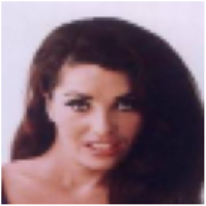} & \includegraphics[width=0.325\textwidth]{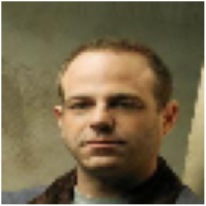} \\ 

             \includegraphics[width=0.325\textwidth]{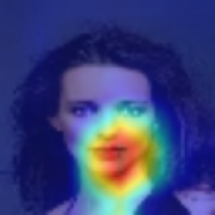} &
             \includegraphics[width=0.325\textwidth]{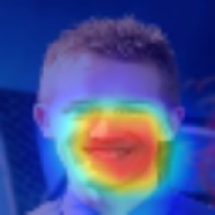} & \includegraphics[width=0.325\textwidth]{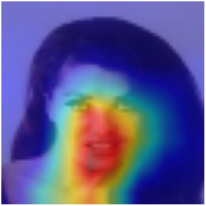} & \includegraphics[width=0.325\textwidth]{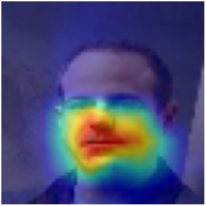} \\

             \includegraphics[width=0.325\textwidth]{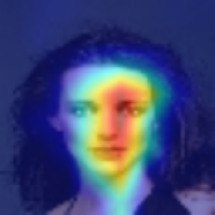} &
             \includegraphics[width=0.325\textwidth]{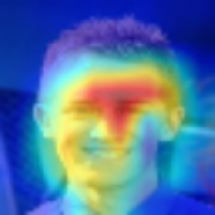} & \includegraphics[width=0.325\textwidth]{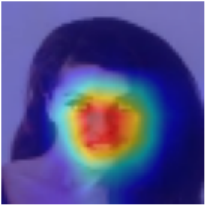} & \includegraphics[width=0.325\textwidth]{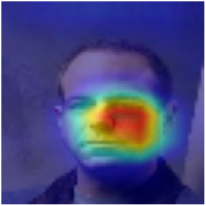} \\

             \includegraphics[width=0.325\textwidth]{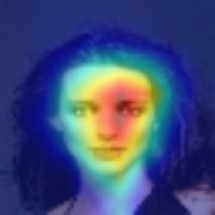} &
             \includegraphics[width=0.325\textwidth]{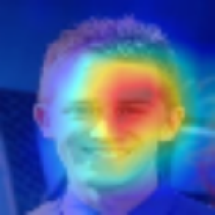} & \includegraphics[width=0.325\textwidth]{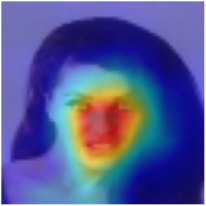} & \includegraphics[width=0.325\textwidth]{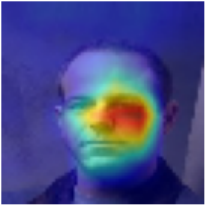} \\

             \multicolumn{2}{c}{\textbf{Bias Dataset}}& \multicolumn{2}{c}{\textbf{Unbias Dataset}}\\
             
            \end{tabular}
        \end{adjustbox}
        \caption{Biased Feature Unlearning}
        \label{fig: bias gradcam}
    \end{subfigure}
    \caption{The attention map of each unlearning method across different unlearning scenarios.}
    \label{fig: attention map}
\end{figure}

\paragraph{Accuracy} $\mathcal{D}_r$ and $\mathcal{D}_u$ represent the unbias and bias datasets, respectively. Successful unlearning results in similar accuracies across both datasets ($Acc_r \approx Acc_u$), ensuring fairness while maintaining high $Acc_r$ and $Acc_u$ for utility. \tab~\ref{tab: effectiveness} shows that the Fine-tune method fails to unlearn bias, as $Acc_u$ remains higher than $Acc_r$, despite slightly higher $Acc_r$ compared to Retrain. FedCDP \cite{FU_Discriminative_Pruning} and FedRecovery \cite{FU_FedRecovery} exhibit catastrophic forgetting, with low $Acc_r$ and $Acc_u$, making them unsuitable for biased feature unlearning. In contrast, \shortname~demonstrates effective unlearning with similar $Acc_r$ and $Acc_u$, and maintains high overall accuracy, indicating a successful biased feature unlearning.


\paragraph{Attention Map} \fig~\ref{fig: bias gradcam} shows attention maps analyzing biased feature unlearning. The Baseline model predominantly focuses on the biased feature region (mouth) in both bias and unbias datasets, suggesting its significant impact on output prediction. However, \shortname~unlearned models redistribute attention across various facial regions in both datasets, similar to the Retrain model. This shift indicates reduced sensitivity to the biased feature, demonstrating successful unlearning. See \appen~\ref{sec: appendix bias gradcam} for supplementary results.

\subsection{Computational Complexity}
\label{sec: time efficiency}

\begin{figure}[t]
    \centering
    \begin{subfigure}[b]{0.45\textwidth}
        \centering
        \includegraphics[width=\textwidth]{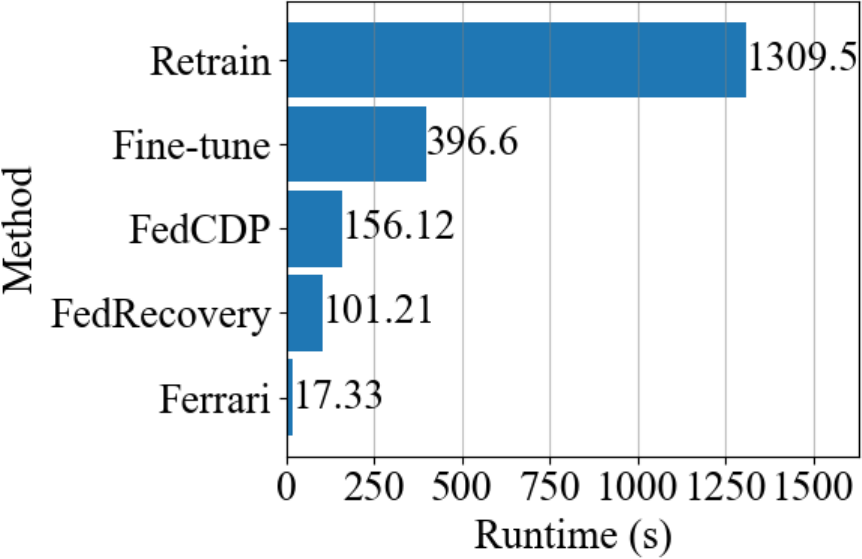} 
        \caption{Runtime(s)}
        \label{fig: runtime}
    \end{subfigure}
    \hfill
    \begin{subfigure}[b]{0.45\textwidth}
        \centering
        \includegraphics[width=\textwidth]{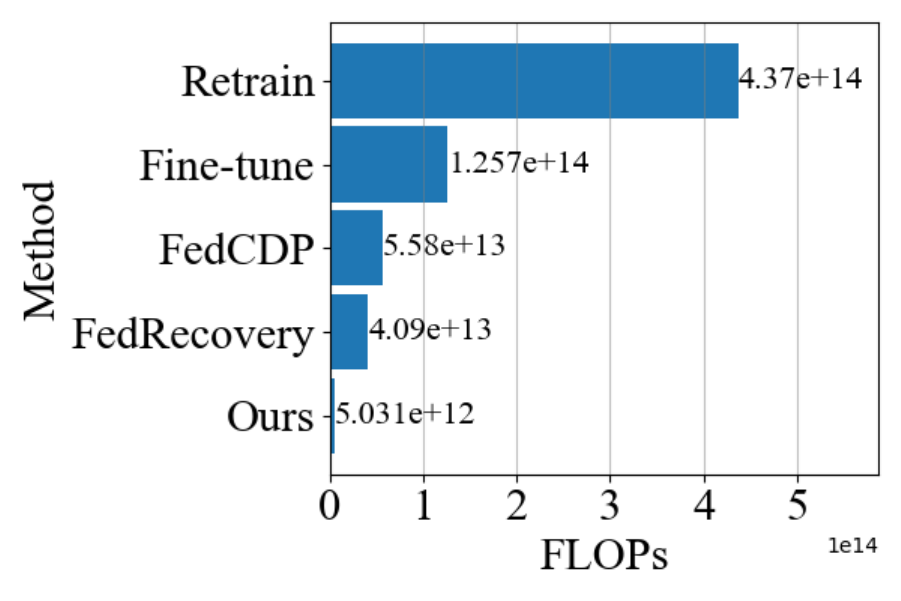} 
        \caption{FLOPs}
        \label{fig: flops}
    \end{subfigure}
    \caption{Computational complexity analysis comparing the runtime(s) and FLOPs for each unlearning method.}
    \label{fig: time efficiency}
\end{figure}

In \fig~\ref{fig: time efficiency}, we evaluate the runtime performance and FLOPs metrics of each unlearning method to demonstrate the computational complexity. The Retrain method  is expected to have the slowest runtime and highest FLOPs, while Fine-tune is fast but still slower than other methods.

Both FedCDP \cite{FU_Discriminative_Pruning} and FedRecovery \cite{FU_FedRecovery} demonstrate faster runtimes and lower FLOPs than the Fine-tune method, but they are still more computationally expensive than \shortname. This is primarily due to the need to access training datasets from all clients and the computational expense of gradient residual calculations \cite{FU_FedRecovery}.


In contrast, \shortname~has the lowest computational complexity, with the fastest runtime and lowest FLOPs. It only requires access to the local dataset of the unlearn client and achieves feature unlearning by minimizing feature sensitivity within a single epoch. 

\subsection{Ablation Study and Hyper-parameter Analysis}
\label{sec: ablation studies}

\begin{figure}[t]
    \centering
    \begin{subfigure}[b]{0.325\textwidth}
        \centering
        \includegraphics[width=\textwidth]{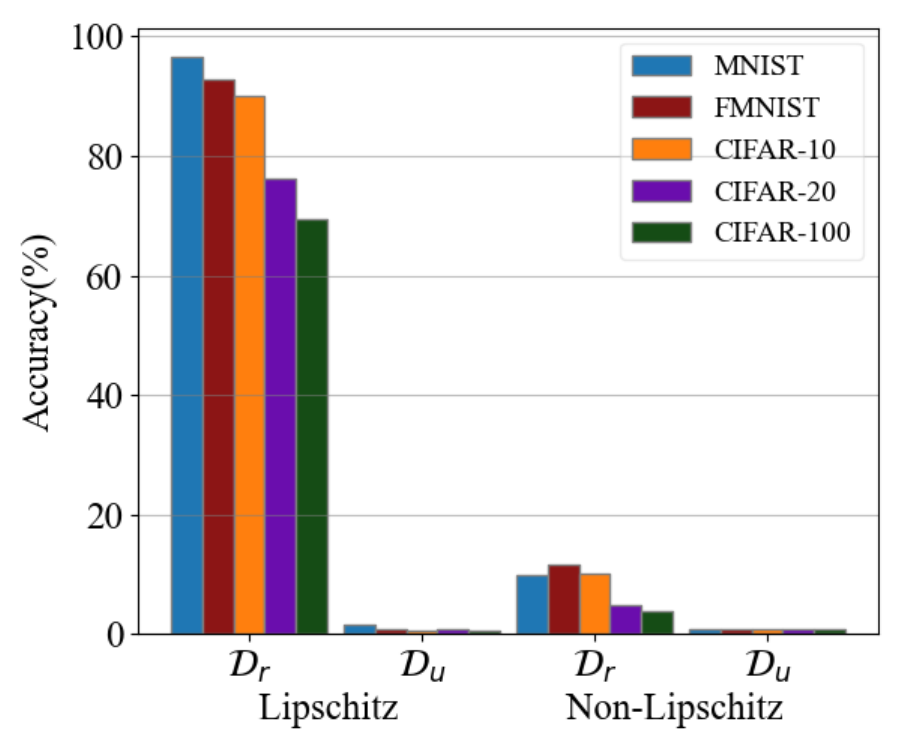} 
        \caption{Non-Lipschitz}
        \label{fig: non-lipschitz experiment}
    \end{subfigure}
    \hfill
    \begin{subfigure}[b]{0.325\textwidth}
        \centering
        \includegraphics[width=\textwidth]{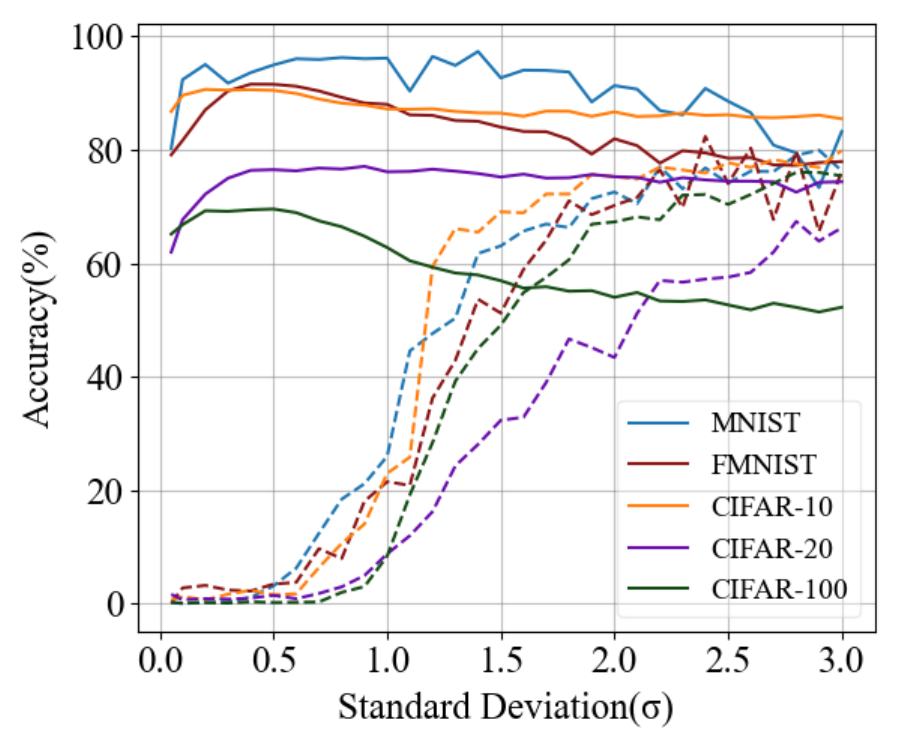} 
        \caption{Gaussian noise level($\sigma$)}
        \label{fig: sigma experiment}
    \end{subfigure}
    \hfill
    \begin{subfigure}[b]{0.325\textwidth}
        \centering
        \includegraphics[width=\textwidth]{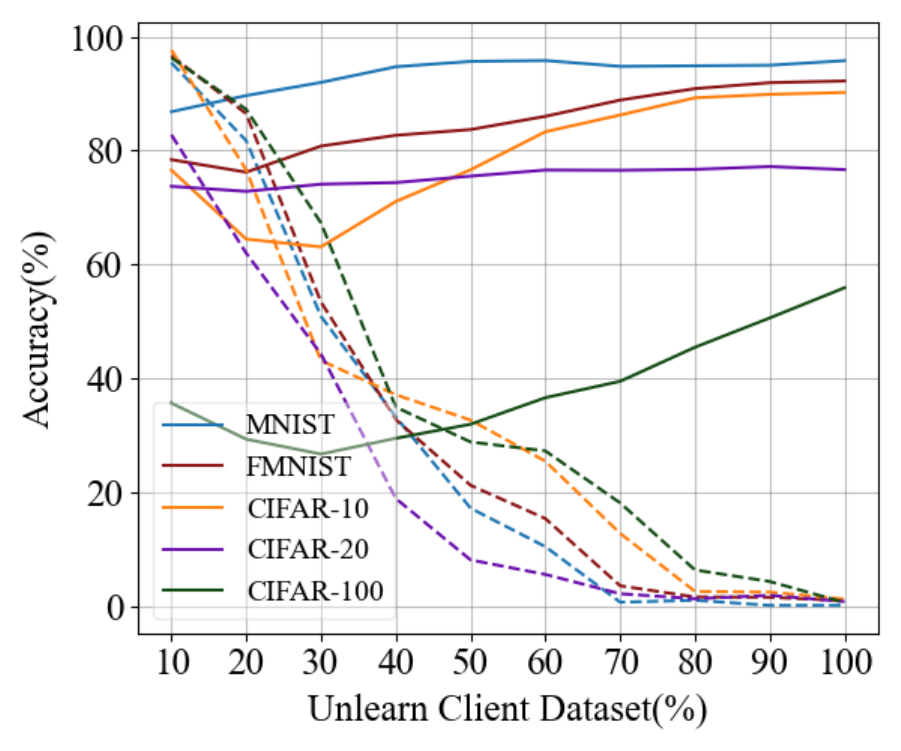} 
        \caption{Number of $\mathcal{D}_u$}
        \label{fig: dataset number experiment}
    \end{subfigure}
    \caption{Ablation and hyper-parameter analysis on \shortname~backdoor feature unlearning. Solid line: $\mathcal{D}_r$; dashed line: $\mathcal{D}_u$.}
    \label{fig: ablation study}
\end{figure}

We conduct an ablation study to analyze how Non-Lipschitz affects the effectiveness of our proposed \shortname~and hyper-parameter analysis of Gaussian noise level ($\sigma$) and number of $\mathcal{D}_u$ in \fig~\ref{fig: ablation study}.

\paragraph{Non-Lipschitz} We evaluate the unlearning performance by removing the denominator in \eq~\ref{eq:loss}, calling this the Non-Lipschitz method, as shown in \fig~\ref{fig: non-lipschitz experiment}. The results indicate catastrophic forgetting: $\mathcal{D}_r$ accuracy drops below 10\%, and the unlearned model misclassifies all inputs into a single random class, rendering it useless. This stems from the unbounded loss function in the non-Lipschitz method, unlike the bounded Lipschitz constant in \eq~\ref{eq:loss}, which provides a theoretical guarantee (see \s~\ref{subsec:verification}). Refer to \appen~\ref{sec: appendix loss analysis} for a detailed analysis of Lipschitz and Non-Lipschitz loss functions.



\paragraph{Gaussian Noise}  The effectiveness of \shortname~is significantly influenced by injected Gaussian noise. \fig~\ref{fig: sigma experiment} shows the accuracy of $\mathcal{D}_r$ and $\mathcal{D}_u$ across different $\sigma$ levels. In the $0.05 \leq \sigma \leq 1.0$ range, $\mathcal{D}_r$ accuracy stays high and $\mathcal{D}_u$ accuracy remains low, indicating a balance. Thus, we implement $\sigma$ values between 0.05 and 1.0 for a balanced accuracy across $\mathcal{D}_r$ and $\mathcal{D}_u$.

\paragraph{Number of Unlearn Dataset} Our analysis illustrated in \fig~\ref{fig: dataset number experiment}, demonstrates that \shortname~remains effective with partial $\mathcal{D}_u$ from $C_u$ for feature unlearning (\ie~data lost). Using 70\% of $\mathcal{D}_u$ yields comparable accuracy to using the full (\ie~100\%) dataset, highlighting the method's flexibility even with partial data.
\section{Conclusion}
\label{sec: conclusion}

This paper introduces \shortname, a federated feature unlearning framework designed to efficiently remove sensitive, backdoor, and biased features without extensive retraining. Leveraging Lipschitz continuity, \shortname~reduces model sensitivity to specific features, ensuring robust and fair models. Uniquely, it requires participation only from the client requesting unlearning, preserving privacy and practicality in FL environments. Experimental results and theoretical analysis demonstrate Ferrari's effectiveness across various data domains, addressing the crucial need for feature-level unlearning in federated learning. This method can serve as a technical solution to meet regulatory requirements for data deletion while maintaining model performance, offering significant value to clients by securing their \textquotedblleft right to be forgotten\textquotedblright~and preventing potential privacy leakage.

\subsection{Limitation and Future Work}
\label{sec: appendix limitation}

The proposed federated feature unlearning method works effectively using only the unlearning client's local data, making it well-suited for real-world scenarios. However, for optimal results, access to the full dataset is required. As demonstrated in Section~\ref{sec: ablation studies}, using 70\% of the data yields comparable performance, but significant data reduction diminishes effectiveness. Future research should focus on developing methods that require only a small portion of the client's data and expanding the approach beyond classification models to include for example, generative models. Additionally, enhancements such as advanced perturbation techniques and integration with privacy-preserving methods should be explored.
\section*{Acknowledgement}
This research is supported by the Fundamental Research Grant Scheme (FRGS/1/2024/ICT02/UM/01/1), awarded by the Ministry of Higher Education, Malaysia.

\bibliographystyle{ieeetr}
\bibliography{main}

\begin{thebibliography}{10}

\bibitem{konevcny2015federated}
J.~Kone{\v{c}}n{\`y}, B.~McMahan, and D.~Ramage, ``Federated optimization: Distributed optimization beyond the datacenter,'' {\em arXiv preprint arXiv:1511.03575}, 2015.

\bibitem{mcmahan2017communication}
B.~McMahan, E.~Moore, D.~Ramage, S.~Hampson, and B.~A. y~Arcas, ``Communication-efficient learning of deep networks from decentralized data,'' in {\em Artificial Intelligence and Statistics}, pp.~1273--1282, PMLR, 2017.

\bibitem{yang2019federated}
Q.~Yang, Y.~Liu, T.~Chen, and Y.~Tong, ``Federated machine learning: Concept and applications,'' {\em ACM Transactions on Intelligent Systems and Technology (TIST)}, vol.~10, no.~2, pp.~1--19, 2019.

\bibitem{harding2019understanding}
E.~L. Harding, J.~J. Vanto, R.~Clark, L.~Hannah~Ji, and S.~C. Ainsworth, ``Understanding the scope and impact of the california consumer privacy act of 2018,'' {\em Journal of Data Protection \& Privacy}, vol.~2, no.~3, pp.~234--253, 2019.

\bibitem{che2023fast}
T.~Che, Y.~Zhou, Z.~Zhang, L.~Lyu, J.~Liu, D.~Yan, D.~Dou, and J.~Huan, ``Fast federated machine unlearning with nonlinear functional theory,'' in {\em International conference on machine learning}, pp.~4241--4268, PMLR, 2023.

\bibitem{FU_Challenges}
Z.~Liu, Y.~Jiang, J.~Shen, M.~Peng, K.-Y. Lam, X.~Yuan, and X.~Liu, ``A survey on federated unlearning: Challenges, methods, and future directions,'' 2024.

\bibitem{FU_Survey}
N.~Romandini, A.~Mora, C.~Mazzocca, R.~Montanari, and P.~Bellavista, ``Federated unlearning: A survey on methods, design guidelines, and evaluation metrics,'' 2024.

\bibitem{FU_Taxonomy}
J.~Yang and Y.~Zhao, ``A survey of federated unlearning: A taxonomy, challenges and future directions,'' 2023.

\bibitem{MU_Features_Labels}
A.~Warnecke, L.~Pirch, C.~Wressnegger, and K.~Rieck, ``Machine unlearning of features and labels,'' in {\em Proc. of the 30th Network and Distributed System Security (NDSS)}, 2023.

\bibitem{MU_Attribute_Unlearning}
T.~Guo, S.~Guo, J.~Zhang, W.~Xu, and J.~Wang, ``Efficient attribute unlearning: Towards selective removal of input attributes from feature representations,'' 2022.

\bibitem{model_inversion_attack_1}
M.~Fredrikson, E.~Lantz, S.~Jha, S.~M. Lin, D.~Page, and T.~Ristenpart, ``Privacy in pharmacogenetics: An end-to-end case study of personalized warfarin dosing,'' {\em Proceedings of the USENIX Security Symposium. UNIX Security Symposium}, vol.~2014, pp.~17--32, 2014.

\bibitem{model_inversion_attack}
M.~Fredrikson, S.~Jha, and T.~Ristenpart, ``Model inversion attacks that exploit confidence information and basic countermeasures,'' in {\em Proceedings of the 22nd ACM SIGSAC conference on computer and communications security}, pp.~1322--1333, 2015.

\bibitem{generative_model_inversion}
Y.~Zhang, R.~Jia, H.~Pei, W.~Wang, B.~Li, and D.~Song, ``The secret revealer: Generative model-inversion attacks against deep neural networks,'' in {\em 2020 IEEE/CVF Conference on Computer Vision and Pattern Recognition (CVPR)}, pp.~250--258, 2020.

\bibitem{tabular_model_inversion}
S.~Mehnaz, S.~V. Dibbo, E.~Kabir, N.~Li, and E.~Bertino, ``Are your sensitive attributes private? novel model inversion attribute inference attacks on classification models,'' in {\em USENIX Security Symposium}, 2022.

\bibitem{backdoor_fl}
E.~Bagdasaryan, A.~Veit, Y.~Hua, D.~Estrin, and V.~Shmatikov, ``How to backdoor federated learning,'' in {\em Proceedings of the Twenty Third International Conference on Artificial Intelligence and Statistics} (S.~Chiappa and R.~Calandra, eds.), vol.~108 of {\em Proceedings of Machine Learning Research}, pp.~2938--2948, PMLR, 26--28 Aug 2020.

\bibitem{backdoor_fl_review}
T.~D. Nguyen, T.~Nguyen, P.~L. Nguyen, H.~H. Pham, K.~D. Doan, and K.-S. Wong, ``Backdoor attacks and defenses in federated learning: Survey, challenges and future research directions,'' {\em Engineering Applications of Artificial Intelligence}, vol.~127, p.~107166, 2024.

\bibitem{backdoor_fl_li}
H.~Li, C.~Wu, S.~Zhu, and Z.~Zheng, ``Learning to backdoor federated learning,'' in {\em ICLR 2023 Workshop on Backdoor Attacks and Defenses in Machine Learning}, 2023.

\bibitem{badnets}
T.~Gu, K.~Liu, B.~Dolan-Gavitt, and S.~Garg, ``Badnets: Evaluating backdooring attacks on deep neural networks,'' {\em IEEE Access}, vol.~7, pp.~47230--47244, 2019.

\bibitem{bias_review}
D.~Pessach and E.~Shmueli, ``A review on fairness in machine learning,'' {\em ACM Comput. Surv.}, vol.~55, feb 2022.

\bibitem{bias_survey}
N.~Mehrabi, F.~Morstatter, N.~Saxena, K.~Lerman, and A.~Galstyan, ``A survey on bias and fairness in machine learning,'' {\em ACM Comput. Surv.}, vol.~54, jul 2021.

\bibitem{bias_sagawa}
S.~Sagawa*, P.~W. Koh*, T.~B. Hashimoto, and P.~Liang, ``Distributionally robust neural networks,'' in {\em International Conference on Learning Representations}, 2020.

\bibitem{bias_seo}
S.~Seo, J.-Y. Lee, and B.~Han, ``Unsupervised learning of debiased representations with pseudo-attributes,'' in {\em 2022 IEEE/CVF Conference on Computer Vision and Pattern Recognition (CVPR)}, pp.~16721--16730, 2022.

\bibitem{machine_unlearning2015}
Y.~Cao and J.~Yang, ``Towards making systems forget with machine unlearning,'' in {\em 2015 IEEE symposium on security and privacy}, pp.~463--480, IEEE, 2015.

\bibitem{machineunlearningprivacy2021}
M.~Chen, Z.~Zhang, T.~Wang, M.~Backes, M.~Humbert, and Y.~Zhang, ``When machine unlearning jeopardizes privacy,'' in {\em Proceedings of the 2021 ACM SIGSAC Conference on Computer and Communications Security}, CCS '21, (New York, NY, USA), p.~896–911, Association for Computing Machinery, 2021.

\bibitem{eurocrypt-2020-30234}
S.~Garg, S.~Goldwasser, and P.~N. Vasudevan, ``Formalizing data deletion in the context of the right to be forgotten,'' in {\em 39th Annual International Conference on the Theory and Applications of Cryptographic Techniques, Zagreb, Croatia, May 10–14, 2020, Proceedings}, vol.~12105 of {\em Lecture Notes in Computer Science}, Springer, 2020.

\bibitem{Thudi2021OnTN}
A.~Thudi, H.~Jia, I.~Shumailov, and N.~Papernot, ``On the necessity of auditable algorithmic definitions for machine unlearning,'' in {\em USENIX Security Symposium}, 2021.

\bibitem{nguyen2022}
T.~T. Nguyen, T.~T. Huynh, P.~L. Nguyen, A.~W.-C. Liew, H.~Yin, and Q.~V.~H. Nguyen, ``A survey of machine unlearning,'' 2022.

\bibitem{SISA2021}
L.~Bourtoule, V.~Chandrasekaran, C.~A. Choquette-Choo, H.~Jia, A.~Travers, B.~Zhang, D.~Lie, and N.~Papernot, ``Machine unlearning,'' in {\em 2021 IEEE Symposium on Security and Privacy (SP)}, pp.~141--159, 2021.

\bibitem{ARCANE2022}
H.~Yan, X.~Li, Z.~Guo, H.~Li, F.~Li, and X.~Lin, ``Arcane: An efficient architecture for exact machine unlearning,'' in {\em Proceedings of the Thirty-First International Joint Conference on Artificial Intelligence, {IJCAI-22}} (L.~D. Raedt, ed.), pp.~4006--4013, International Joint Conferences on Artificial Intelligence Organization, 7 2022.
\newblock Main Track.

\bibitem{Jie2023}
J.~Xu, Z.~Wu, C.~Wang, and X.~Jia, ``Machine unlearning: Solutions and challenges,'' {\em ArXiv}, vol.~abs/2308.07061, 2023.

\bibitem{Xu2023}
H.~Xu, T.~Zhu, L.~Zhang, W.~Zhou, and P.~S. Yu, ``Machine unlearning: A survey,'' {\em ACM Comput. Surv.}, vol.~56, aug 2023.

\bibitem{AmnesiacUnlearning}
L.~Graves, V.~Nagisetty, and V.~Ganesh, ``Amnesiac machine learning,'' in {\em AAAI Conference on Artificial Intelligence}, 2020.

\bibitem{Two-stageModelRetraining}
J.~Kim and S.~S. Woo, ``Efficient two-stage model retraining for machine unlearning,'' in {\em 2022 IEEE/CVF Conference on Computer Vision and Pattern Recognition Workshops (CVPRW)}, pp.~4360--4368, 2022.

\bibitem{UNDO}
S.~Lee and S.~S. Woo, ``Undo: Effective and accurate unlearning method for deep neural networks,'' in {\em Proceedings of the 32nd ACM International Conference on Information and Knowledge Management}, pp.~4043--4047, 2023.

\bibitem{BoundaryUnlearning}
M.~Chen, W.~Gao, G.~Liu, K.~Peng, and C.~Wang, ``Boundary unlearning: Rapid forgetting of deep networks via shifting the decision boundary,'' in {\em 2023 IEEE/CVF Conference on Computer Vision and Pattern Recognition (CVPR)}, pp.~7766--7775, 2023.

\bibitem{LSF}
T.~Shibata, G.~Irie, D.~Ikami, and Y.~Mitsuzumi, ``Learning with selective forgetting,'' in {\em International Joint Conference on Artificial Intelligence}, 2021.

\bibitem{huang2021unlearnable}
H.~Huang, X.~Ma, S.~M. Erfani, J.~Bailey, and Y.~Wang, ``Unlearnable examples: Making personal data unexploitable,'' in {\em International Conference on Learning Representations}, 2021.

\bibitem{di2022hidden}
J.~Z. Di, J.~Douglas, J.~Acharya, G.~Kamath, and A.~Sekhari, ``Hidden poison: Machine unlearning enables camouflaged poisoning attacks,'' in {\em Workshop on Trustworthy and Socially Responsible Machine Learning, NeurIPS 2022}, 2022.

\bibitem{UNSIR}
A.~K. Tarun, V.~S. Chundawat, M.~Mandal, and M.~Kankanhalli, ``Fast yet effective machine unlearning,'' {\em IEEE Transactions on Neural Networks and Learning Systems}, pp.~1--10, 2023.

\bibitem{Chundawat_Tarun_Mandal_Kankanhalli_2023}
V.~S. Chundawat, A.~K. Tarun, M.~Mandal, and M.~Kankanhalli, ``Can bad teaching induce forgetting? unlearning in deep networks using an incompetent teacher,'' {\em Proceedings of the AAAI Conference on Artificial Intelligence}, vol.~37, pp.~7210--7217, Jun. 2023.

\bibitem{StochasticTeacher}
X.~Zhang, J.~Wang, N.~Cheng, Y.~Sun, C.~Zhang, and J.~Xiao, ``Machine unlearning methodology based on stochastic teacher network,'' in {\em Advanced Data Mining and Applications} (X.~Yang, H.~Suhartanto, G.~Wang, B.~Wang, J.~Jiang, B.~Li, H.~Zhu, and N.~Cui, eds.), (Cham), pp.~250--261, Springer Nature Switzerland, 2023.

\bibitem{kurmanji2023towards}
M.~Kurmanji, P.~Triantafillou, J.~Hayes, and E.~Triantafillou, ``Towards unbounded machine unlearning,'' in {\em Thirty-seventh Conference on Neural Information Processing Systems}, 2023.

\bibitem{jung2024attack}
Y.~Jung, I.~Cho, S.-H. Hsu, and J.~Hockenmaier, ``Attack and reset for unlearning: Exploiting adversarial noise toward machine unlearning through parameter re-initialization,'' 2024.

\bibitem{Hoang_2024_WACV}
T.~Hoang, S.~Rana, S.~Gupta, and S.~Venkatesh, ``Learn to unlearn for deep neural networks: Minimizing unlearning interference with gradient projection,'' in {\em Proceedings of the IEEE/CVF Winter Conference on Applications of Computer Vision (WACV)}, pp.~4819--4828, January 2024.

\bibitem{abbasi2023covarnav}
A.~Abbasi, C.~Thrash, E.~Akbari, D.~Zhang, and S.~Kolouri, ``Covarnav: Machine unlearning via model inversion and covariance navigation,'' 2023.

\bibitem{choi2023machine}
D.~Choi and D.~Na, ``Towards machine unlearning benchmarks: Forgetting the personal identities in facial recognition systems,'' 2023.

\bibitem{goel2023adversarial}
S.~Goel, A.~Prabhu, A.~Sanyal, S.-N. Lim, P.~Torr, and P.~Kumaraguru, ``Towards adversarial evaluations for inexact machine unlearning,'' 2023.

\bibitem{FisherForgetting}
A.~Golatkar, A.~Achille, and S.~Soatto, ``Eternal sunshine of the spotless net: Selective forgetting in deep networks,'' in {\em 2020 IEEE/CVF Conference on Computer Vision and Pattern Recognition (CVPR)}, pp.~9301--9309, 2020.

\bibitem{NTK}
A.~Golatkar, A.~Achille, and S.~Soatto, ``Forgetting outside the box: Scrubbing deep networks of information accessible from input-output observations,'' in {\em Computer Vision -- ECCV 2020} (A.~Vedaldi, H.~Bischof, T.~Brox, and J.-M. Frahm, eds.), (Cham), pp.~383--398, Springer International Publishing, 2020.

\bibitem{Mixed-PrivacyForgetting}
A.~Golatkar, A.~Achille, A.~Ravichandran, M.~Polito, and S.~Soatto, ``Mixed-privacy forgetting in deep networks,'' in {\em 2021 IEEE/CVF Conference on Computer Vision and Pattern Recognition (CVPR)}, pp.~792--801, 2021.

\bibitem{CertifiedDataRemoval}
C.~Guo, T.~Goldstein, A.~Hannun, and L.~Van Der~Maaten, ``Certified data removal from machine learning models,'' in {\em Proceedings of the 37th International Conference on Machine Learning} (H.~D. III and A.~Singh, eds.), vol.~119 of {\em Proceedings of Machine Learning Research}, pp.~3832--3842, PMLR, 13--18 Jul 2020.

\bibitem{foster2023fast}
J.~Foster, S.~Schoepf, and A.~Brintrup, ``Fast machine unlearning without retraining through selective synaptic dampening,'' 2023.

\bibitem{Guo_Certified_Data_Removal}
C.~Guo, T.~Goldstein, A.~Hannun, and L.~Van Der~Maaten, ``Certified data removal from machine learning models,'' in {\em Proceedings of the 37th International Conference on Machine Learning} (H.~D. III and A.~Singh, eds.), vol.~119 of {\em Proceedings of Machine Learning Research}, pp.~3832--3842, PMLR, 13--18 Jul 2020.

\bibitem{FU_Knowledge_Distillation}
C.~Wu, S.~Zhu, and P.~Mitra, ``Federated unlearning with knowledge distillation,'' 2022.

\bibitem{FU_FedEraser}
G.~Liu, X.~Ma, Y.~Yang, C.~Wang, and J.~Liu, ``Federaser: Enabling efficient client-level data removal from federated learning models,'' in {\em 2021 IEEE/ACM 29th International Symposium on Quality of Service (IWQOS)}, pp.~1--10, 2021.

\bibitem{FU_on-Device_Reccomendation}
W.~Yuan, H.~Yin, F.~Wu, S.~Zhang, T.~He, and H.~Wang, ``Federated unlearning for on-device recommendation,'' in {\em Proceedings of the Sixteenth ACM International Conference on Web Search and Data Mining}, WSDM '23, (New York, NY, USA), p.~393–401, Association for Computing Machinery, 2023.

\bibitem{FU_FedRecover}
X.~Cao, J.~Jia, Z.~Zhang, and N.~Z. Gong, ``Fedrecover: Recovering from poisoning attacks in federated learning using historical information,'' in {\em 2023 IEEE Symposium on Security and Privacy (SP)}, pp.~1366--1383, 2023.

\bibitem{FU_Verifi}
X.~Gao, X.~Ma, J.~Wang, Y.~Sun, B.~Li, S.~Ji, P.~Cheng, and J.~Chen, ``Verifi: Towards verifiable federated unlearning,'' 2022.

\bibitem{FU_HDUS}
G.~Ye, T.~Chen, Q.~V. Hung~Nguyen, and H.~Yin, ``Heterogeneous decentralised machine unlearning with seed model distillation,'' {\em CAAI Transactions on Intelligence Technology}, 2024.

\bibitem{FU_Asynchoronous}
N.~Su and B.~Li, ``Asynchronous federated unlearning,'' in {\em IEEE INFOCOM 2023 - IEEE Conference on Computer Communications}, pp.~1--10, 2023.

\bibitem{FU_FedRecovery}
L.~Zhang, T.~Zhu, H.~Zhang, P.~Xiong, and W.~Zhou, ``Fedrecovery: Differentially private machine unlearning for federated learning frameworks,'' {\em IEEE Transactions on Information Forensics and Security}, vol.~18, pp.~4732--4746, 2023.

\bibitem{FU_Efficiently_Erase_a_Client}
A.~Halimi, S.~Kadhe, A.~Rawat, and N.~Baracaldo, ``Federated unlearning: How to efficiently erase a client in fl?,'' 2023.

\bibitem{FU_Subspace}
G.~Li, L.~Shen, Y.~Sun, Y.~Hu, H.~Hu, and D.~Tao, ``Subspace based federated unlearning,'' 2023.

\bibitem{FU_Erasing_Backdoors}
M.~Alam, H.~Lamri, and M.~Maniatakos, ``Get rid of your trail: Remotely erasing backdoors in federated learning,'' 2023.

\bibitem{FU_Discriminative_Pruning}
J.~Wang, S.~Guo, X.~Xie, and H.~Qi, ``Federated unlearning via class-discriminative pruning,'' in {\em Proceedings of the ACM Web Conference 2022}, WWW '22, (New York, NY, USA), p.~622–632, Association for Computing Machinery, 2022.

\bibitem{FU_Momentum_Degradation}
Y.~Zhao, P.~Wang, H.~Qi, J.~Huang, Z.~Wei, and Q.~Zhang, ``Federated unlearning with momentum degradation,'' {\em IEEE Internet of Things Journal}, pp.~1--1, 2023.

\bibitem{FU_TheRightToBeForgotten}
Y.~Liu, L.~Xu, X.~Yuan, C.~Wang, and B.~Li, ``The right to be forgotten in federated learning: An efficient realization with rapid retraining,'' in {\em IEEE INFOCOM 2022 - IEEE Conference on Computer Communications}, pp.~1749--1758, 2022.

\bibitem{FU_QuickDrop}
A.~Dhasade, Y.~Ding, S.~Guo, A.~marie Kermarrec, M.~D. Vos, and L.~Wu, ``Quickdrop: Efficient federated unlearning by integrated dataset distillation,'' 2023.

\bibitem{FU_FedFilter}
P.~Wang, Z.~Yan, M.~S. Obaidat, Z.~Yuan, L.~Yang, J.~Zhang, Z.~Wei, and Q.~Zhang, ``Edge caching with federated unlearning for low-latency v2x communications,'' {\em IEEE Communications Magazine}, pp.~1--7, 2023.

\bibitem{FU_Fedme2}
H.~Xia, S.~Xu, J.~Pei, R.~Zhang, Z.~Yu, W.~Zou, L.~Wang, and C.~Liu, ``Fedme2: Memory evaluation \& erase promoting federated unlearning in dtmn,'' {\em IEEE Journal on Selected Areas in Communications}, vol.~41, no.~11, pp.~3573--3588, 2023.

\bibitem{MU_GAN1}
S.~Moon, S.~Cho, and D.~Kim, ``Feature unlearning for pre-trained gans and vaes,'' in {\em AAAI Conference on Artificial Intelligence}, 2023.

\bibitem{MU_GAN3}
S.~Bae, S.~Kim, H.~Jung, and W.~Lim, ``Gradient surgery for one-shot unlearning on generative model,'' 2023.

\bibitem{MU_GAN4}
Anonymous, ``Machine unlearning for image-to-image generative models,'' in {\em The Twelfth International Conference on Learning Representations}, 2024.

\bibitem{MU_GAN5}
W.~Wang, H.~Bai, J.~tse Huang, Y.~Wan, Y.~Yuan, H.~Qiu, N.~Peng, and M.~R. Lyu, ``New job, new gender? measuring the social bias in image generation models,'' 2024.

\bibitem{MU_LLM1}
Y.~Yao, X.~Xu, and Y.~Liu, ``Large language model unlearning,'' 2023.

\bibitem{MU_LLM2}
C.~Yu, S.~Jeoung, A.~Kasi, P.~Yu, and H.~Ji, ``Unlearning bias in language models by partitioning gradients,'' in {\em Findings of the Association for Computational Linguistics: ACL 2023} (A.~Rogers, J.~Boyd-Graber, and N.~Okazaki, eds.), (Toronto, Canada), pp.~6032--6048, Association for Computational Linguistics, July 2023.

\bibitem{MU_LLM3}
J.~Chen and D.~Yang, ``Unlearn what you want to forget: Efficient unlearning for {LLM}s,'' in {\em Proceedings of the 2023 Conference on Empirical Methods in Natural Language Processing} (H.~Bouamor, J.~Pino, and K.~Bali, eds.), (Singapore), pp.~12041--12052, Association for Computational Linguistics, Dec. 2023.

\bibitem{SpectralNorm}
Y.~Yoshida and T.~Miyato, ``Spectral norm regularization for improving the generalizability of deep learning,'' {\em arXiv preprint arXiv:1705.10941}, 2017.

\bibitem{LipschitzRegularization}
A.~M. Oberman and J.~Calder, ``Lipschitz regularized deep neural networks converge and generalize,'' {\em CoRR}, vol.~abs/1808.09540, 2018.

\bibitem{LipschitzContinuityFundementalAspect}
G.~Khromov and S.~P. Singh, ``Some fundemental aspects about lipschitz continuity of neural networks,'' 2023.

\bibitem{usama2018robustnn}
M.~Usama and D.~E. Chang, ``Towards robust neural networks with lipschitz continuity,'' in {\em Digital Forensics and Watermarking} (C.~D. Yoo, Y.-Q. Shi, H.~J. Kim, A.~Piva, and G.~Kim, eds.), (Cham), pp.~373--389, Springer International Publishing, 2019.

\bibitem{weng2018evaluating}
T.-W. Weng, H.~Zhang, P.-Y. Chen, J.~Yi, D.~Su, Y.~Gao, C.-J. Hsieh, and L.~Daniel, ``Evaluating the robustness of neural networks: An extreme value theory approach,'' in {\em International Conference on Learning Representations}, 2018.

\bibitem{yoshida2017spectral}
Y.~Yoshida and T.~Miyato, ``Spectral norm regularization for improving the generalizability of deep learning,'' 2017.

\bibitem{Celeba}
Z.~Liu, P.~Luo, X.~Wang, and X.~Tang, ``Deep learning face attributes in the wild,'' in {\em 2015 IEEE International Conference on Computer Vision (ICCV)}, pp.~3730--3738, 2015.

\bibitem{adult_income_dataset}
Wenruliu, ``Adult income dataset.'' Kaggle, 2024.

\bibitem{diabetes_dataset}
M.~Akturk, ``Diabetes dataset.'' Kaggle, 2024.

\bibitem{bias_dje}
Y.~Djebrouni, N.~Benarba, O.~Touat, P.~De~Rosa, S.~Bouchenak, A.~Bonifati, P.~Felber, V.~Marangozova, and V.~Schiavoni, ``Bias mitigation in federated learning for edge computing,'' {\em Proc. ACM Interact. Mob. Wearable Ubiquitous Technol.}, vol.~7, jan 2024.

\bibitem{bias_chen}
R.~Chen, J.~Yang, H.~Xiong, J.~Bai, T.~Hu, J.~Hao, Y.~FENG, J.~T. Zhou, J.~Wu, and Z.~Liu, ``Fast model debias with machine unlearning,'' in {\em Thirty-seventh Conference on Neural Information Processing Systems}, 2023.

\bibitem{Mnist}
Y.~Lecun, L.~Bottou, Y.~Bengio, and P.~Haffner, ``Gradient-based learning applied to document recognition,'' {\em Proceedings of the IEEE}, vol.~86, no.~11, pp.~2278--2324, 1998.

\bibitem{Resnet18}
K.~He, X.~Zhang, S.~Ren, and J.~Sun, ``Deep residual learning for image recognition,'' in {\em 2016 IEEE Conference on Computer Vision and Pattern Recognition (CVPR)}, pp.~770--778, 2016.

\bibitem{fmnist}
H.~Xiao, K.~Rasul, and R.~Vollgraf, ``Fashion-mnist: a novel image dataset for benchmarking machine learning algorithms,'' {\em arXiv preprint arXiv:1708.07747}, 2017.

\bibitem{Cifar10}
A.~Krizhevsky, ``Learning multiple layers of features from tiny images,'' Toronto, ON, Canada, 2009.

\bibitem{imagenet}
J.~Deng, W.~Dong, R.~Socher, L.-J. Li, K.~Li, and L.~Fei-Fei, ``Imagenet: A large-scale hierarchical image database,'' in {\em 2009 IEEE Conference on Computer Vision and Pattern Recognition}, pp.~248--255, 2009.

\bibitem{bert}
J.~Devlin, M.-W. Chang, K.~Lee, and K.~Toutanova, ``Bert: Pre-training of deep bidirectional transformers for language understanding,'' in {\em North American Chapter of the Association for Computational Linguistics}, 2019.

\bibitem{imdb}
A.~L. Maas, R.~E. Daly, P.~T. Pham, D.~Huang, A.~Y. Ng, and C.~Potts, ``Learning word vectors for sentiment analysis,'' in {\em Proceedings of the 49th Annual Meeting of the Association for Computational Linguistics: Human Language Technologies} (D.~Lin, Y.~Matsumoto, and R.~Mihalcea, eds.), (Portland, Oregon, USA), pp.~142--150, Association for Computational Linguistics, June 2011.

\bibitem{gradcam}
R.~R. Selvaraju, M.~Cogswell, A.~Das, R.~Vedantam, D.~Parikh, and D.~Batra, ``Grad-cam: Visual explanations from deep networks via gradient-based localization,'' in {\em 2017 IEEE International Conference on Computer Vision (ICCV)}, pp.~618--626, 2017.

\end{thebibliography}

\newpage
\appendix
\section{Appendix}
\label{sec: appendix}

\subsection{Proof of Theorem 1} 
\label{sec: appendix theorem 1 proof}
As illustrated in \s~\ref{subsec:challenge}, it is hard to build the unlearned data $x^u$ for the feature unlearning since adding the perturbation may influence the model accuracy seriously. Suppose the feature is successfully removed when the norm of perturbation is larger than $C$. We define the  utility loss $\ell_{1}$ with unlearning feature successfully:
\begin{equation}
    \ell_{1} = \min_{\|\delta_\calF\| \geq C}\EE_{(x,y)\in \calD} \min_\theta \ell \big( f_\theta(x+\delta_\calF), y \big ) 
\end{equation}
And we define the maximum utility loss with the norm perturbation less than $C$ as:
\begin{equation}
    \ell_{2} = \max_{\|\delta_\calF\| \leq C}\EE_{(x,y)\in \calD} \min_\theta \ell \big( f_\theta(x+\delta_\calF), y \big ) 
\end{equation}

\begin{assumption}\label{assum:1-app}
Assume $\ell_2 \leq \ell_1$
\end{assumption}
Assumption \ref{assum:1-app} elucidates that the utility loss associated with a perturbation norm less than $C $ is smaller than the utility loss when the perturbation norm is greater than $C$. This assumption is logical, as larger perturbations would naturally lead to greater utility loss.

\begin{assumption} \label{assum:2-app}
Suppose the federated model achieves zero training loss.
\end{assumption}
We have the following theorem to elucidate the relation between feature sensitivity removing via \alg~\ref{alg: federated feature unlearning pseudocode} and exact unlearning (see proof in Appendix).

\begin{theorem} \label{theo:thm1-app}
If Assumption \ref{assum:1-app} and \ref{assum:2-app} hold, the utility loss of unlearned model obtained by \alg~\ref{alg: federated feature unlearning pseudocode} is less than the utility loss with unlearning successfully, \ie,
\begin{equation}
    \ell_u \leq \ell_1,
\end{equation}
where $\ell_u =\EE_{(x,y)\in \calD}\big(\ell (f_{\theta^u}(x), y)$
\end{theorem}

\begin{proof}
When the unlearning happens during the federated training, the unlearning clients would also optimize the training loss and feature sensitivity simultaneously. Specifically, the optimization process could be written as:

\begin{equation*}
    \theta_u = \argmin_\theta \EE_{(x,y)\in \calD}\big(\ell (f_\theta(x), y) +  \lambda \EE_{\delta_\calF}\frac{\|f_\theta(x) - f_\theta(x + \delta_\calF)\|_2}{\|\delta_\calF\|_2} \big ),
\end{equation*}

where $\lambda \geq \frac{1}{C}$ is one coefficient.  
Without loss of generality, we assume the $\ell(f_\theta(x),y) = \|f_\theta(x)-y) \|$. Denote

\begin{equation*}
    \Theta^* = \argmin_\theta \EE_{(x,y)\in \calD} \ell (f_\theta(x), y).
\end{equation*} 

If Assumption \ref{assum:2-app} holds, then  $f_{\theta^*}(x) = y $ for any $\theta^* \in \Theta^*$. Therefore, for any $\|\delta_\calF\| \geq \frac{1}{\lambda}$ such that 

\begin{equation}
\begin{split}
    &\EE_{(x,y)\in \calD} \big(\ell (f_{\theta^*}(x), y) +  \lambda \EE_{\|\delta_\calF\| \geq \frac{1}{\lambda}}\frac{\|f_\theta(x) - f_{\theta^*}(x + \delta_\calF)\|_2}{\|\delta_\calF\|_2} \big) \\
    &=\lambda \EE_{(x,y)\in \calD} \EE_{\|\delta_\calF\| \geq \frac{1}{\lambda}}\frac{\|y - f_{\theta^*}(x + \delta_\calF)\|_2}{\|\delta_\calF\|_2} \\
    & \leq \EE_{(x,y)\in \calD}\EE_{\|\delta_\calF\| \geq \frac{1}{\lambda}}\|y - f_{\theta^*}(x + \delta_\calF)\|_2.
    \end{split}
\end{equation}

Extending Assumption \ref{assum:2-app} to the case of non-zero training loss and assuming it holds, $f_{\theta^*}(x) = y $ for any $\theta^* \in \Theta^*$. Therefore, for any $\|\delta_\calF\| \geq \lambda$, such that

\begin{equation}
\begin{split}
    &\EE_{(x,y)\in \calD} \big(\|(f_{\theta^*}(x)-y\|+  \lambda \EE_{\|\delta_\calF\| \geq {\lambda}}\frac{\|f_\theta(x) - f_{\theta^*}(x + \delta_\calF)\|_2}{\|\delta_\calF\|_2} \big) \\
    &= \EE_{(x,y)\in \calD} \ell \|f_{\theta^*}(x)- y\|+ \lambda \EE_{(x,y)\in \calD} \EE_{\|\delta_\calF\| \geq {\lambda}}\frac{\|f_\theta(x) - f_{\theta^*}(x + \delta_\calF)\|_2}{\|\delta_\calF\|_2} \\
    & \leq \EE_{(x,y)\in \calD} \ell \|f_{\theta^*}(x) - y\|+\EE_{(x,y)\in \calD}\EE_{\|\delta_\calF\| \geq {\lambda}}\|f_\theta(x) - f_{\theta^*}(x + \delta_\calF)\|_2 \\
    & \leq \EE_{(x,y)\in \calD}\EE_{\|\delta_\calF\| \geq {\lambda}}\|y - f_{\theta^*}(x + \delta_\calF)\|_2
    \end{split}
    \label{eq:non zero training loss}
\end{equation}

 Therefore, we further obtain:
\begin{equation}
    \begin{split}
        \ell_u &\leq\min_{\theta \in \RR^d}\EE_{(x,y)\in \calD}  \big(\ell (f_\theta(x), y) + \lambda \EE_{\|\delta_\calF\| \geq \frac{1}{\lambda}}\frac{\|f_\theta(x) - f_\theta(x + \delta_\calF)\|_2}{\|\delta_\calF\|_2} \big )\\
        &\leq \min_{\theta \in \Theta^*} \EE_{(x,y)\in \calD} \big(\ell (f_\theta(x), y) +  \lambda \EE_{\|\delta_\calF\| \geq \frac{1}{\lambda}}\frac{\|f_\theta(x) - f_\theta(x + \delta_\calF)\|_2}{\|\delta_\calF\|_2} \big ) \\
        & \leq \min_{\theta \in \Theta^*} \EE_{(x,y)\in \calD} \EE_{\|\delta_\calF\| \geq \frac{1}{\lambda}}\|y - f_{\theta^*}(x + \delta_\calF)\|_2 \\
         & \leq \EE_{(x,y)\in \calD}  \EE_{\|\delta_\calF\| \geq \frac{1}{\lambda}}\min_{\theta \in \Theta^*} \|y - f_{\theta^*}(x + \delta_\calF)\|_2 \\
        &=  \EE_{\|\delta_\calF\| \geq \frac{1}{\lambda}}\EE_{(x,y)\in \calD}\min_{\theta \in \Theta^*}\|y - f_{\theta^*}(x + \delta_\calF)\|_2 \\
       & \leq  \max_{\|\delta_\calF\| \geq \frac{1}{\lambda}}\EE_{(x,y)\in \calD}\min_{\theta \in \RR^d}\|y - f_{\theta^*}(x + \delta_\calF)\|_2 \\
       & \leq  \max_{\|\delta_\calF \|\leq C}\EE_{(x,y)\in \calD}\min_{\theta \in \RR^d}\|y - f_{\theta^*}(x + \delta_\calF)\|_2 \\
       & =\ell_2,
    \end{split}
\end{equation}
where the last inequality is due to $\lambda\geq \frac{1}{C}$. According to Assumption \ref{assum:1-app}, we have $\ell_u \leq \ell_1$
\end{proof}

\subsection{Experimental Setup}
\label{sec: appendix experimental setup}

\paragraph{Datasets} \textit{MNIST} \cite{Mnist}: Both the \textit{MNIST} \cite{Mnist} and \textit{Fashion-MNIST(FMNIST)} \cite{fmnist} datasets contain images of handwritten digits and attire, respectively. Each dataset comprises 60,000 training examples and 10,000 test examples. In both datasets, each example is represented as a single-channel image with dimensions of 28$\times$28 pixels, categorized into one of 10 classes. Additionally, the \textit{Colored-MNIST(CMNIST)} \cite{Mnist} dataset, an extension of the original MNIST, introduces color into the digits of each example. Consequently, images in the Colored MNIST dataset are represented in three channels. \textit{CIFAR} \cite{Cifar10}: The \textit{CIFAR-10} \cite{Cifar10} dataset comprises 60,000 images, each with dimensions of 32$\times$32 pixels and three color channels, distributed across 10 classes. This dataset includes 6,000 images per class and is partitioned into 50,000 training examples and 10,000 test examples. Similarly, the \textit{CIFAR-100} \cite{Cifar10} dataset shares the same image dimensions and structure as \textit{CIFAR-10} but extends to 100 classes, with each class containing 600 images. Within each class, there are 500 training images and 100 test images. Moreover, \textit{CIFAR-100} organizes its 100 classes into 20 superclasses, forming the \textit{CIFAR-20 dataset} \cite{Cifar10}. \textit{CelebA} \cite{Celeba}: A face recognition dataset featuring 40 attributes such as gender and facial characteristics, comprising 162,770 training examples and 19,962 test examples. This study will focus on utilizing the \textit{CelebA} \cite{Celeba} dataset primarily for gender classification tasks. ImageNet \cite{imagenet}: A large-scale image dataset which contains 1.2 million training samples across 1,000 categories.

\textit{Adult Census Income (Adult)} \cite{adult_income_dataset} includes 48, 842 records with 14 attributes such as age, gender, education, marital status, etc. The classification task of this dataset is to predict if a person earns over $\$50$K a year based on the census attributes. We then consider marital status as the sensitive feature that aim to unlearn in this study. \textit{Diabetes} \cite{diabetes_dataset} includes 768 personal health records of females at least 21 years old with 8 attributes such as blood pressure, insulin level, age and etc. The classification task of this dataset is to predict if a person has diabetes. We then consider number of pregnancies as the sensitive feature that aim to unlearn in this study.

The IMDB movie reviews dataset \cite{imdb} is widely used for binary sentiment analysis, where the task is to determine whether a review expresses a positive or negative sentiment. It comprises 50,000 movie reviews, each labeled as either positive or negative. In this study, we focus on unlearning the influence of specific sensitive features, particularly the names of celebrities. Each client’s local dataset includes names of specific celebrities, which are treated as sensitive features for this analysis.

\paragraph{Baselines}The baseline methods in this study: 

\textit{Baseline}: Original model before unlearning. 

\textit{Retrain}: In scenarios involving sensitive feature unlearning, the retrained model was simply trained using a dataset where Gaussian noise was applied to the unlearned feature region. This approach may lead to performance deterioration, as discussed in \s~\ref{subsec:challenge}. For backdoor feature unlearning scenarios, the retrained model was trained using the retain dataset $\mathcal{D}_r$, also referred to as the clean dataset. In biased feature unlearning scenarios, the retrained model was trained using a combination of 50\% from each of the retain dataset $\mathcal{D}_r$ (bias dataset) and the unlearn client local dataset $\mathcal{D}_u$ (unbias dataset). This ensures fairness in the model's performance across both datasets.

\textit{Fine-tune}: The baseline model is fine-tuned using the retained dataset $\mathcal{D}_r$ for 5 epochs. 

\textit{Class-Discriminative Pruning(FedCDP)} \cite{FU_Discriminative_Pruning}: A FU framework that achieves class unlearning by utilizing Term Frequency-Inverse Document Frequency (TF-IDF) guided channel pruning, which selectively removes the most discriminative channels related to the target category and followed by fine-tuning without retraining from scratch. 

\textit{FedRecovery} \cite{FU_FedRecovery}: A FU framework that achieves client unlearning by removing the influence of a client’s data from the global model using a differentially private machine unlearning algorithm that leverages historical gradient submissions without the need for retraining.

\subsection{Attention Map}
\label{sec: appendix attention map}

In this section, we provide additional results from attention map analysis based on GradCAM \cite{gradcam} for backdoor feature unlearning (refer to \s~\ref{sec: appendix backdoor gradcam}) and biased feature unlearning (refer to \s~\ref{sec: appendix bias gradcam})

\subsubsection{Backdoor Feature Unlearning}
\label{sec: appendix backdoor gradcam}

\begin{figure}[t]
    \centering
    \begin{adjustbox}{max width=\textwidth}
\begin{tabular}{c c c c c c c c c c}
            
                \textbf{Label}& 1& 2& 3& 4& 5& 6& 7& 8& 9\\
                
                \textbf{Input}& \includegraphics[width=0.09\textwidth]{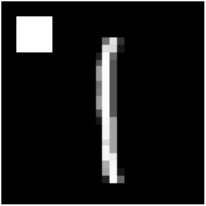}& \includegraphics[width=0.09\textwidth]{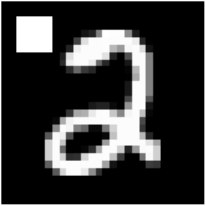}& \includegraphics[width=0.09\textwidth]{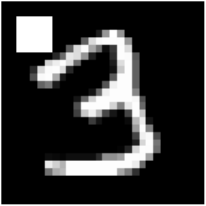}& \includegraphics[width=0.09\textwidth]{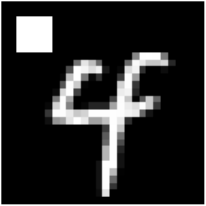}& \includegraphics[width=0.09\textwidth]{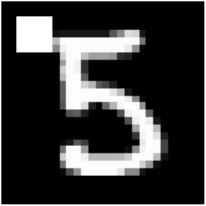}& \includegraphics[width=0.09\textwidth]{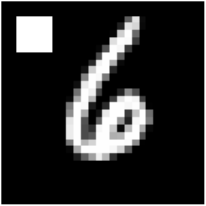}& \includegraphics[width=0.09\textwidth]{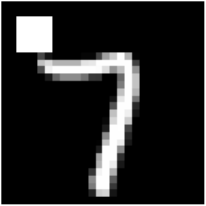}& \includegraphics[width=0.09\textwidth]{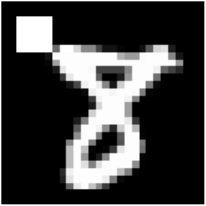}& \includegraphics[width=0.09\textwidth]{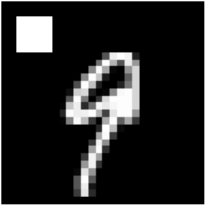}\\

                \textbf{Baseline}& \includegraphics[width=0.09\textwidth]{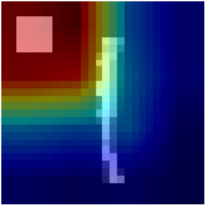}& \includegraphics[width=0.09\textwidth]{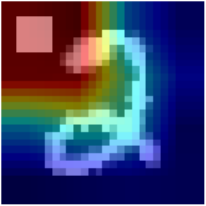}& \includegraphics[width=0.09\textwidth]{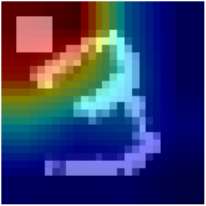}& \includegraphics[width=0.09\textwidth]{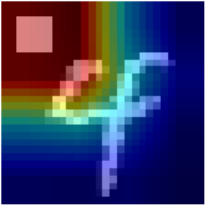}& \includegraphics[width=0.09\textwidth]{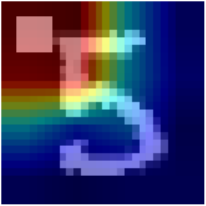}& \includegraphics[width=0.09\textwidth]{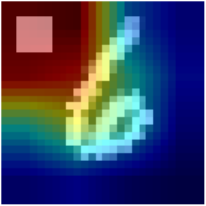}& \includegraphics[width=0.09\textwidth]{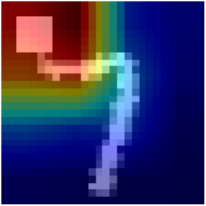}& \includegraphics[width=0.09\textwidth]{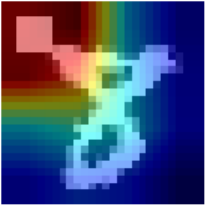}& \includegraphics[width=0.09\textwidth]{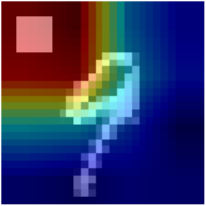}\\
\end{tabular}
        \end{adjustbox}
        \caption{MNIST}
    \label{fig: appendix mnist gradcam}
\end{figure}

\begin{figure}[H]
    \centering
    \begin{adjustbox}{max width=\textwidth}
            \begin{tabular}{c c c c c c c c c c}
                \textbf{Label}& Trouser& Pullover& Dress& Coat& Sandal& Shirt& Sneaker& Bag& Boot\\
                
                \textbf{Input}& \includegraphics[width=0.09\textwidth]{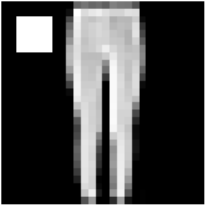}& \includegraphics[width=0.09\textwidth]{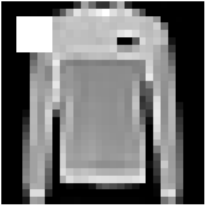}& \includegraphics[width=0.09\textwidth]{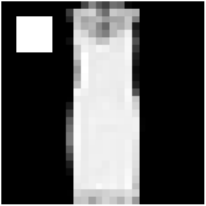}& \includegraphics[width=0.09\textwidth]{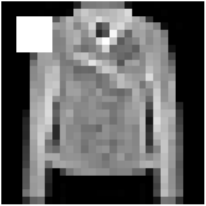}& \includegraphics[width=0.09\textwidth]{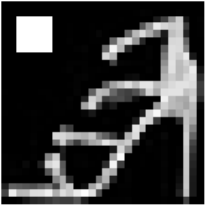}& \includegraphics[width=0.09\textwidth]{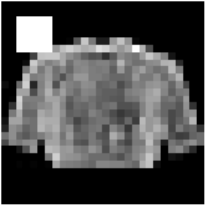}& \includegraphics[width=0.09\textwidth]{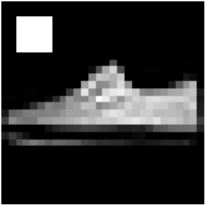}& \includegraphics[width=0.09\textwidth]{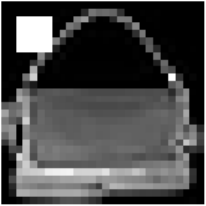}& \includegraphics[width=0.09\textwidth]{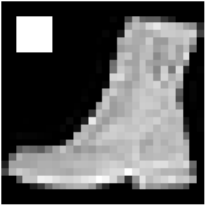}\\

                \textbf{Baseline}& \includegraphics[width=0.09\textwidth]{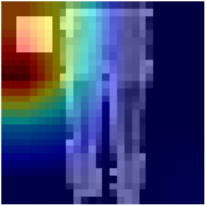}& \includegraphics[width=0.09\textwidth]{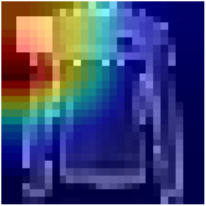}& \includegraphics[width=0.09\textwidth]{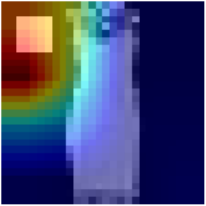}& \includegraphics[width=0.09\textwidth]{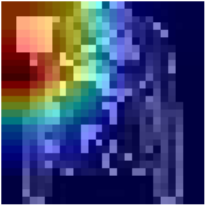}& \includegraphics[width=0.09\textwidth]{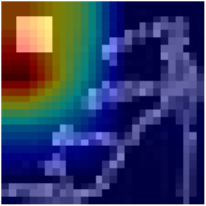}& \includegraphics[width=0.09\textwidth]{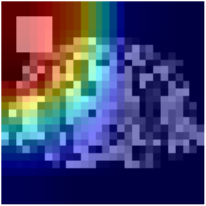}& \includegraphics[width=0.09\textwidth]{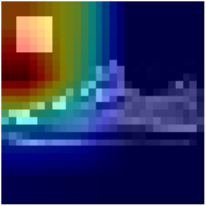}& \includegraphics[width=0.09\textwidth]{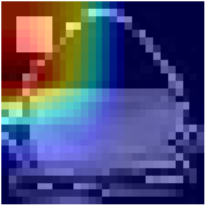}& \includegraphics[width=0.09\textwidth]{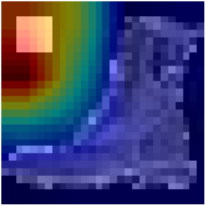}\\
\end{tabular}
        \end{adjustbox}
        \caption{FMNIST}
    \label{fig: appendix fmnist gradcam}
\end{figure}

\begin{figure}[H]
    \centering
    \begin{adjustbox}{max width=\textwidth}
            \begin{tabular}{c c c c c c c c c c}
                \textbf{Label}& Car& Bird& Cat& Deer& Dog& Frog& Horse& Ship& Truck\\

                \textbf{Input}& \includegraphics[width=0.09\textwidth]{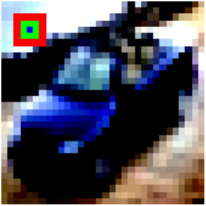}& \includegraphics[width=0.09\textwidth]{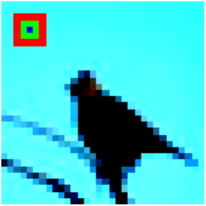}& \includegraphics[width=0.09\textwidth]{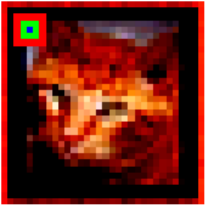}& \includegraphics[width=0.09\textwidth]{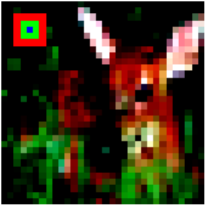}& \includegraphics[width=0.09\textwidth]{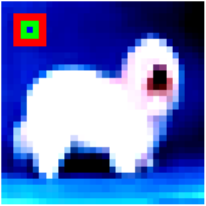}& \includegraphics[width=0.09\textwidth]{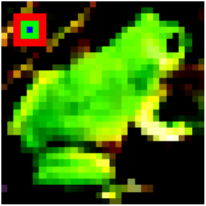}& \includegraphics[width=0.09\textwidth]{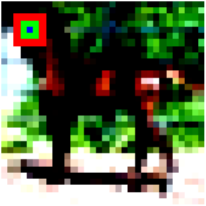}& \includegraphics[width=0.09\textwidth]{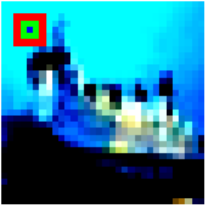}& \includegraphics[width=0.09\textwidth]{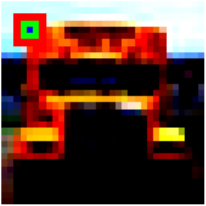}\\

                \textbf{Baseline}& \includegraphics[width=0.09\textwidth]{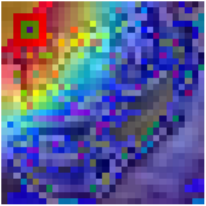}& \includegraphics[width=0.09\textwidth]{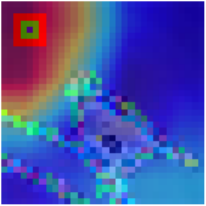}& \includegraphics[width=0.09\textwidth]{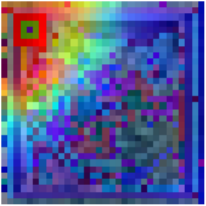}& \includegraphics[width=0.09\textwidth]{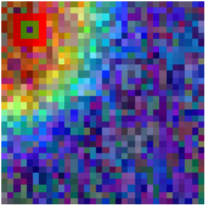}& \includegraphics[width=0.09\textwidth]{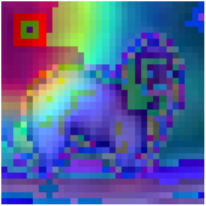}& \includegraphics[width=0.09\textwidth]{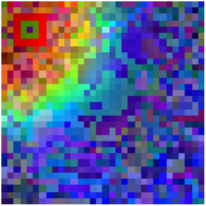}& \includegraphics[width=0.09\textwidth]{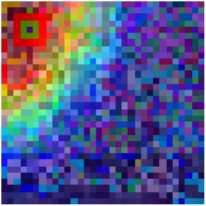}& \includegraphics[width=0.09\textwidth]{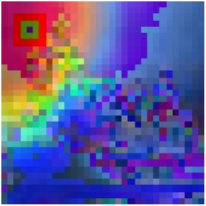}& \includegraphics[width=0.09\textwidth]{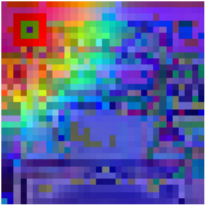}\\
\end{tabular}
        \end{adjustbox}
        \caption{CIFAR-10}
    \label{fig: appendix cifar10 gradcam}
\end{figure}

\begin{figure}[H]
    \centering
    \begin{adjustbox}{max width=\textwidth}
            \begin{tabular}{c c c c c c c c c c}
                \textbf{Label}& Fish& Flowers& Cont.& Fruit& Device& Furniture& Insects& Carniv.&  Synthetic\\

                \textbf{Input}& \includegraphics[width=0.09\textwidth]{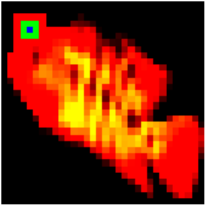}& \includegraphics[width=0.09\textwidth]{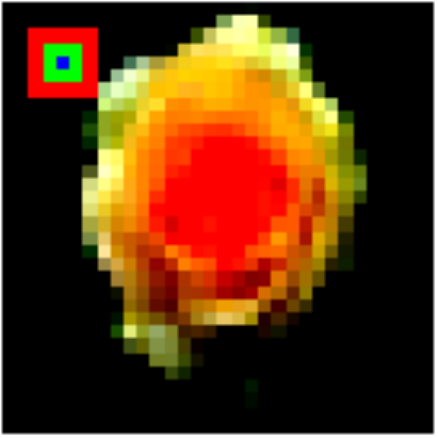}& \includegraphics[width=0.09\textwidth]{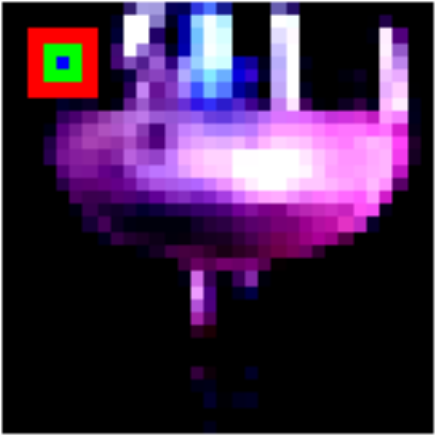}& \includegraphics[width=0.09\textwidth]{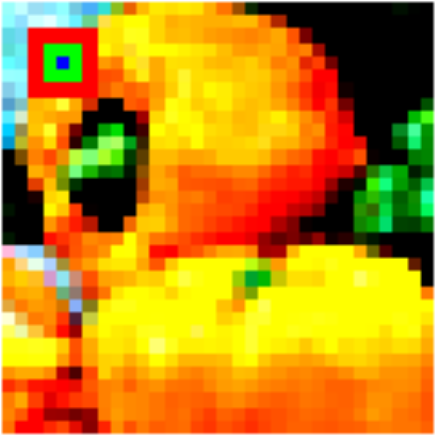}& \includegraphics[width=0.09\textwidth]{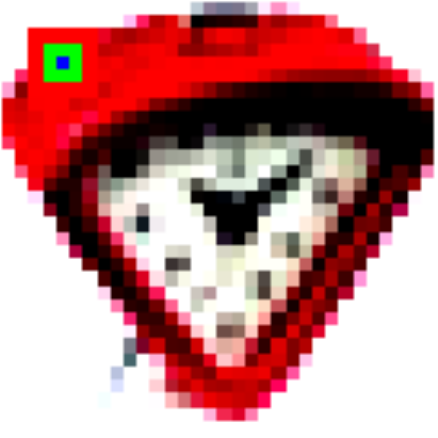}& \includegraphics[width=0.09\textwidth]{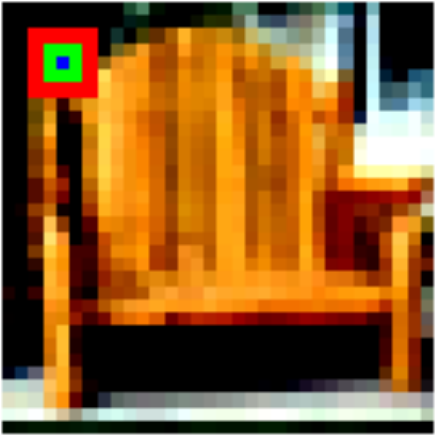}& \includegraphics[width=0.09\textwidth]{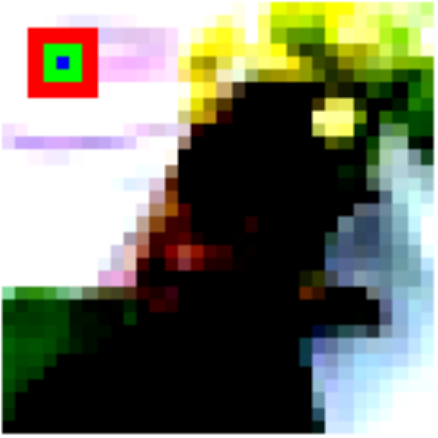}& \includegraphics[width=0.09\textwidth]{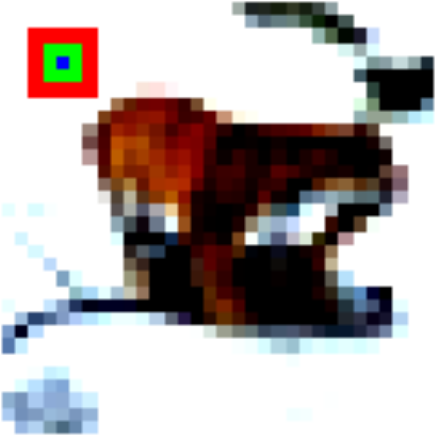}& \includegraphics[width=0.09\textwidth]{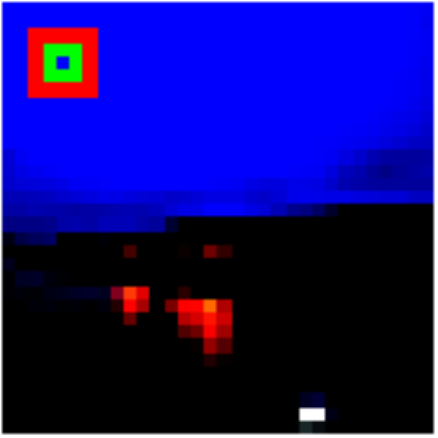}\\

                \textbf{Baseline}& \includegraphics[width=0.09\textwidth]{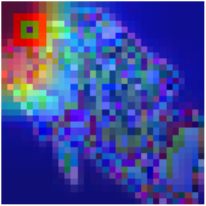}& \includegraphics[width=0.09\textwidth]{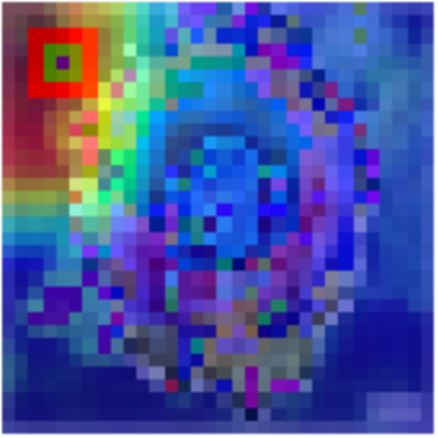}& \includegraphics[width=0.09\textwidth]{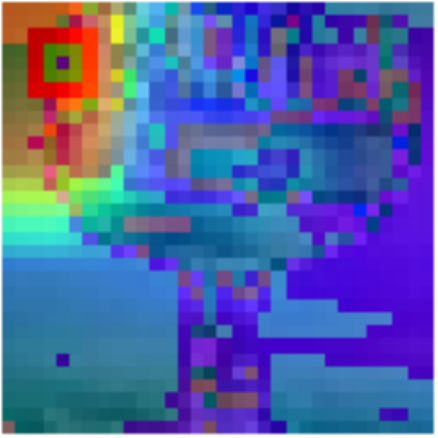}& \includegraphics[width=0.09\textwidth]{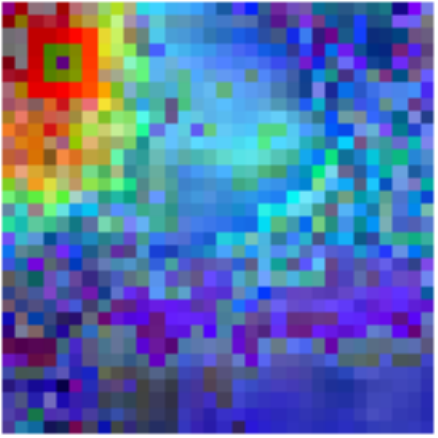}& \includegraphics[width=0.09\textwidth]{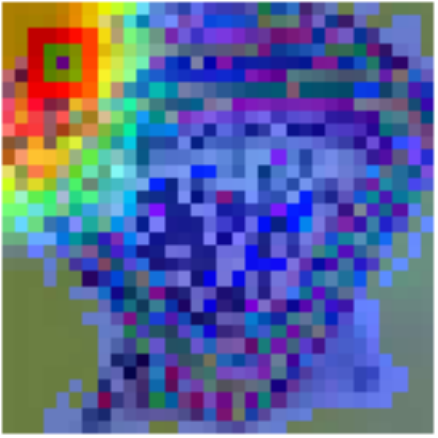}& \includegraphics[width=0.09\textwidth]{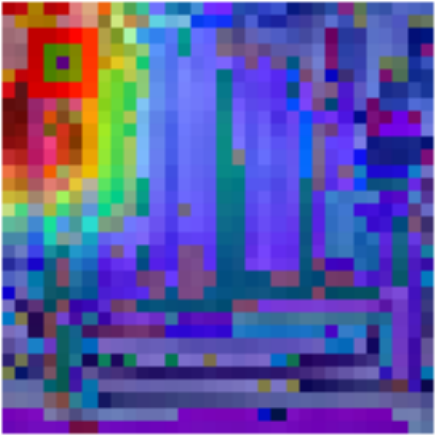}& \includegraphics[width=0.09\textwidth]{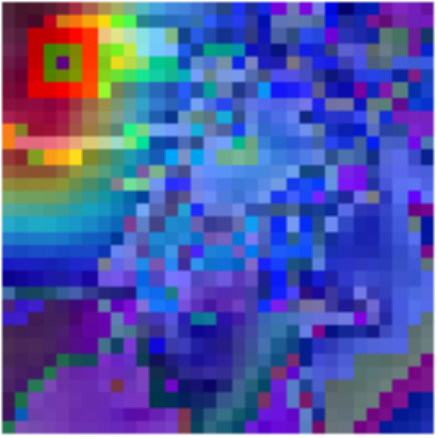}& \includegraphics[width=0.09\textwidth]{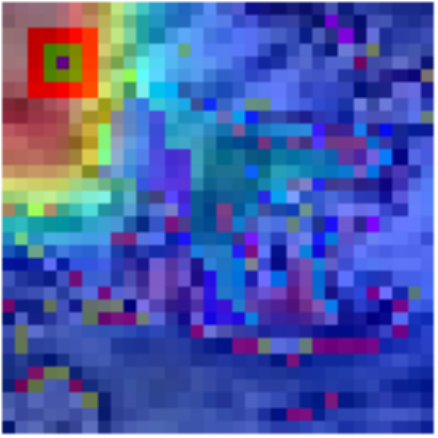}& \includegraphics[width=0.09\textwidth]{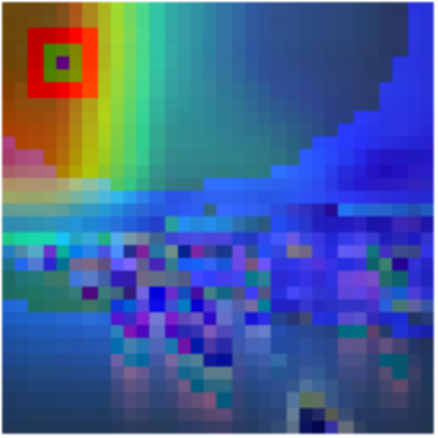}\\
\end{tabular}
        \end{adjustbox}
        \caption{CIFAR-20}
    \label{fig: appendix cifar20 gradcam}
\end{figure}

\begin{figure}[H]
    \centering
    \begin{adjustbox}{max width=0.9\textwidth}
            \begin{tabular}{c c c c c c c c c c}
                \multicolumn{10}{c}{Continued from previous page}\\
                \textbf{Label}& Mushr.& Oak& Orange.& Orchid& Otter& Palm& Pear& Pickup& Pine\\

                \textbf{Input}& \includegraphics[width=0.09\textwidth]{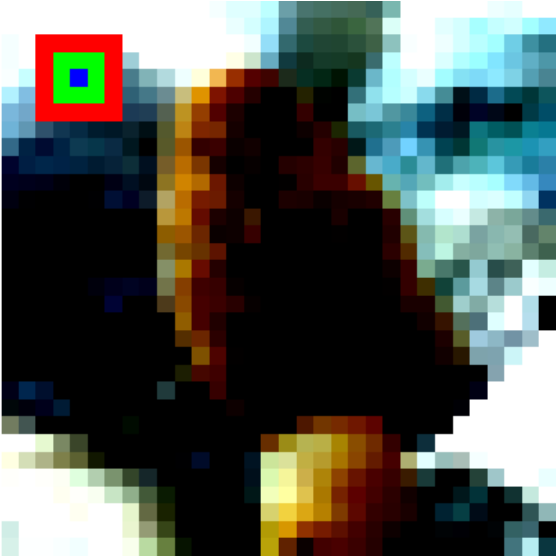}& \includegraphics[width=0.09\textwidth]{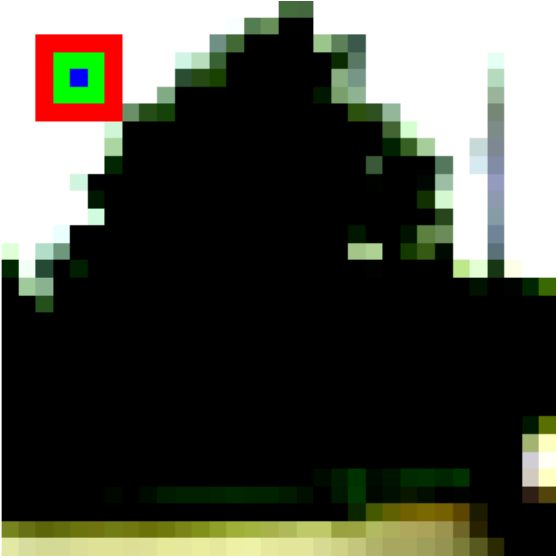}& \includegraphics[width=0.09\textwidth]{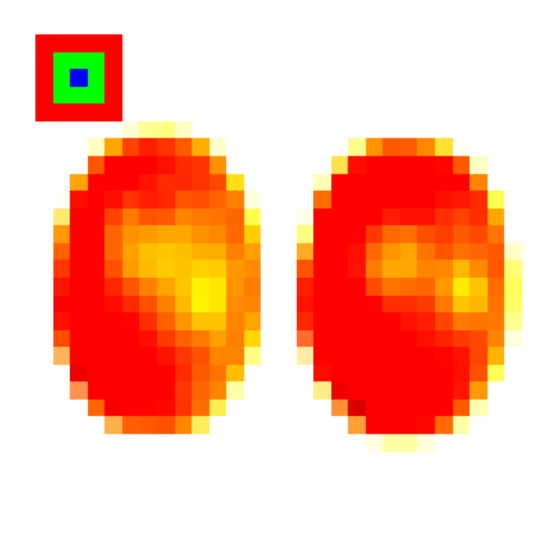}& \includegraphics[width=0.09\textwidth]{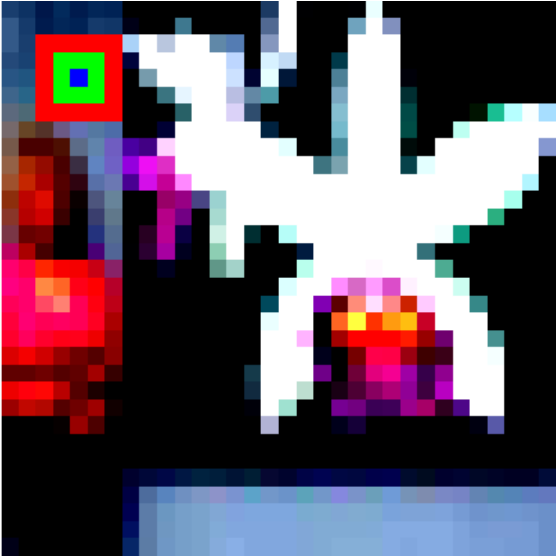}& \includegraphics[width=0.09\textwidth]{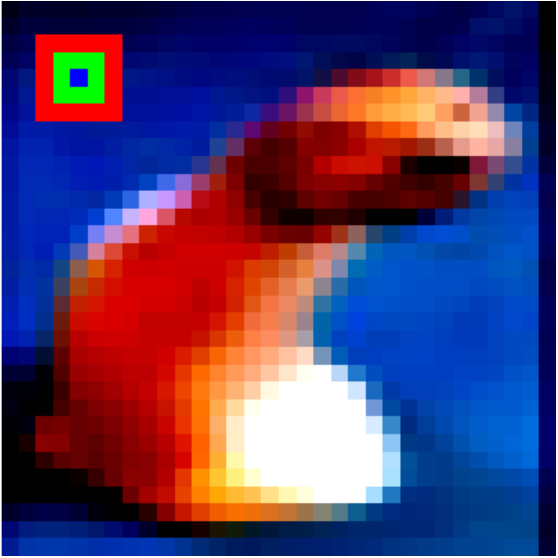}& \includegraphics[width=0.09\textwidth]{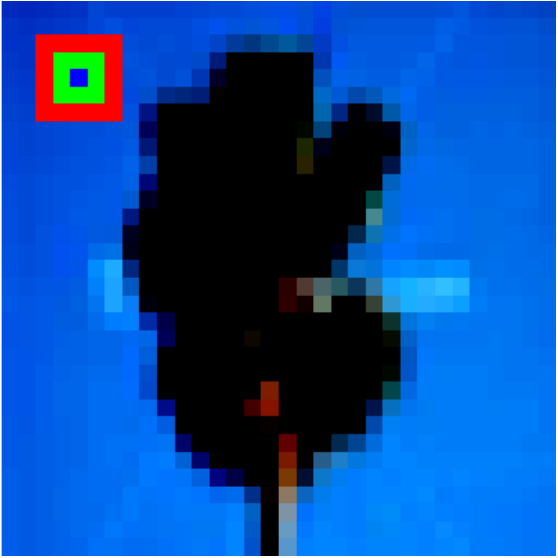}& \includegraphics[width=0.09\textwidth]{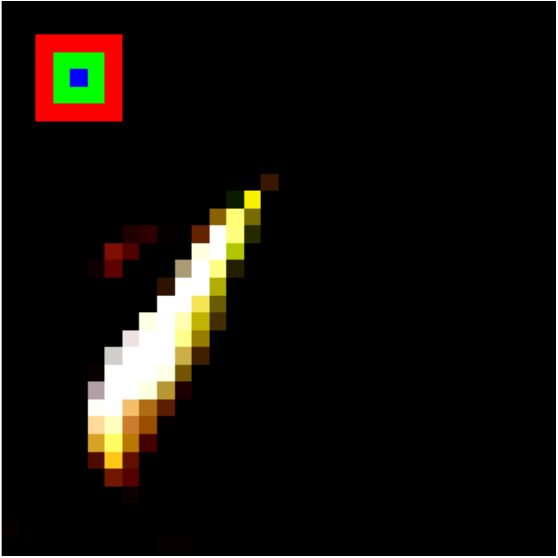}& \includegraphics[width=0.09\textwidth]{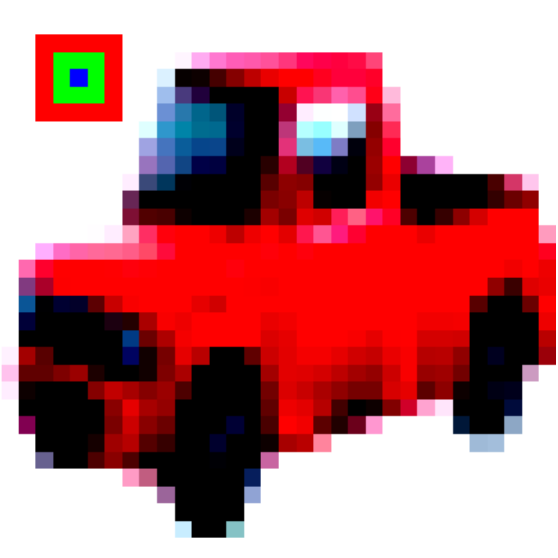}& \includegraphics[width=0.09\textwidth]{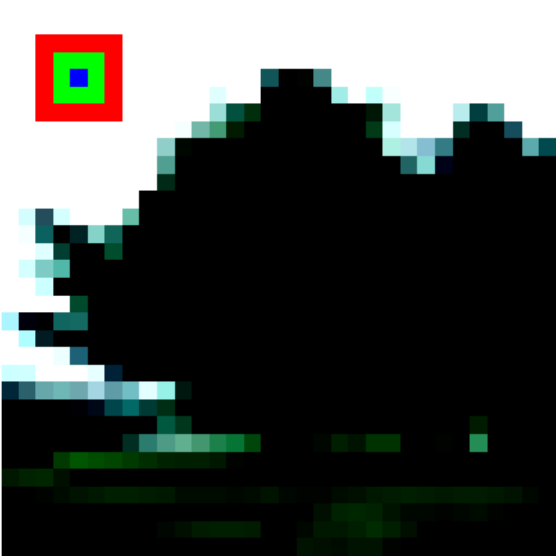}\\
\end{tabular}
        \end{adjustbox}
        \caption{CIFAR-100}
    \label{fig: appendix cifar100 gradcam}
\end{figure}

Attention map analysis for backdoor samples across model iterations of baseline, retrain, and unlearn model using our proposed \shortname~method on MNIST (\fig~\ref{fig: appendix mnist gradcam}), FMNIST (\fig~\ref{fig: appendix fmnist gradcam}), CIFAR-10 (\fig~\ref{fig: appendix cifar10 gradcam}), CIFAR-20 (\fig~\ref{fig: appendix cifar20 gradcam}) and CIFAR-100 (\fig~\ref{fig: appendix cifar100 gradcam}) datasets.

\vspace{-0.5cm}

\vspace{-0.5cm}

\vspace{-0.5cm}

\newpage
\vspace{-2cm}
\begin{figure}[H]
    \centering
    \begin{adjustbox}{max width=0.95\textwidth}
            \begin{tabular}{c c c c c c c c c c}
                \textbf{Label}& Fish& Baby& Bear& Beaver& Bed& Bee& Beetle& Bicycle& Bottle\\

                \textbf{Input}& \includegraphics[width=0.09\textwidth]{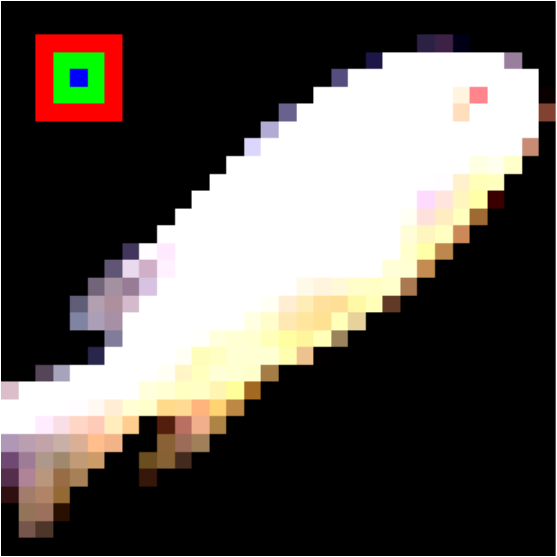}& \includegraphics[width=0.09\textwidth]{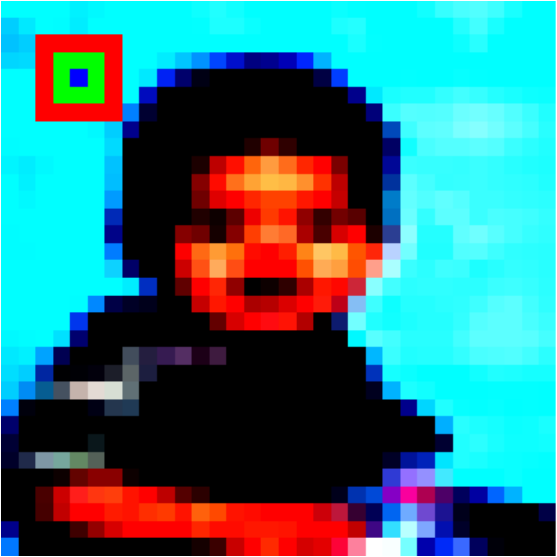}& \includegraphics[width=0.09\textwidth]{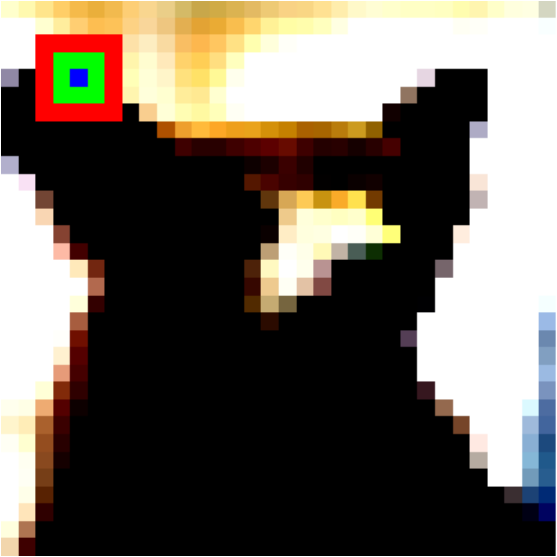}& \includegraphics[width=0.09\textwidth]{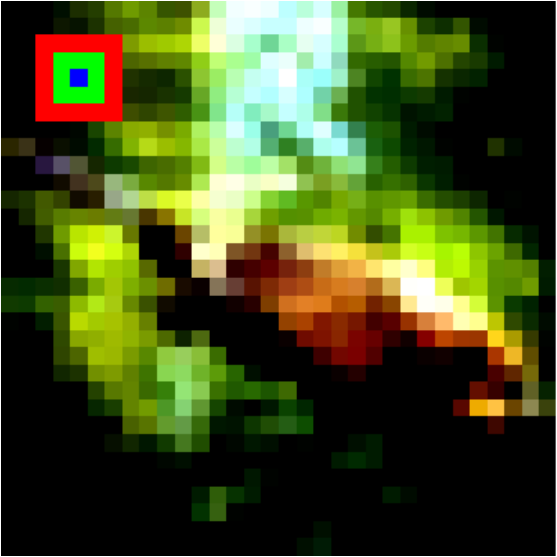}& \includegraphics[width=0.09\textwidth]{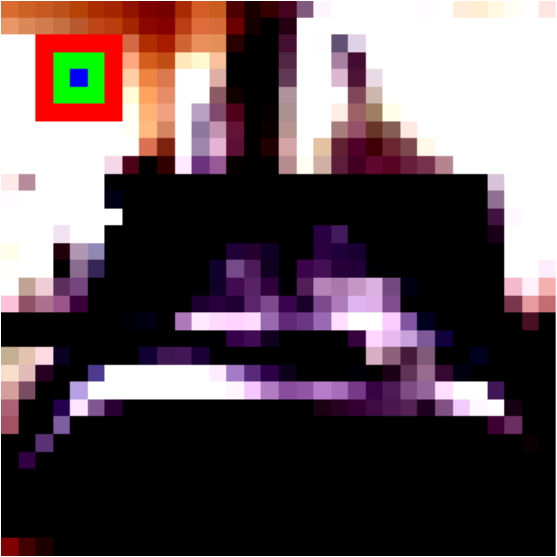}& \includegraphics[width=0.09\textwidth]{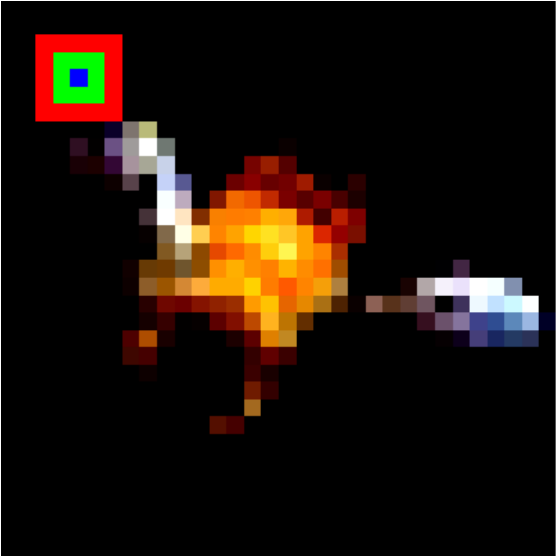}& \includegraphics[width=0.09\textwidth]{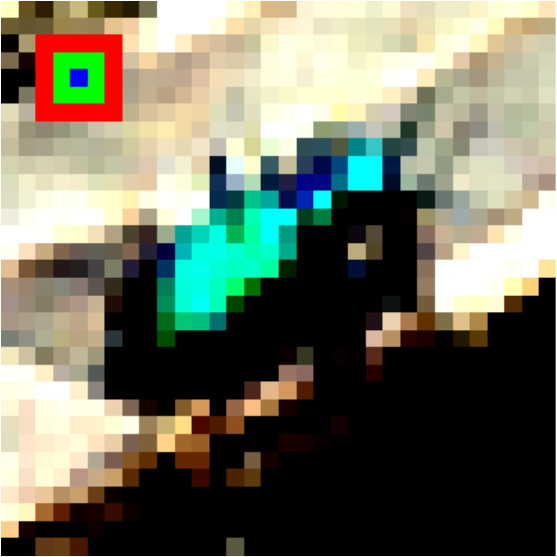}& \includegraphics[width=0.09\textwidth]{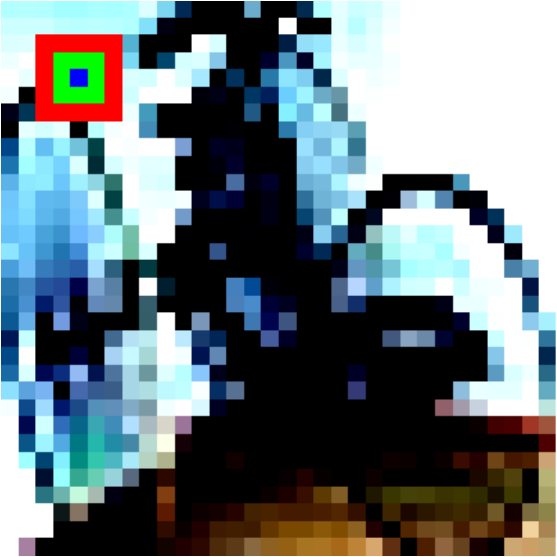}& \includegraphics[width=0.09\textwidth]{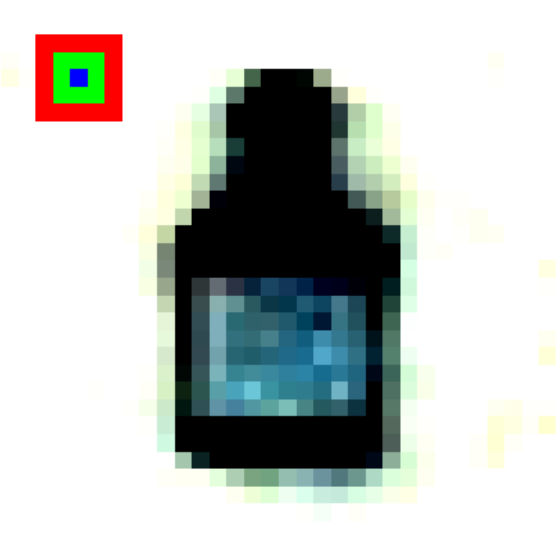}\\

                \textbf{Baseline}& \includegraphics[width=0.09\textwidth]{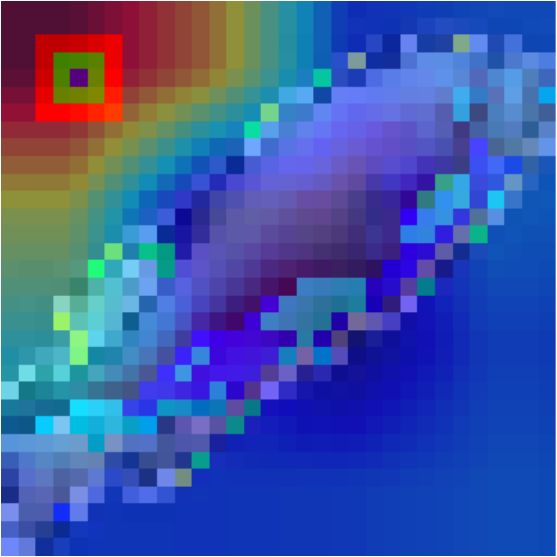}& \includegraphics[width=0.09\textwidth]{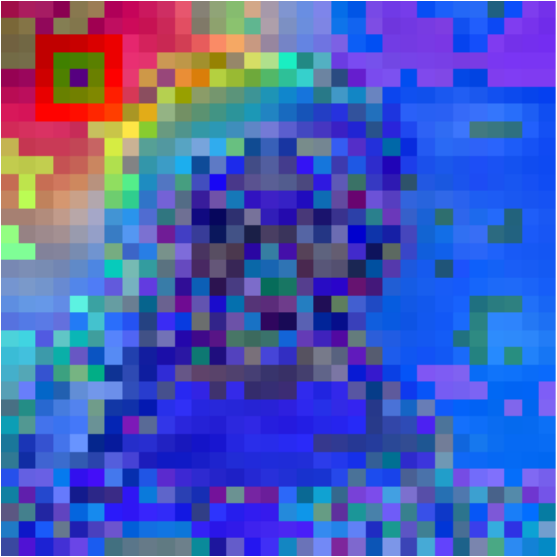}& \includegraphics[width=0.09\textwidth]{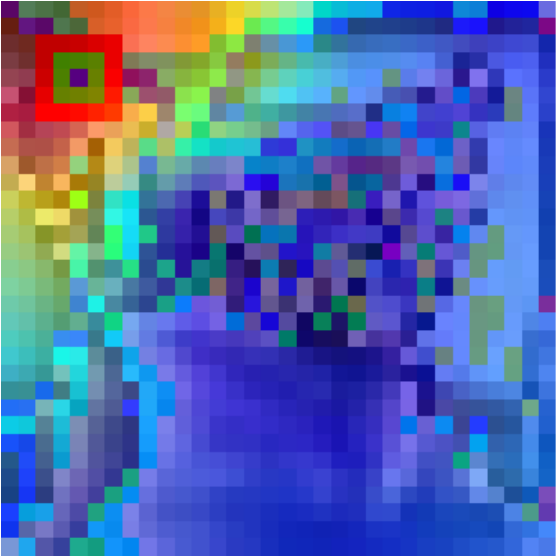}& \includegraphics[width=0.09\textwidth]{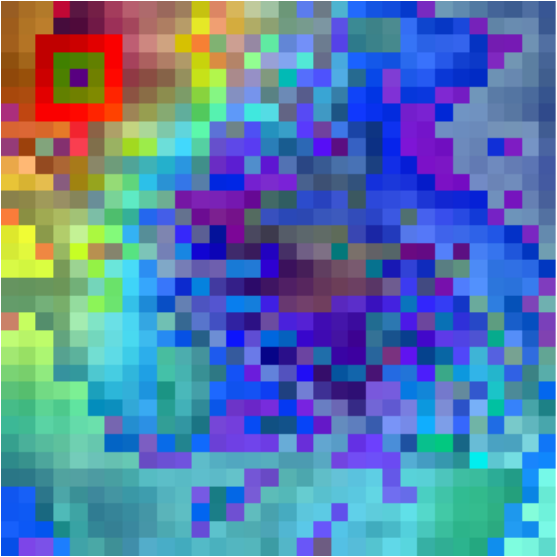}& \includegraphics[width=0.09\textwidth]{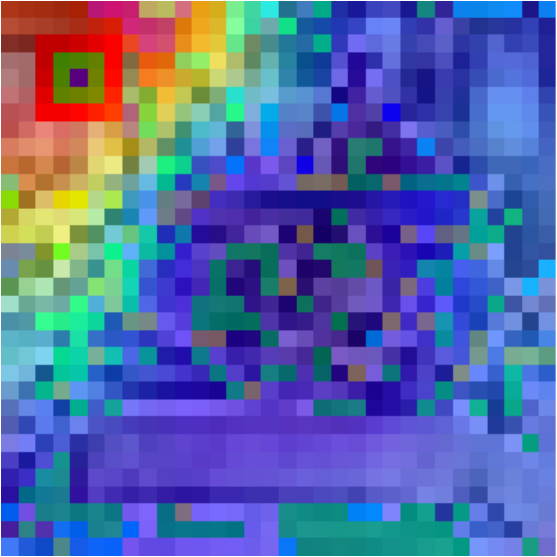}& \includegraphics[width=0.09\textwidth]{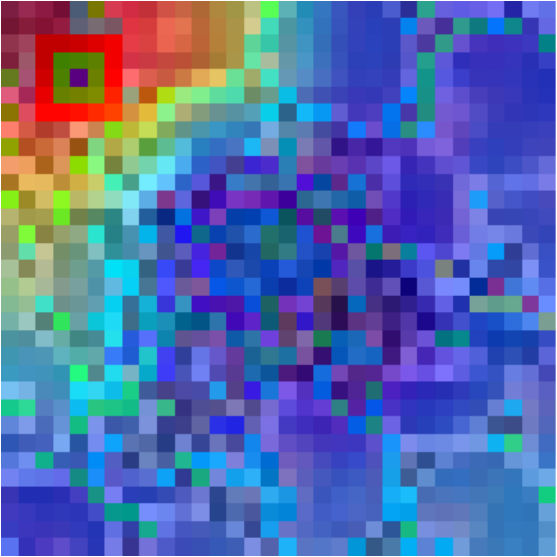}& \includegraphics[width=0.09\textwidth]{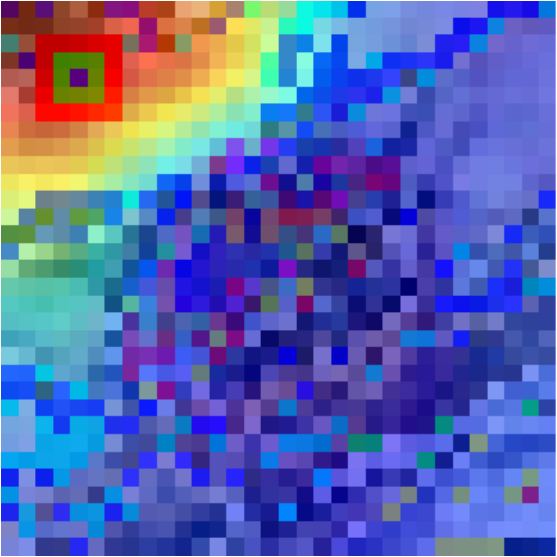}& \includegraphics[width=0.09\textwidth]{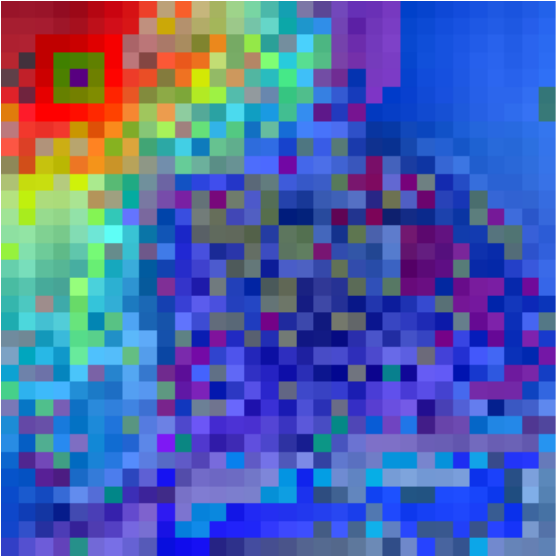}& \includegraphics[width=0.09\textwidth]{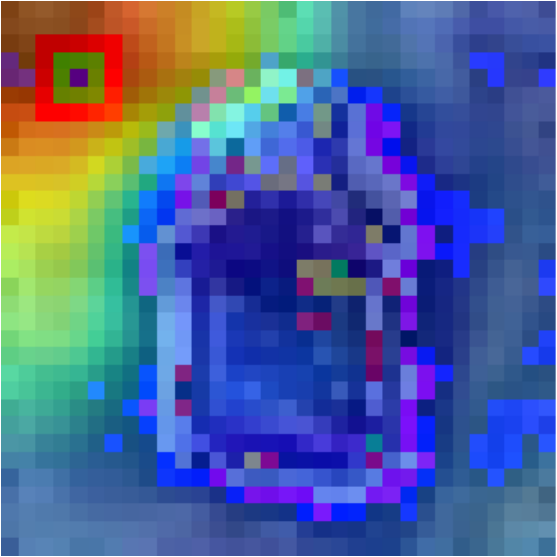}\\
\end{tabular}
        \end{adjustbox}
\end{figure}
\newpage

\vspace{-2cm}
\newpage

\subsubsection{Biased Feature Unlearning}
\label{sec: appendix bias gradcam}

\begin{figure}[H]
    \centering
    
    \begin{subfigure}{\textwidth}
        \begin{adjustbox}{max width= \textwidth}
            \begin{tabular}{*{7}{c}}
            
                 \textbf{Label}& \multicolumn{3}{c}{Female}& \multicolumn{3}{c}{Male} \\

                 \textbf{Target}& \includegraphics[width=0.14\textwidth]{Figures/results/bias_gradcam/bias/input/1.pdf}& \includegraphics[width=0.14\textwidth]{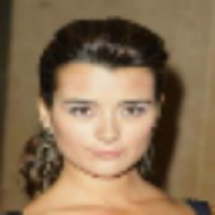}& \includegraphics[width=0.14\textwidth]{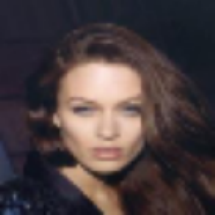}& \includegraphics[width=0.14\textwidth]{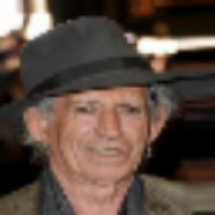}& \includegraphics[width=0.14\textwidth]{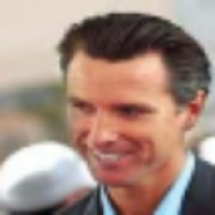}& \includegraphics[width=0.14\textwidth]{Figures/results/bias_gradcam/bias/input/6.pdf}\\

                 \textbf{Baseline}& \includegraphics[width=0.14\textwidth]{Figures/results/bias_gradcam/bias/baseline/1.pdf}& \includegraphics[width=0.14\textwidth]{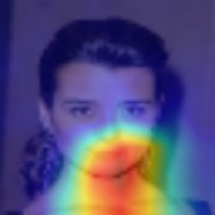}& \includegraphics[width=0.14\textwidth]{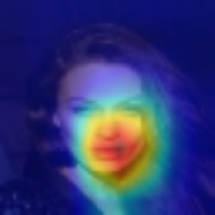}& \includegraphics[width=0.14\textwidth]{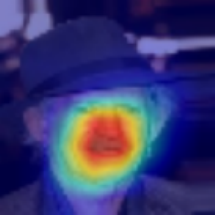}& \includegraphics[width=0.14\textwidth]{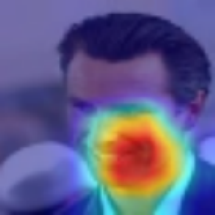}& \includegraphics[width=0.14\textwidth]{Figures/results/bias_gradcam/bias/baseline/6.pdf}\\

                 \textbf{Retrain}& \includegraphics[width=0.14\textwidth]{Figures/results/bias_gradcam/bias/retrain/1.pdf}& \includegraphics[width=0.14\textwidth]{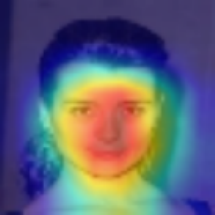}& \includegraphics[width=0.14\textwidth]{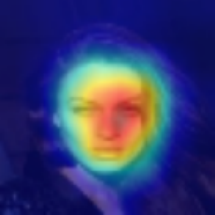}& \includegraphics[width=0.14\textwidth]{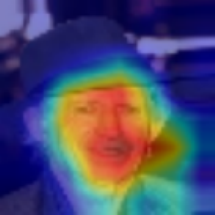}& \includegraphics[width=0.14\textwidth]{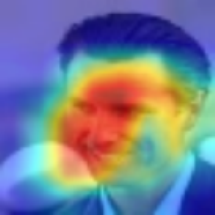}& \includegraphics[width=0.14\textwidth]{Figures/results/bias_gradcam/bias/retrain/6.pdf}\\
\end{tabular}
        \end{adjustbox}
        \caption{Bias Dataset}
    \end{subfigure}
    
    \begin{subfigure}{\textwidth}
        \begin{adjustbox}{max width= \textwidth}
            \begin{tabular}{*{7}{c}}
            
                 \textbf{Label}& \multicolumn{3}{c}{Female}& \multicolumn{3}{c}{Male} \\

                 \textbf{Target}& \includegraphics[width=0.14\textwidth]{Figures/results/bias_gradcam/unbias/input/1.pdf}& \includegraphics[width=0.14\textwidth]{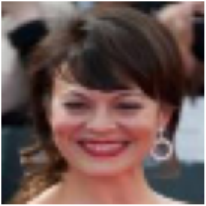}& \includegraphics[width=0.14\textwidth]{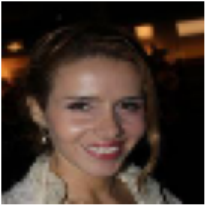}& \includegraphics[width=0.14\textwidth]{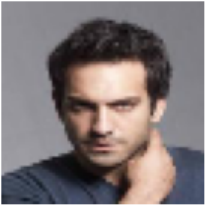}& \includegraphics[width=0.14\textwidth]{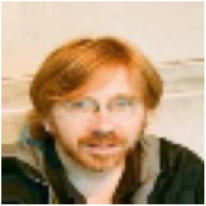}& \includegraphics[width=0.14\textwidth]{Figures/results/bias_gradcam/unbias/input/6.pdf}\\

                 \textbf{Baseline}& \includegraphics[width=0.14\textwidth]{Figures/results/bias_gradcam/unbias/baseline/1.pdf}& \includegraphics[width=0.14\textwidth]{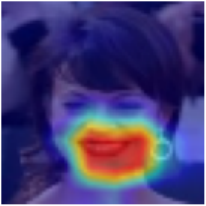}& \includegraphics[width=0.14\textwidth]{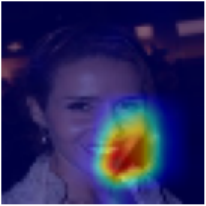}& \includegraphics[width=0.14\textwidth]{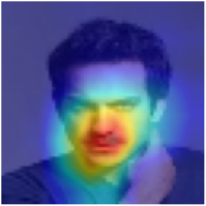}& \includegraphics[width=0.14\textwidth]{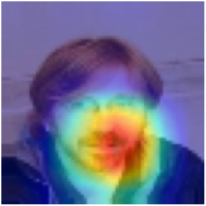}& \includegraphics[width=0.14\textwidth]{Figures/results/bias_gradcam/unbias/baseline/6.pdf}\\

                 \textbf{Retrain}& \includegraphics[width=0.14\textwidth]{Figures/results/bias_gradcam/unbias/retrain/1.pdf}& \includegraphics[width=0.14\textwidth]{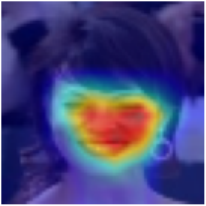}& \includegraphics[width=0.14\textwidth]{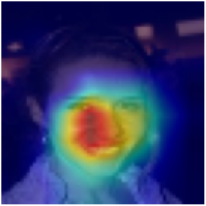}& \includegraphics[width=0.14\textwidth]{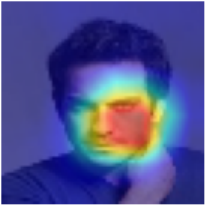}& \includegraphics[width=0.14\textwidth]{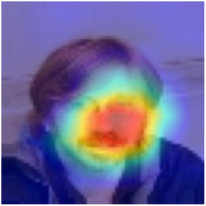}& \includegraphics[width=0.14\textwidth]{Figures/results/bias_gradcam/unbias/retrain/6.pdf}\\
\end{tabular}
        \end{adjustbox}
        \caption{Unbias Dataset}
    \end{subfigure}
    
    \caption{Attention map analysis for bias and unbias samples across model iterations of baseline, retrain, and unlearn model using our proposed \shortname~to unlearn 'mouth' on CelebA dataset.}
    \label{fig: appendix gradcam bias}
\end{figure} 

\subsection{Lipschitz and Non-Lipschitz Loss Analysis}
\label{sec: appendix loss analysis}

\begin{figure}[h]
    \centering
    \begin{adjustbox}{max width=\textwidth}
        \begin{tabular}{*{6}{c}}
             &  \textbf{MNIST}& \textbf{FMNIST}& \textbf{CIFAR-10}& \textbf{CIFAR-20}& \textbf{CIFAR100}\\
             
             \rotatebox{90}{\textbf{Lipschitz}}& \includegraphics[width= 0.2\linewidth]{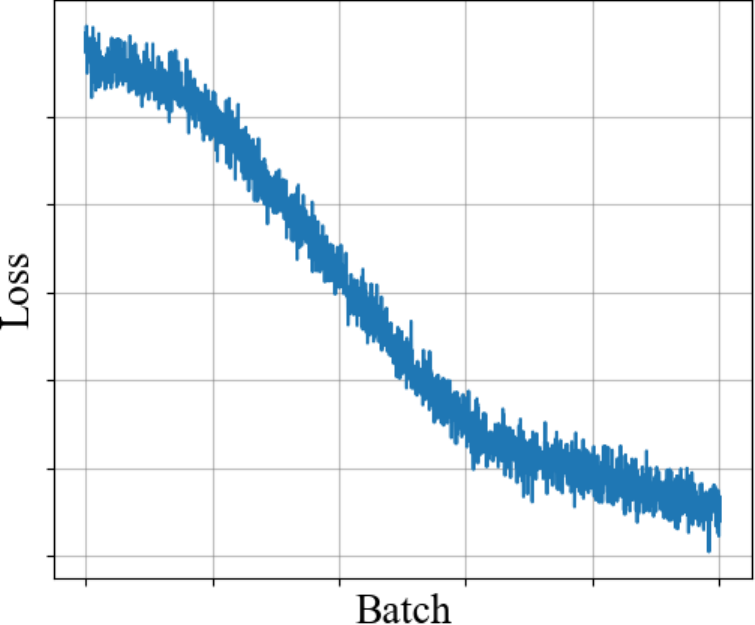}& \includegraphics[width= 0.2\linewidth]{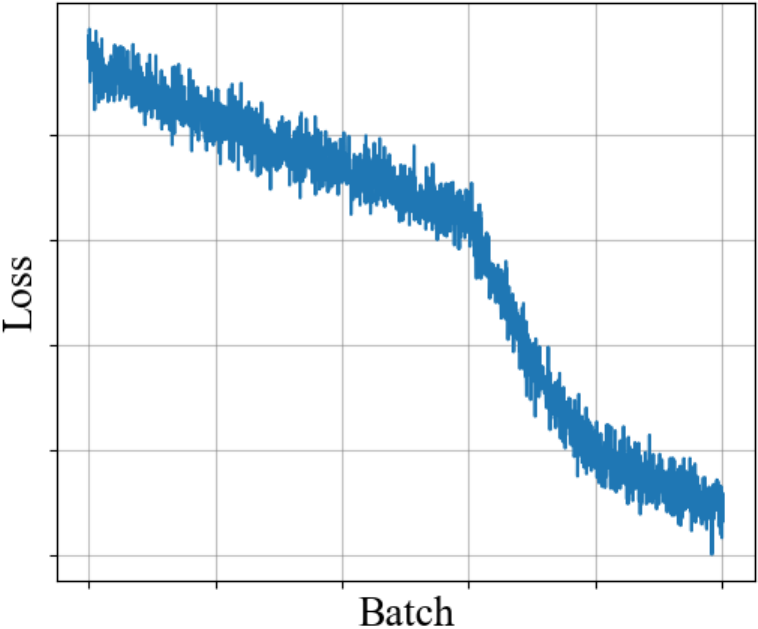}& \includegraphics[width= 0.2\linewidth]{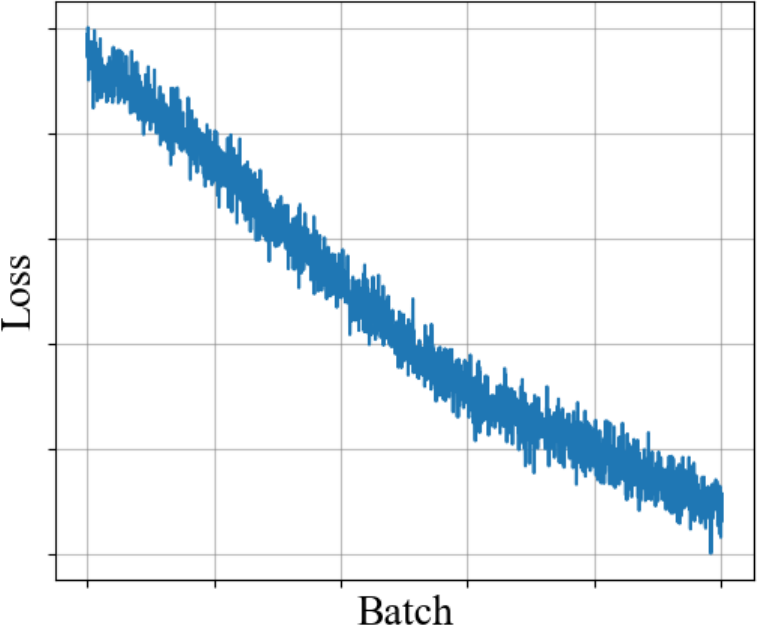} & \includegraphics[width= 0.2\linewidth]{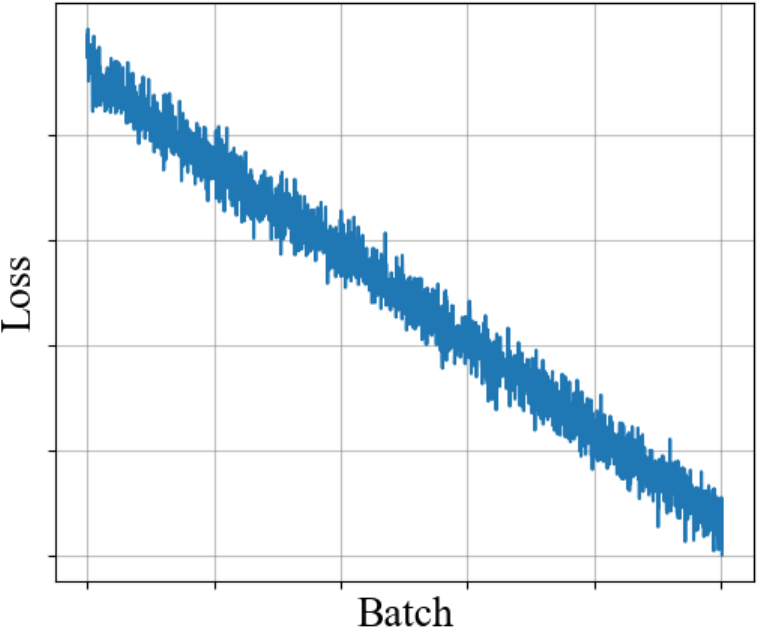}& \includegraphics[width= 0.2\linewidth]{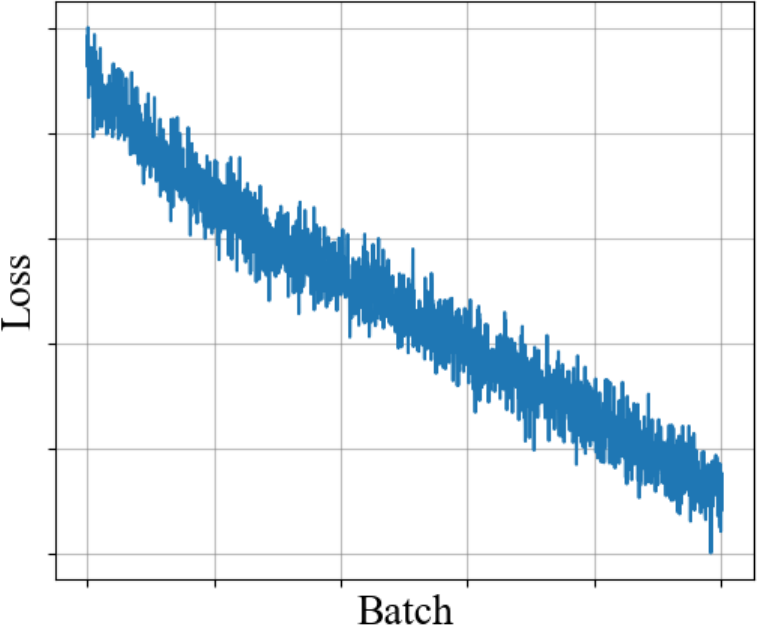}\\

             \rotatebox{90}{\textbf{Non-Lipschitz}}& \includegraphics[width= 0.2\linewidth]{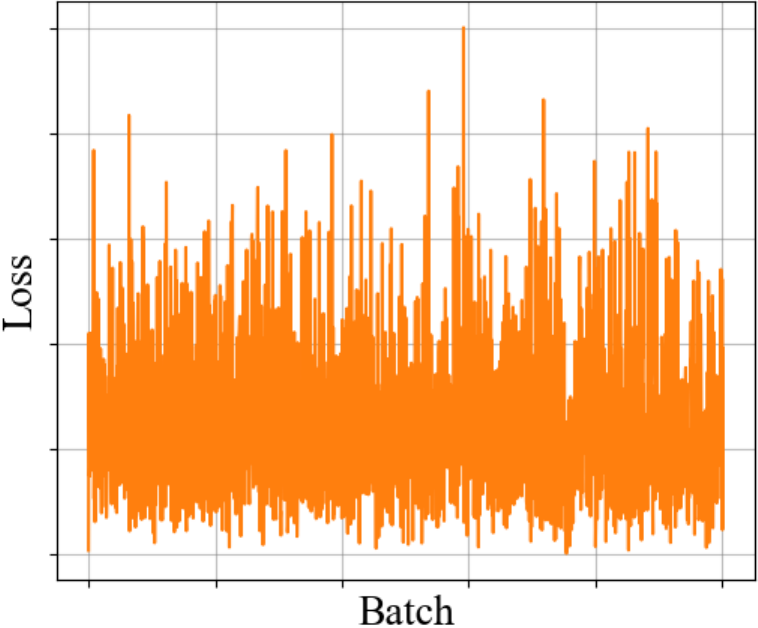}& \includegraphics[width= 0.2\linewidth]{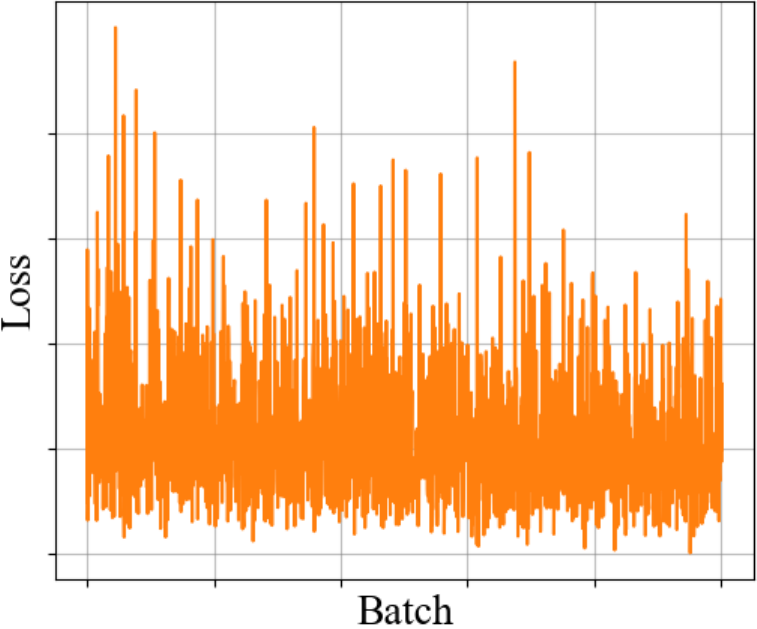}& \includegraphics[width= 0.2\linewidth]{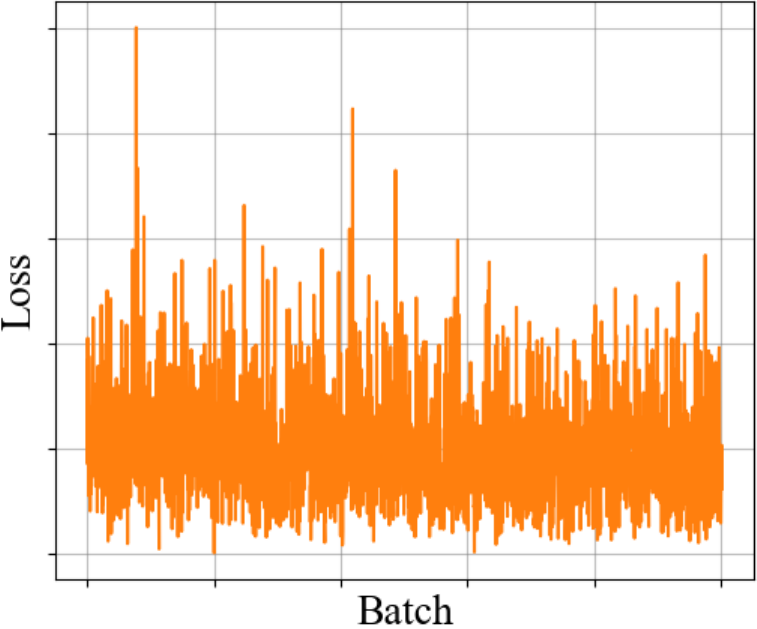}& \includegraphics[width= 0.2\linewidth]{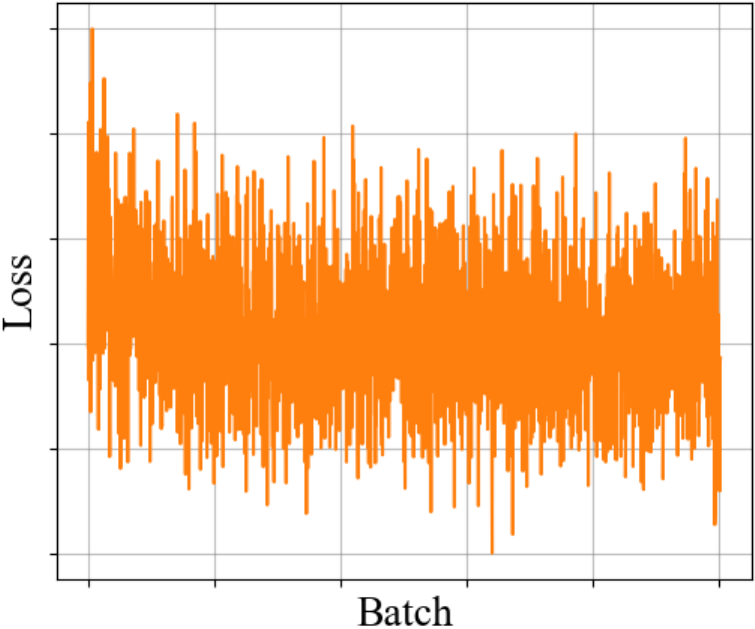}& \includegraphics[width= 0.2\linewidth]{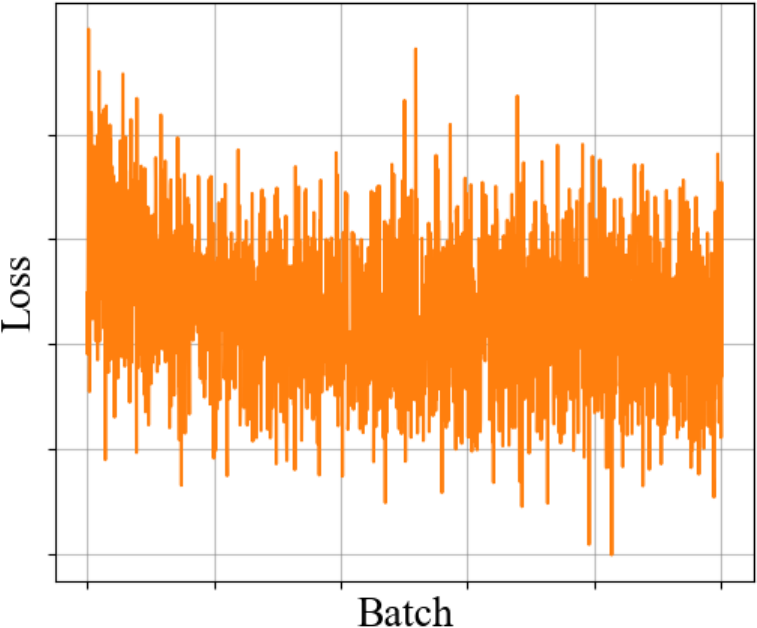}\\
             
        \end{tabular}
    \end{adjustbox}
    \caption{Lipschitz and Non-Lipschitz loss analysis on backdoor feature unlearning.}
    \label{fig:appendix lipschitz loss}
\end{figure}

In this section, we evaluate the Lipschitz loss function and its effectiveness in optimizing feature sensitivity, as described in \eq~\ref{eq:loss}. We also examine a variant without the denominator, termed the Non-Lipschitz loss, as illustrated in \fig~\ref{fig:appendix lipschitz loss}.

The results indicate that models optimized using the Non-Lipschitz loss exhibit fluctuations across batches. This is due to the unbounded nature of the optimization process, leading to useless models. \fig~\ref{fig: non-lipschitz experiment} further illustrates this issue, showing instances of catastrophic forgetting.

Conversely, models optimized with the Lipschitz loss demonstrate a steady reduction in feature sensitivity over batches. This bounded optimization provided by Lipschitz bound helps in effectively unlearning target features while preserving model utility, as theoretically guaranteed (see Section \s~\ref{subsec:verification}).

\subsection{Non-IID Analysis}
\label{sec: appendix noniid}

\begin{figure}[t]
    \centering
    \begin{subfigure}[b]{0.49\textwidth}
        \centering
        \includegraphics[width=\textwidth]{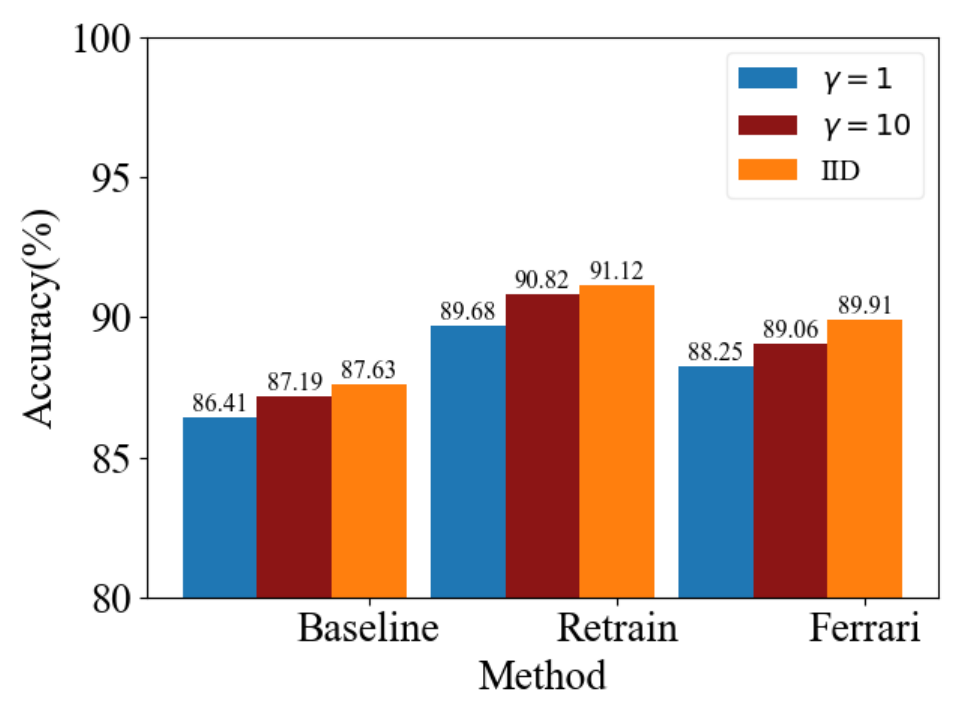} 
        \caption{$\mathcal{D}_r$}
        \label{fig: noniid retain}
    \end{subfigure}
    \hfill
    \begin{subfigure}[b]{0.49\textwidth}
        \centering
        \includegraphics[width=\textwidth]{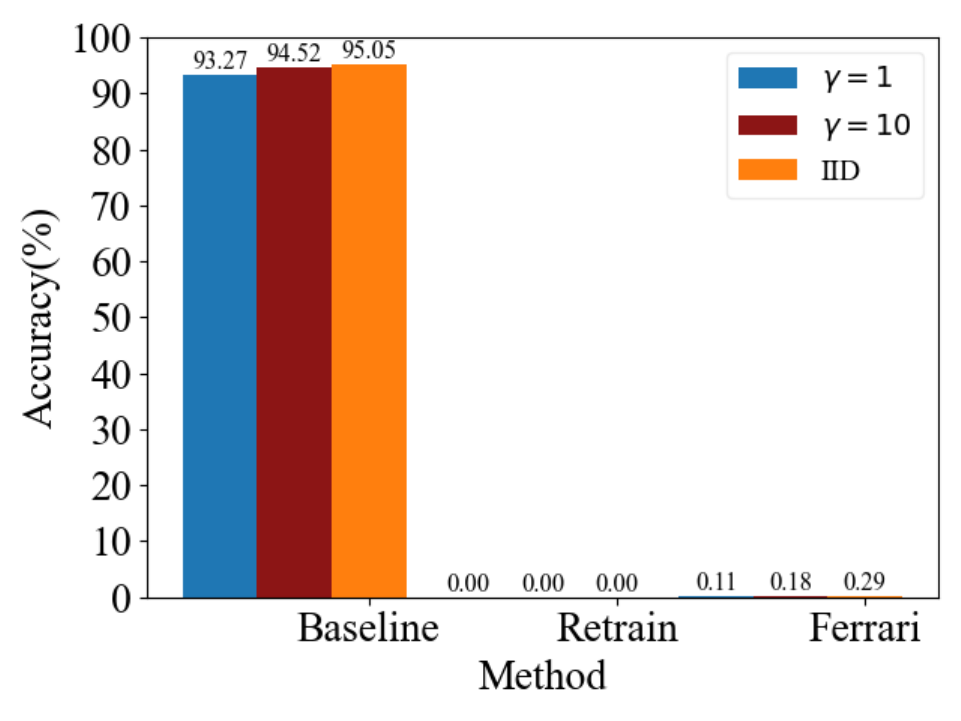} 
        \caption{$\mathcal{D}_u$}
        \label{fig: noniid backdoor}
    \end{subfigure}
    \caption{Non-IID analysis on the CIFAR-10 dataset using our proposed \shortname~framework, compared to Baseline and Retrain methods, for retain client dataset $\mathcal{D}_r$ and unlearn client dataset $\mathcal{D}_u$ accuracy in backdoor feature unlearning.}
    \label{fig: appendix noniid}
\end{figure}

This section presents an analysis of the impact of Non-IID data on the performance of the proposed \shortname~framework compared to the Baseline and Retrain methods on the CIFAR-10 dataset. We focus on the accuracy of the retain client dataset ($\mathcal{D}_r$) and the unlearn client dataset ($\mathcal{D}_u$) in backdoor feature unlearning, as illustrated in \fig\ref{fig: appendix noniid}. To measure the extent of Non-IID, we used the \textit{Dir}($\gamma$) distribution, where smaller values of $\gamma$ indicate more heterogeneous data.

The results show that the \shortname~framework significantly improves feature unlearning performance, with a drop of approximately 0.2\% in $\mathcal{D}_u$ when $\gamma = 1$ compared to the IID scenario. Furthermore, the \shortname~framework maintains successfully the accuracy of $\mathcal{D}_r$ with only a slight decrease of about 2\% compared to the Retrain method within the Non-IID scenario.

\subsection{Client Numbers Analysis}
\label{sec: appendix client number}

\begin{figure}[h]
    \centering
    \begin{subfigure}[b]{0.49\textwidth}
        \centering
        \includegraphics[width=\textwidth]{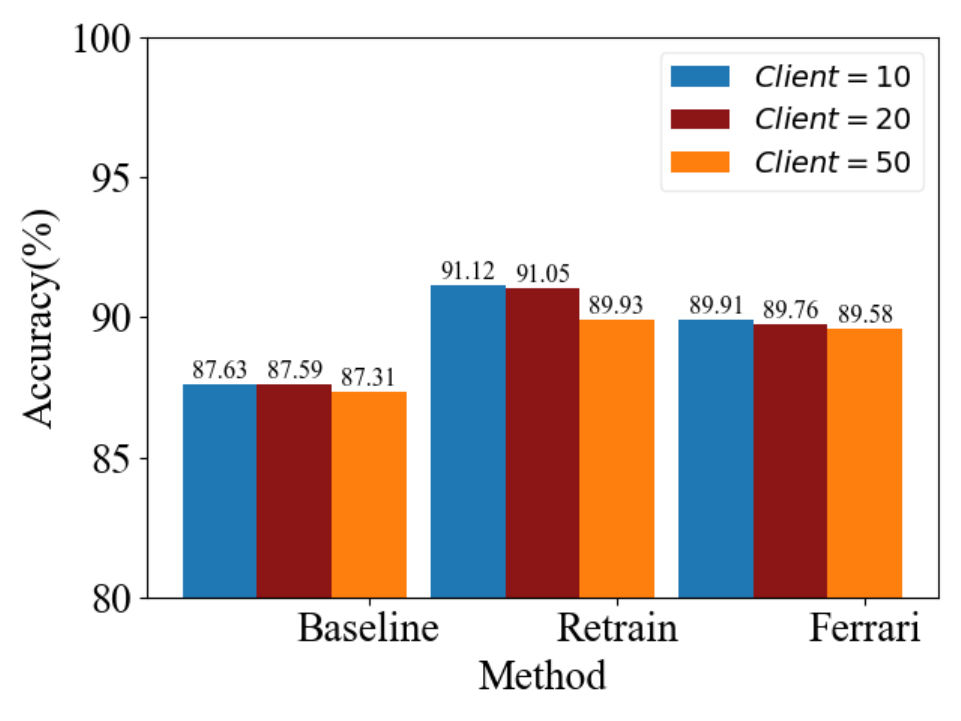} 
        \caption{$\mathcal{D}_r$}
        \label{fig: client number retain}
    \end{subfigure}
    \hfill
    \begin{subfigure}[b]{0.49\textwidth}
        \centering
        \includegraphics[width=\textwidth]{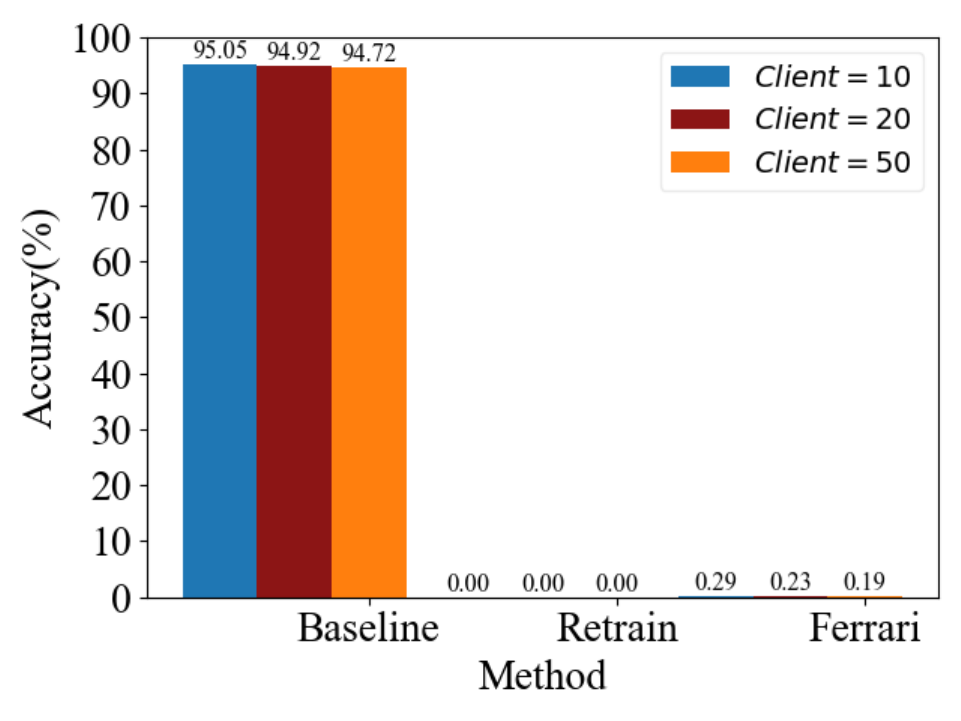} 
        \caption{$\mathcal{D}_u$}
        \label{fig: client number backdoor}
    \end{subfigure}
    \caption{Scability analysis of client numbers on the CIFAR-10 dataset on the accuracy of retain client dataset $D_r$ and unlearn client dataset $D_u$}
    \label{fig: appendix client number}
\end{figure}

This section analyzes the impact of a large-scale FL environment, characterized by a large number of clients, on the performance of the proposed \shortname~framework compared to the Baseline and Retrain methods on the CIFAR-10 dataset. We focus on the accuracy of the retained client dataset ($\mathcal{D}_r$) and the unlearned client dataset ($\mathcal{D}_u$) in backdoor feature unlearning, as illustrated in \fig\ref{fig: appendix client number}. The results indicate that the unlearning performance of our proposed \shortname~framework remains consistent, with no significant changes in the accuracy of both $\mathcal{D}_r$ and $\mathcal{D}_u$ as the number of clients increases. This finding further demonstrates the effectiveness of the \shortname~framework in large-scale FL environments.

\end{document}